\PassOptionsToPackage{table}{xcolor}

\documentclass{article}

\usepackage[preprint]{acl}

\usepackage{times}
\usepackage{latexsym}

\usepackage[T1]{fontenc}

\usepackage[utf8]{inputenc}

\usepackage{microtype}

\usepackage{inconsolata}

\usepackage{graphicx}

\usepackage{arydshln}
\usepackage{tikz}
\usepackage{caption}
\usepackage{arydshln}
\usepackage{makecell}
\usepackage{tcolorbox}  
\usepackage{listings}    
\newcommand{\model}{MAmmoTH-VL\xspace}
\usepackage{titletoc}
\usepackage{etoolbox} 
\usepackage{xspace}  
\usepackage{wrapfig}
\usepackage{amsmath}
\usepackage{pifont}

\usepackage{color-edits}
\addauthor{gn}{magenta}

\newcommand\DoToC{%
  \startcontents
  \printcontents{}{1}{\noindent \textbf{\Large{Table of Contents in Appendix}}\vskip3pt\vskip5pt}
  \vskip3pt\vskip5pt
}

\newcommand{\correct}{\textcolor{green}{\ding{51}}}
\newcommand{\wrong}{\textcolor{red}{\ding{55}}}

\usepackage[utf8]{inputenc} 
\usepackage[T1]{fontenc}    
\usepackage{hyperref}       
\usepackage{url}            
\usepackage{booktabs}       
\usepackage{amsfonts}       
\usepackage{nicefrac}       
\usepackage{microtype}      
\usepackage{fontawesome}
\usepackage{booktabs}
\usepackage{enumerate}
\usepackage{enumitem}
\usepackage{cuted}

\title{\model: Eliciting Multimodal Reasoning with \\ Instruction Tuning at Scale }

\author{Jarvis Guo$^{\clubsuit}$\thanks{Equal contribution.}\;,
    Tuney Zheng$^{\clubsuit}$\footnotemark[1]\;, 
    Yuelin Bai$^{\clubsuit}$\;,
    Bo Li$^{\triangle}$\;\\
    \textbf{Yubo Wang}$^{\heartsuit}$\;,
    \textbf{King Zhu}$^{\clubsuit}$\;,
    \textbf{Yizhi Li}$^{\diamondsuit}$\;\\
     \textbf{Graham Neubig}$^{\spadesuit}$\;,
    \textbf{Wenhu Chen}$^{\heartsuit}$\;,
    \textbf{Xiang Yue}$^{\spadesuit}$\thanks{\faEnvelope~ Corresponding author: xyue2@andrew.cmu.edu } \\[2mm]
    \small $^{\spadesuit}$Carnegie Mellon University $^{\clubsuit}$M-A-P\\
    \small $^{\triangle}$Nanyang Technological University
    \small $^{\heartsuit}$University of Waterloo\\
    \small $^{\diamondsuit}$The University of Manchester
    \\[2mm]
    \faHome: \url{https://mammoth-vl.github.io}
}

\begin{document}
\maketitle

\begin{strip}
    \vspace*{-2cm}
    \centering
    \includegraphics[width=0.24\linewidth]{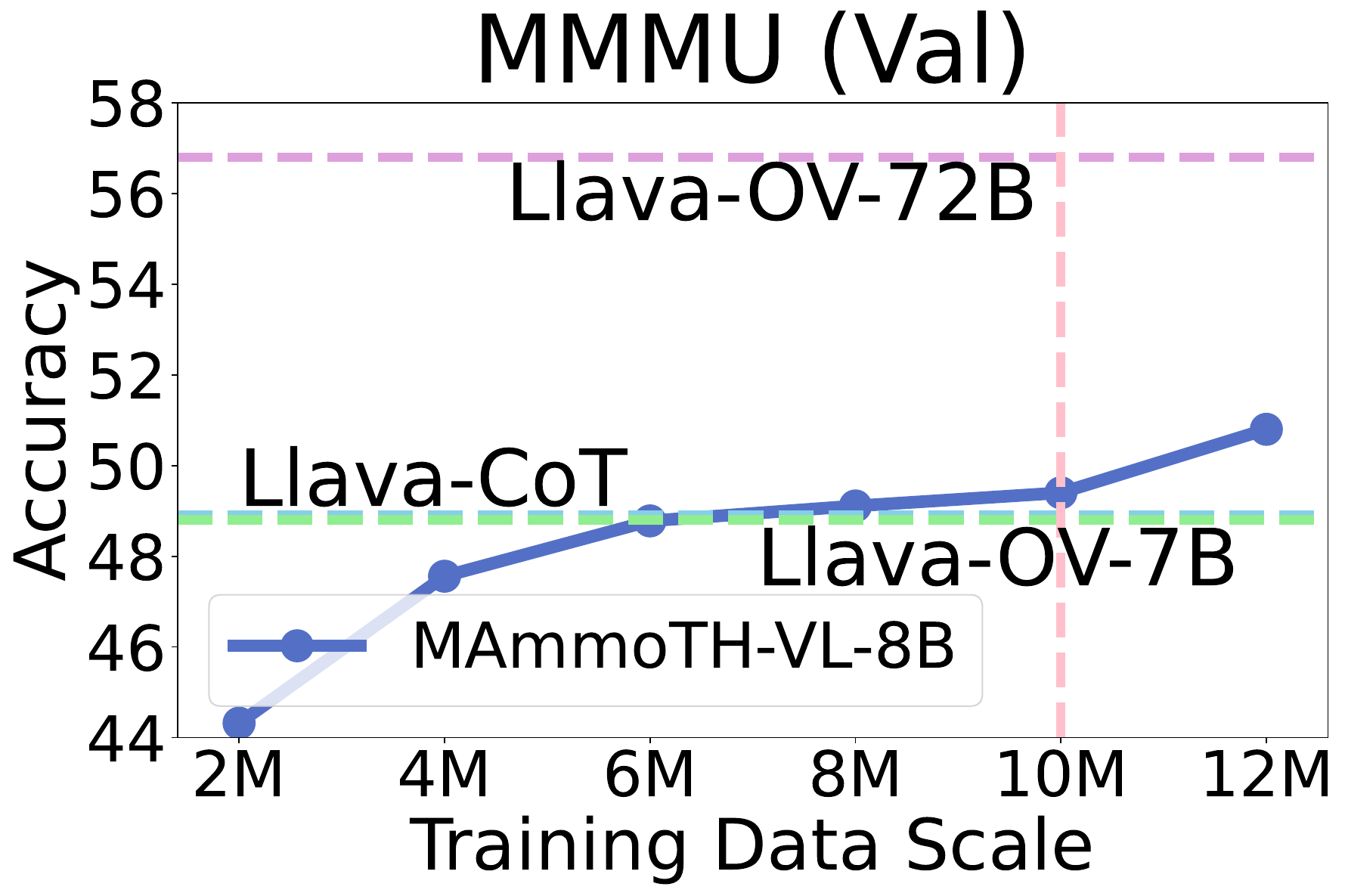}
    \includegraphics[width=0.24\linewidth]{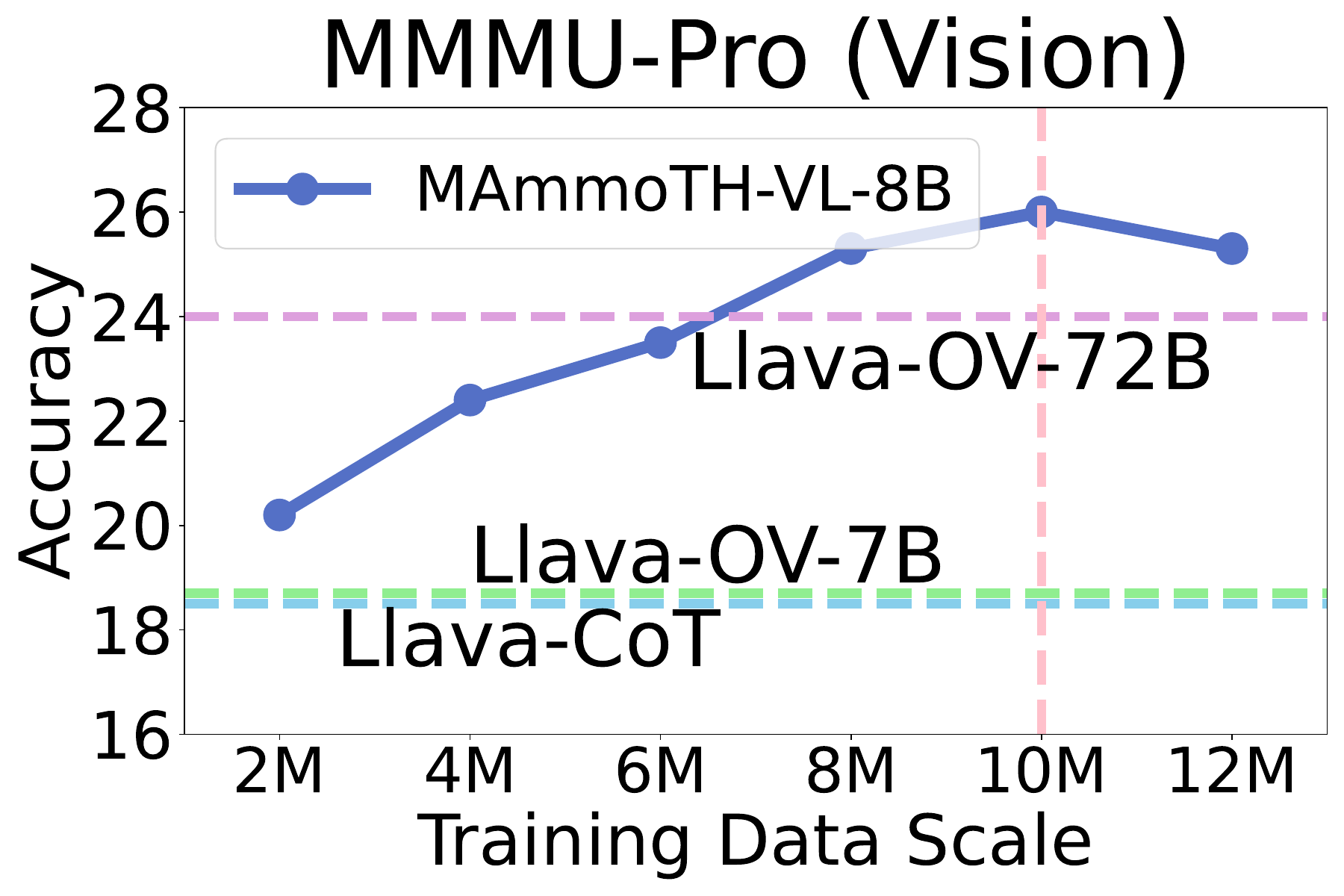}
    \includegraphics[width=0.24\linewidth]{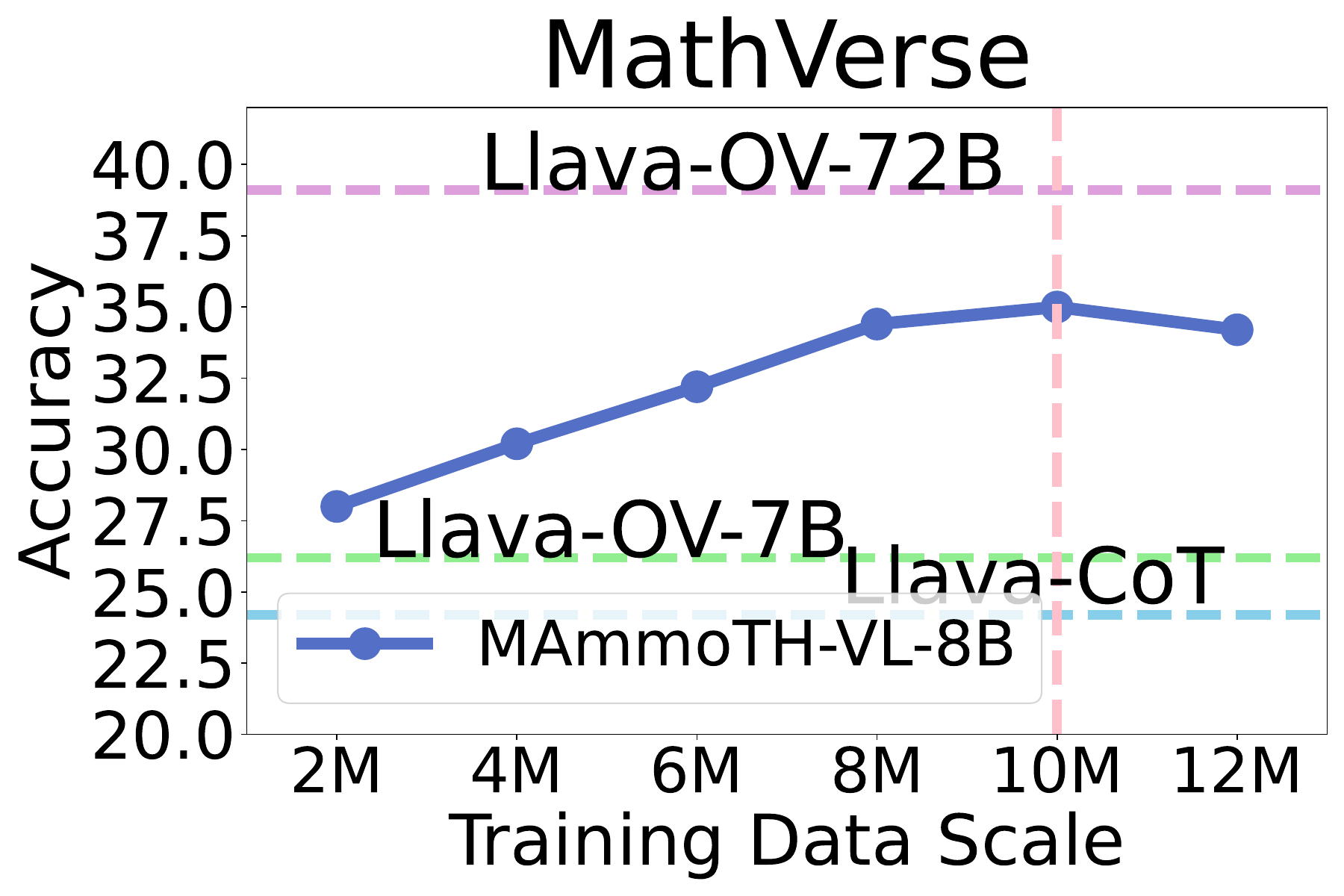}
    \includegraphics[width=0.24\linewidth]{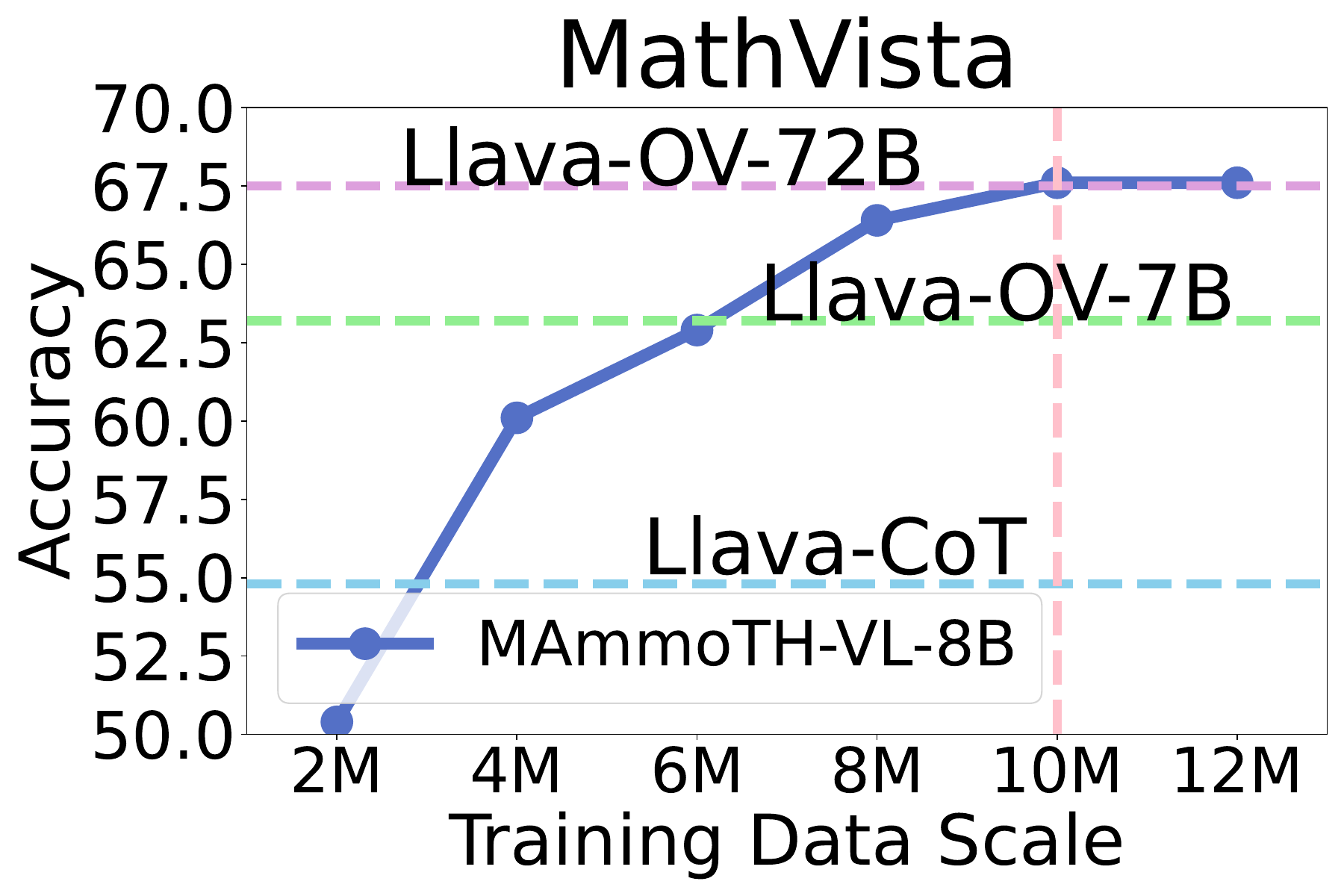}\\[2cm]
    \includegraphics[width=0.24\linewidth]{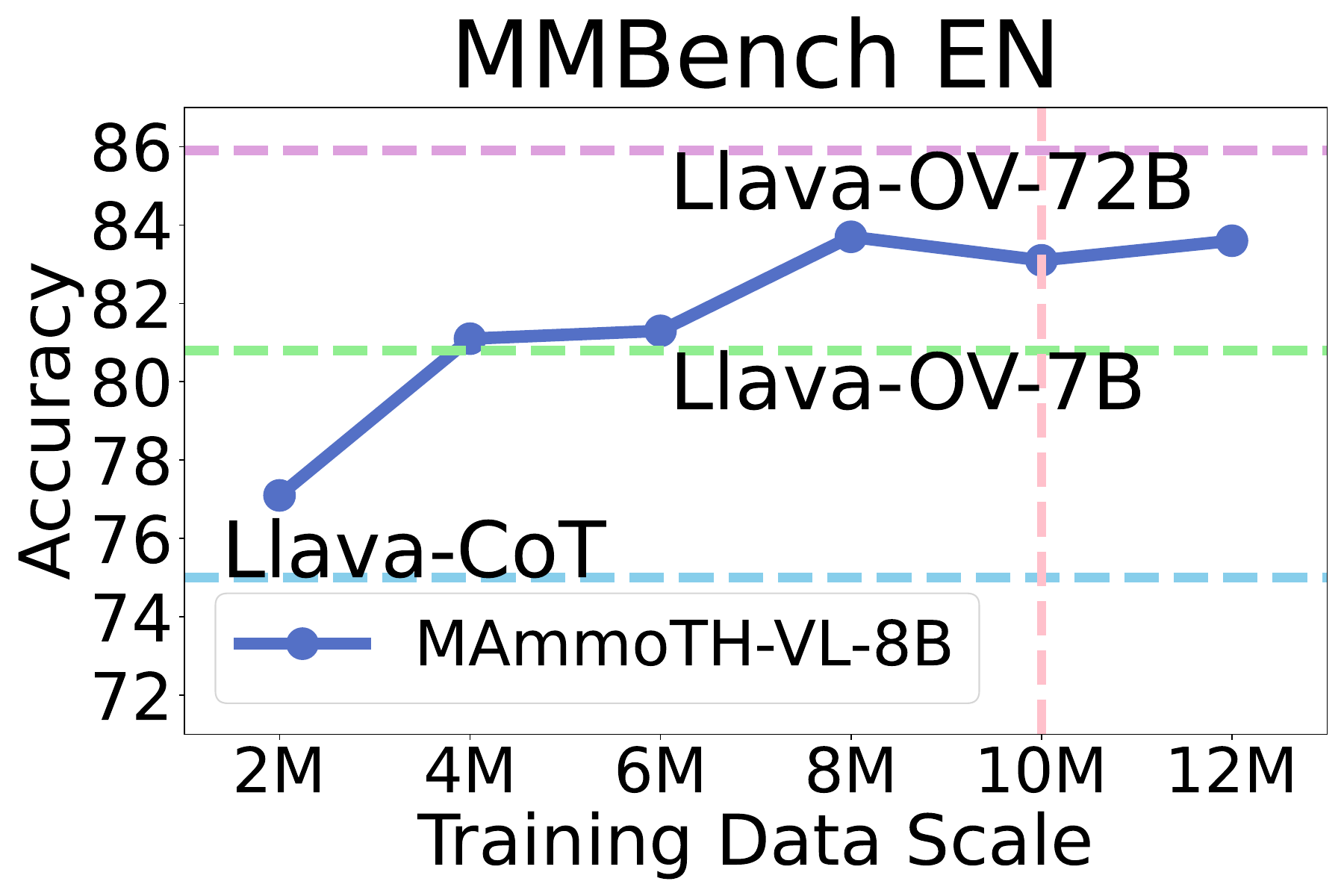}
    \includegraphics[width=0.24\linewidth]{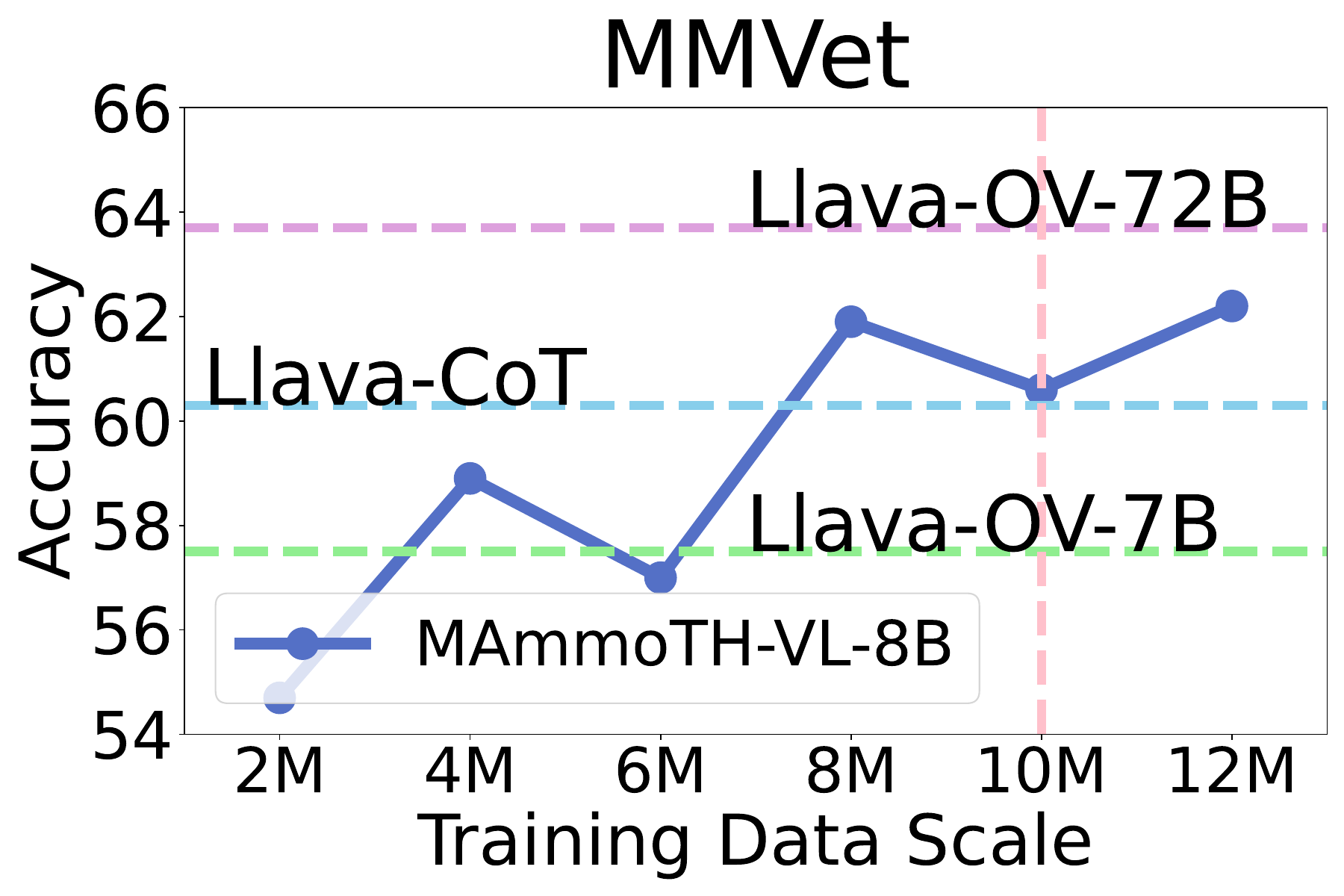}
    \includegraphics[width=0.24\linewidth]{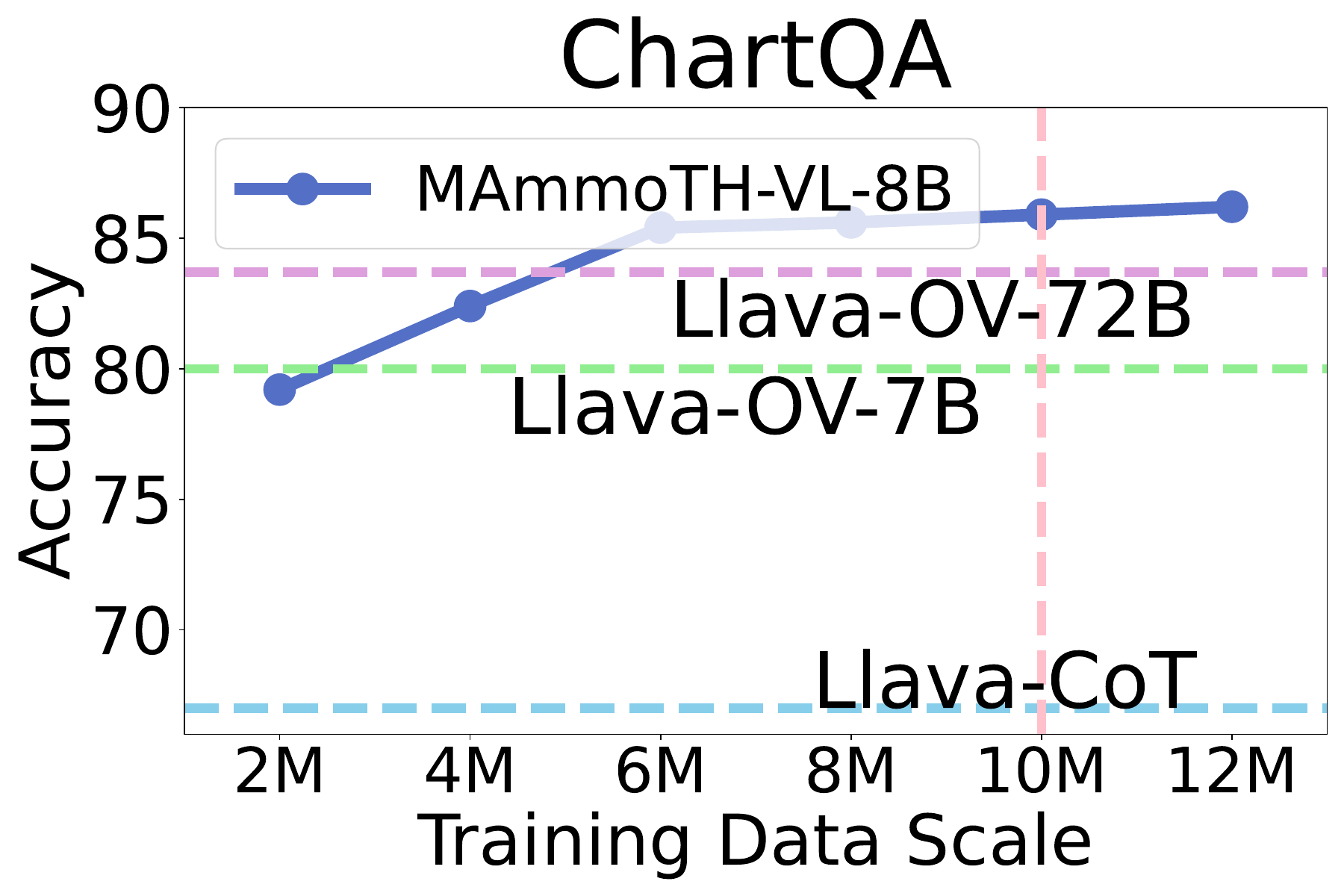}
    \includegraphics[width=0.24\linewidth]{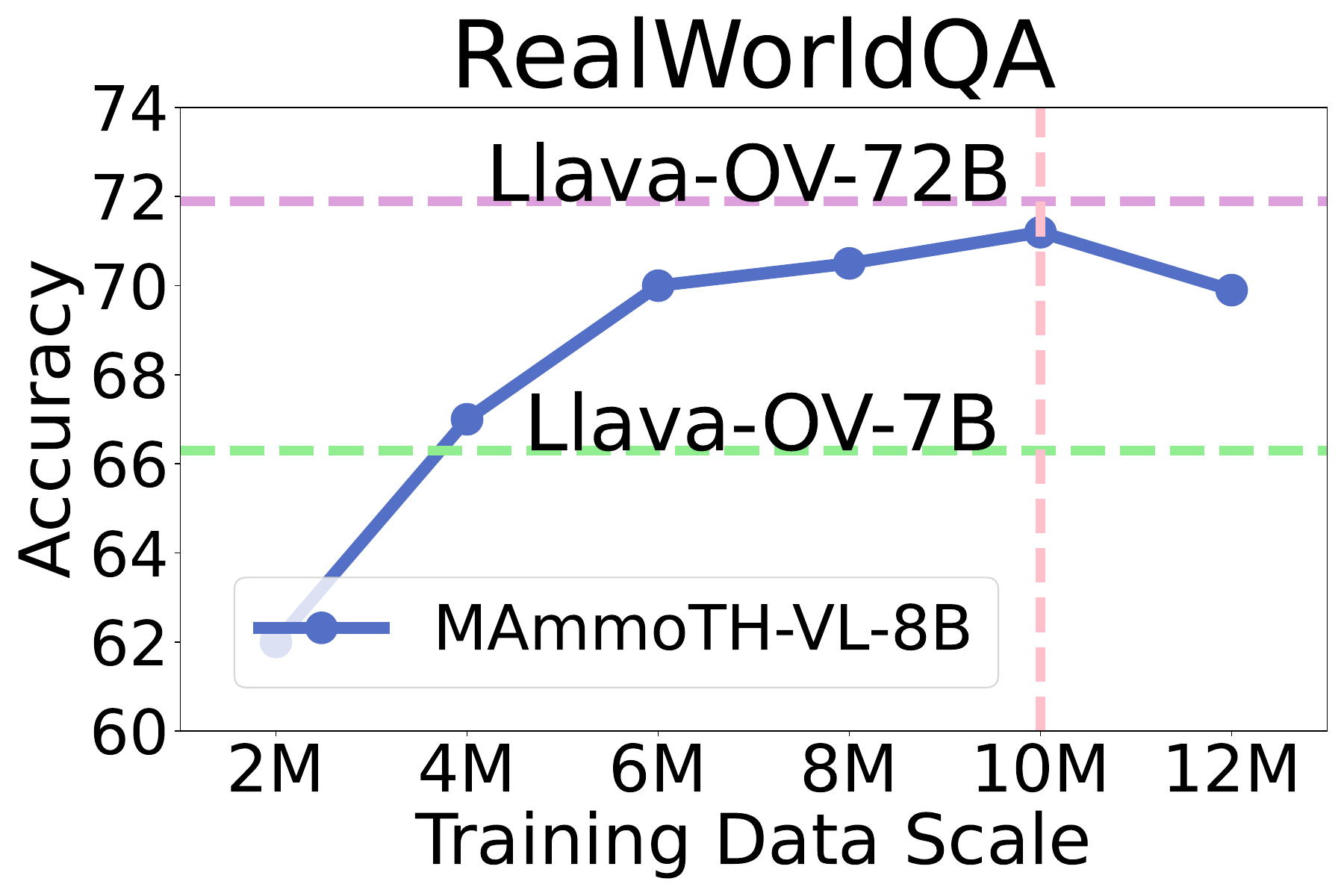}
    
\captionof{figure}{Scaling effects of \model-8B on eight multimodal evaluation datasets. A \textit{simple} rewriting approach using open models improves the quality of visual instruction data by eliciting chain-of-thought (CoT) reasoning. Training on this \textit{rewritten} data demonstrates significant performance gains through increased model scale. Llava-OneVision-7B\&72B~\citep{li2024llava} and Llava-CoT~\citep{xu2024llava} are included as references.}
    \label{fig:enter-label}
\end{strip}

\begin{abstract}
Open-source multimodal large language models (MLLMs) have shown significant potential in a broad range of multimodal tasks. However, their reasoning capabilities remain constrained by existing instruction-tuning datasets, which were predominately repurposed from academic datasets such as VQA, AI2D, and ChartQA. These datasets target simplistic tasks, and only provide phrase-level answers without any intermediate rationales.
To address these challenges, we introduce a \textit{scalable and cost-effective method} to construct a \textit{large-scale multimodal instruction-tuning dataset with rich intermediate rationales designed to elicit CoT reasoning}. Using only open models, we create a dataset containing 12M instruction-response pairs to cover diverse, reasoning-intensive tasks with detailed and faithful rationales. Experiments demonstrate that training MLLMs on this dataset significantly improves reasoning capabilities, achieving state-of-the-art performance on benchmarks such as MathVerse (+8.1\%), MMMU-Pro (+7\%), and MuirBench (+13.3\%). Additionally, the model demonstrates notable improvements of up to 4\% on non-reasoning-based benchmarks. Ablation studies further highlight the importance of key components, such as rewriting and self-filtering, in the dataset construction process.
\end{abstract}    
\begin{figure*}[!t]
    \centering
    \vspace{-5pt}
    \includegraphics[width=1.0\textwidth]{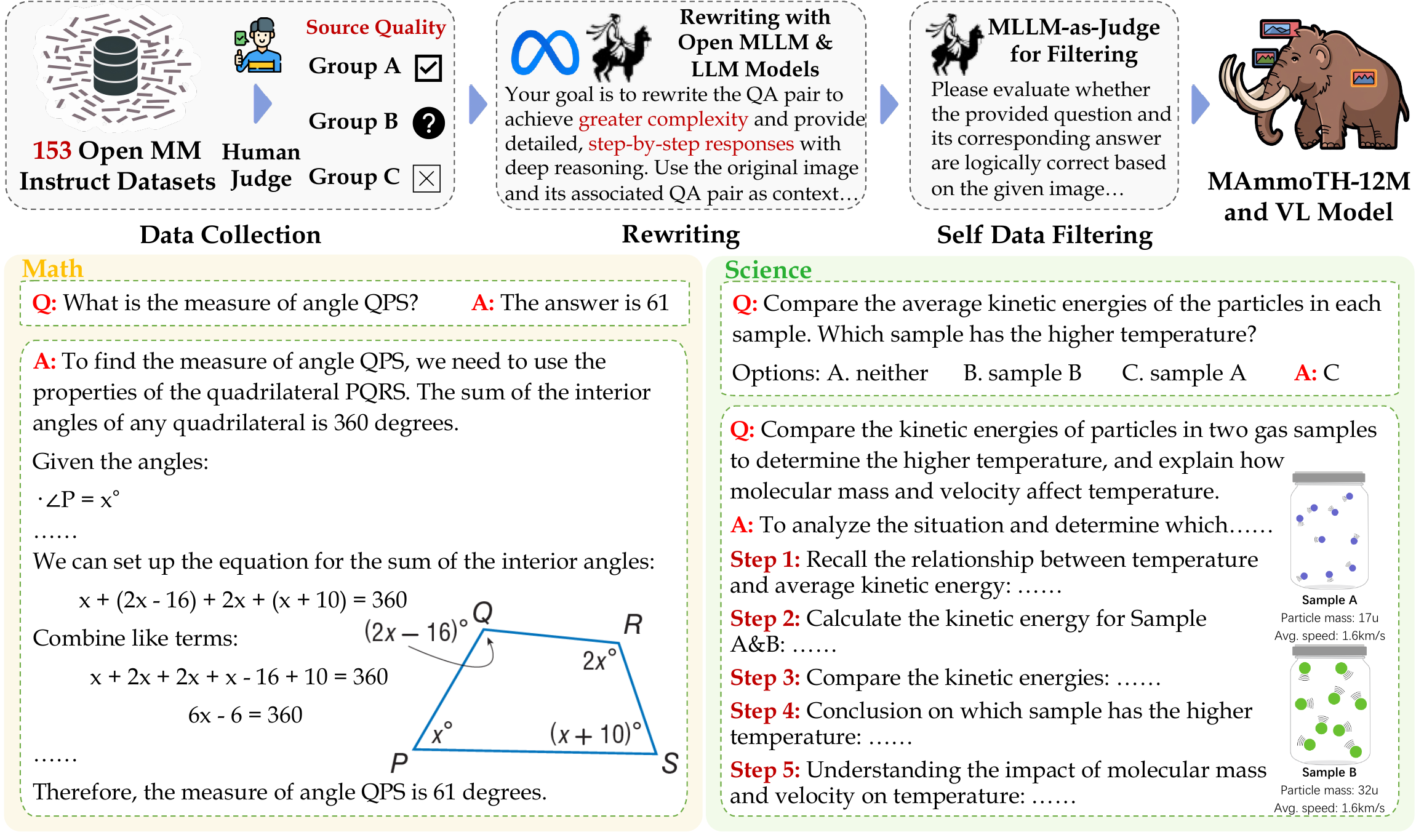}
\caption{Overview of our \textit{simple yet scalable} visual instruction data rewriting pipeline with three steps: manual data source collection, rewriting using MLLMs/LLMs, and filtering via the same MLLM as a judge. Examples below illustrate transformations in math and science categories, showcasing detailed, step-by-step responses.}
\vspace{-10pt}
    \label{fig:flow}
\end{figure*}

\section{Introduction}
Recent advances in open-source multimodal large language models (MLLMs), such as LLaVA~\citep{liu2024llava, li2024llava}, demonstrate the potential of integrating pre-trained visual encoders~\citep{radford2021learning, zhai2023sigmoid} with large language models (LLMs)~\citep{vicuna2023, bai2023qwen, touvron2023llama, young2024yi} to achieve robust multimodal capabilities across a range of tasks. While promising, these models often struggle with reasoning-heavy tasks~\citep{lu2024mathvistaevaluatingmathematicalreasoning, zhang2024mathversedoesmultimodalllm, yue2023mmmu, yue2024mmmu}, a limitation largely rooted in the shortcomings of existing instruction-tuning datasets.

Current multimodal instruction datasets predominantly target simplistic visual question answering (VQA) tasks, such as identifying objects in images (\textit{“What is shown in the image?”}) or generating straightforward captions (\textit{“Can you describe the image?”}). A key reason for this is that many of these datasets are repurposed from academic VQA datasets~\citep{antol2015vqa, textvqa}, which focus on phrase-based answers rather than nuanced reasoning~\citep{tong2024cambrian, bai2024survey}. Consequently, these datasets fail to elicit deliberate reasoning from multimodal models. The absence of deliberate reasoning not only limits interpretability but also hampers performance on tasks that demand contextual understanding.

Chain-of-Thought (CoT) reasoning has proven highly effective in addressing similar challenges in text-based LLMs~\citep{kojima2023largelanguagemodelszeroshot, wei2022chain}. By requiring step-by-step reasoning, CoT enhances interpretability and reasoning capabilities. Despite CoT's transformative potential, constructing datasets that elicit CoT reasoning remains a significant challenge~\citep{yue2024mammoth2}, particularly at the scale required to support robust multimodal learning. The creation of such datasets faces key obstacles: 1) ensuring instruction diversity and complexity, and 2) generating coherent responses with detailed rationales. Human-annotated CoT responses, while ideal, are prohibitively costly and lack scalability~\citep{xu2024vision, deitke2024molmo}. Furthermore, reliance on proprietary tools such as GPT-4~\citep{luo2024mmevol, zhang2024improve, xu2024llava} produces high-quality data but involves substantial costs and licensing issues for the open-source community, further exacerbating these challenges.

In this work, we address these challenges by introducing a simple, scalable, and cost-effective methodology for \textit{constructing instruction-tuning datasets at scale designed to elicit multimodal CoT reasoning}. As illustrated in \autoref{fig:flow}, we develop a high-quality dataset comprising 12 million entries using only open-weight LLMs~\citep{touvron2023llama} and MLLMs~\citep{chen2023internvl}, focusing on diverse, real-world tasks such as mathematical problem-solving, OCR, and domain-specific reasoning. The dataset is built through a three-step pipeline: (1) collecting and categorizing diverse image data into task-specific categories, (2) augmenting and rewriting tasks with CoT-style rationales using open models, and (3) rigorously filtering the data to ensure coherence and accuracy while minimizing hallucinations.

\begin{figure*}[tp]
\centering
\begin{minipage}{.24\textwidth}
\includegraphics[height=185pt]{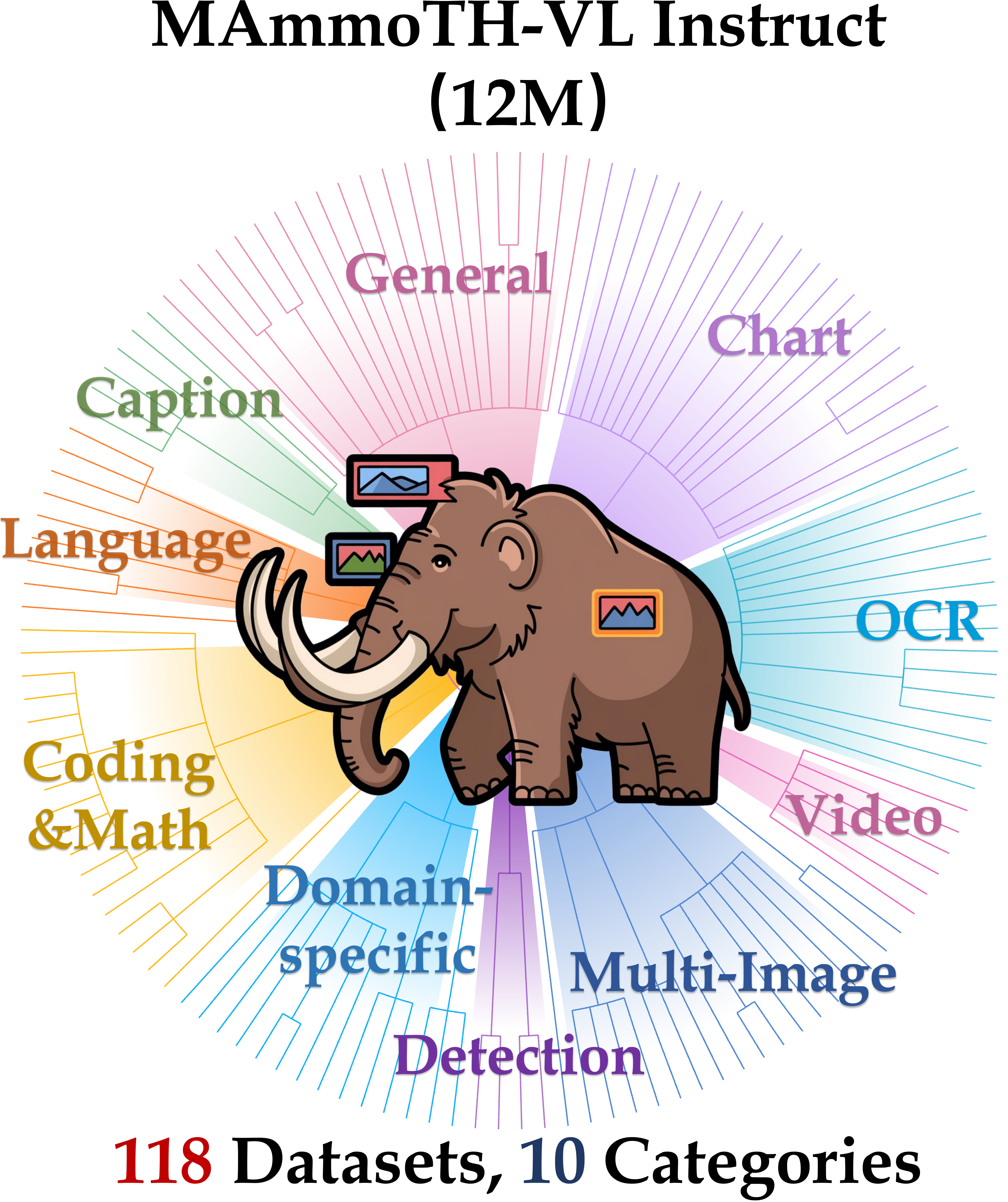}
\end{minipage}%
\begin{minipage}{.9\textwidth}
\centering
\renewcommand{\arraystretch}{1.1}
\setlength\tabcolsep{1pt}
\fontsize{6pt}{6.5pt}\selectfont
\begin{tabular}{lllll}

\makecell[l]{\cellcolor[RGB]{234,142,178} \textcolor{white}{General (15.4\%)}} &
\tikz[baseline=0.05em] \fill [color={rgb,255:red,234; green,142; blue,178}] (0,0) rectangle (0.75em,0.75em); ALLaVA  &
\tikz[baseline=0.05em] \fill [color={rgb,255:red,236; green,152; blue,185}] (0,0) rectangle (0.75em,0.75em); SVITCore  &
\tikz[baseline=0.05em] \fill [color={rgb,255:red,238; green,162; blue,192}] (0,0) rectangle (0.75em,0.75em); ALLaVA-zh  &

\tikz[baseline=0.05em] \fill [color={rgb,255:red,240; green,173; blue,199}] (0,0) rectangle (0.75em,0.75em); ShareGPT4V  \\
\tikz[baseline=0.05em] \fill [color={rgb,255:red,242; green,183; blue,206}] (0,0) rectangle (0.75em,0.75em); CLlava Instruct  &
\tikz[baseline=0.05em] \fill [color={rgb,255:red,243; green,193; blue,213}] (0,0) rectangle (0.75em,0.75em); idefics375k  & 
\tikz[baseline=0.05em] \fill [color={rgb,255:red,245; green,202; blue,220}] (0,0) rectangle (0.75em,0.75em); LVIS-InstructV4  &

\tikz[baseline=0.05em] \fill [color={rgb,255:red,240; green,173; blue,199}] (0,0) rectangle (0.75em,0.75em); WildVision Chat  &
\tikz[baseline=0.05em] \fill [color={rgb,255:red,242; green,183; blue,206}] (0,0) rectangle (0.75em,0.75em); GQA  \\
\tikz[baseline=0.05em] \fill [color={rgb,255:red,243; green,193; blue,213}] (0,0) rectangle (0.75em,0.75em); AlfWorld  & 
\tikz[baseline=0.05em] \fill [color={rgb,255:red,245; green,202; blue,220}] (0,0) rectangle (0.75em,0.75em); IDK  &

\tikz[baseline=0.05em] \fill [color={rgb,255:red,240; green,173; blue,199}] (0,0) rectangle (0.75em,0.75em); GPT4V77  &
\tikz[baseline=0.05em] \fill [color={rgb,255:red,242; green,183; blue,206}] (0,0) rectangle (0.75em,0.75em); Laion GPT4V  &
\tikz[baseline=0.05em] \fill [color={rgb,255:red,243; green,193; blue,213}] (0,0) rectangle (0.75em,0.75em); Sherlock  \\
\tikz[baseline=0.05em] \fill [color={rgb,255:red,245; green,202; blue,220}] (0,0) rectangle (0.75em,0.75em); Irv-Normal  &

\tikz[baseline=0.05em] \fill [color={rgb,255:red,247; green,213; blue,227}] (0,0) rectangle (0.75em,0.75em); LLaVA-zh  &
\tikz[baseline=0.05em] \fill [color={rgb,255:red,249; green,223; blue,233}] (0,0) rectangle (0.75em,0.75em); SVITCore  &
\tikz[baseline=0.05em] \fill [color={rgb,255:red,251; green,234; blue,240}] (0,0) rectangle (0.75em,0.75em); Cambrian (Filter)  &
\tikz[baseline=0.05em] \fill [color={rgb,255:red,253; green,244; blue,247}] (0,0) rectangle (0.75em,0.75em); Visual7W  \\

\makecell[l]{\cellcolor[RGB]{193,149,245} \textcolor{white}{Chart (15.4\%)}} &
\tikz[baseline=0.05em] \fill [color={rgb,255:red,193; green,149; blue,245}] (0,0) rectangle (0.75em,0.75em); mPLUG-DocOwlchart  &
\tikz[baseline=0.05em] \fill [color={rgb,255:red,199; green,159; blue,246}] (0,0) rectangle (0.75em,0.75em); Ureader Chart  &
\tikz[baseline=0.05em] \fill [color={rgb,255:red,204; green,168; blue,247}] (0,0) rectangle (0.75em,0.75em); Ureader QA  &

\tikz[baseline=0.05em] \fill [color={rgb,255:red,210; green,178; blue,248}] (0,0) rectangle (0.75em,0.75em); DVQA \\
\tikz[baseline=0.05em] \fill [color={rgb,255:red,215; green,187; blue,249}] (0,0) rectangle (0.75em,0.75em); ArXiv-Chart-GPT4o  &
\tikz[baseline=0.05em] \fill [color={rgb,255:red,221; green,197; blue,250}] (0,0) rectangle (0.75em,0.75em); PlotQA  &
\tikz[baseline=0.05em] \fill [color={rgb,255:red,226; green,206; blue,250}] (0,0) rectangle (0.75em,0.75em); ArxivQA  &

\tikz[baseline=0.05em] \fill [color={rgb,255:red,215; green,187; blue,249}] (0,0) rectangle (0.75em,0.75em); InfographicVQA  &
\tikz[baseline=0.05em] \fill [color={rgb,255:red,221; green,197; blue,250}] (0,0) rectangle (0.75em,0.75em); Robut-WTQ  \\
\tikz[baseline=0.05em] \fill [color={rgb,255:red,226; green,206; blue,250}] (0,0) rectangle (0.75em,0.75em); Robut-SQA  &
\tikz[baseline=0.05em] \fill [color={rgb,255:red,210; green,178; blue,248}] (0,0) rectangle (0.75em,0.75em); Hitab  &

\tikz[baseline=0.05em] \fill [color={rgb,255:red,215; green,187; blue,249}] (0,0) rectangle (0.75em,0.75em); TAT-QA  &
\tikz[baseline=0.05em] \fill [color={rgb,255:red,221; green,197; blue,250}] (0,0) rectangle (0.75em,0.75em); FinQA  &
\tikz[baseline=0.05em] \fill [color={rgb,255:red,232; green,216; blue,251}] (0,0) rectangle (0.75em,0.75em); Vistext \\
\tikz[baseline=0.05em] \fill [color={rgb,255:red,238; green,225; blue,252}] (0,0) rectangle (0.75em,0.75em); ChartQA  &
\tikz[baseline=0.05em] \fill [color={rgb,255:red,232; green,216; blue,251}] (0,0) rectangle (0.75em,0.75em); Robut-WikiSQL &
\tikz[baseline=0.05em] \fill [color={rgb,255:red,238; green,225; blue,252}] (0,0) rectangle (0.75em,0.75em); Ureader KG  &
\tikz[baseline=0.05em] \fill [color={rgb,255:red,243; green,235; blue,253}] (0,0) rectangle (0.75em,0.75em); Chart2Text  &
\tikz[baseline=0.05em] \fill [color={rgb,255:red,249; green,244; blue,254}] (0,0) rectangle (0.75em,0.75em); Irv-Chart \\

\makecell[l]{\cellcolor[RGB]{38,172,204} \textcolor{white}{OCR (13.7\%)}} & 
\tikz[baseline=0.05em] \fill [color={rgb,255:red,38; green,172; blue,204}] (0,0) rectangle (0.75em,0.75em); MultiUI  &
\tikz[baseline=0.05em] \fill [color={rgb,255:red,51; green,177; blue,207}] (0,0) rectangle (0.75em,0.75em); OCRVQA  &
\tikz[baseline=0.05em] \fill [color={rgb,255:red,64; green,182; blue,210}] (0,0) rectangle (0.75em,0.75em); ScreenQA  &
\tikz[baseline=0.05em] \fill [color={rgb,255:red,77; green,187; blue,213}] (0,0) rectangle (0.75em,0.75em); TextVQA  \\
\tikz[baseline=0.05em] \fill [color={rgb,255:red,90; green,192; blue,216}] (0,0) rectangle (0.75em,0.75em); TextOCR  &
\tikz[baseline=0.05em] \fill [color={rgb,255:red,103; green,197; blue,219}] (0,0) rectangle (0.75em,0.75em); LLaVAR GPT4  &
\tikz[baseline=0.05em] \fill [color={rgb,255:red,116; green,202; blue,222}] (0,0) rectangle (0.75em,0.75em); ReCTs  &
\tikz[baseline=0.05em] \fill [color={rgb,255:red,77; green,187; blue,213}] (0,0) rectangle (0.75em,0.75em); Chrome-Writing  &
\tikz[baseline=0.05em] \fill [color={rgb,255:red,90; green,192; blue,216}] (0,0) rectangle (0.75em,0.75em); IAM  \\
\tikz[baseline=0.05em] \fill [color={rgb,255:red,116; green,202; blue,222}] (0,0) rectangle (0.75em,0.75em); UreaderOCR  &
\tikz[baseline=0.05em] \fill [color={rgb,255:red,129; green,207; blue,224}] (0,0) rectangle (0.75em,0.75em); ST-VQA  &
\tikz[baseline=0.05em] \fill [color={rgb,255:red,142; green,212; blue,227}] (0,0) rectangle (0.75em,0.75em); DocVQA  &
\tikz[baseline=0.05em] \fill [color={rgb,255:red,155; green,217; blue,230}] (0,0) rectangle (0.75em,0.75em); RenderedText &
\tikz[baseline=0.05em] \fill [color={rgb,255:red,168; green,222; blue,233}] (0,0) rectangle (0.75em,0.75em); VisualMRC \\

\makecell[l]{\cellcolor[RGB]{46,151,78} \textcolor{white}{Caption (10.9\%)}} &
\tikz[baseline=0.05em] \fill [color={rgb,255:red,46; green,151; blue,78}] (0,0) rectangle (0.75em,0.75em); ShareGPT4v  &
\tikz[baseline=0.05em] \fill [color={rgb,255:red,66; green,161; blue,92}] (0,0) rectangle (0.75em,0.75em); ShareGPT4o  &
\tikz[baseline=0.05em] \fill [color={rgb,255:red,86; green,171; blue,106}] (0,0) rectangle (0.75em,0.75em); Sharegpt4v (SAM)  &
\tikz[baseline=0.05em] \fill [color={rgb,255:red,106; green,181; blue,120}] (0,0) rectangle (0.75em,0.75em); Infographic  \\
\tikz[baseline=0.05em] \fill [color={rgb,255:red,126; green,191; blue,134}] (0,0) rectangle (0.75em,0.75em); Sharegpt4v (COCO)  &
\tikz[baseline=0.05em] \fill [color={rgb,255:red,146; green,201; blue,148}] (0,0) rectangle (0.75em,0.75em); Sharegpt4v (LLAVA) & & & \\

\makecell[l]{\cellcolor[RGB]{226,85,8} \textcolor{white}{Language (16\%)}} &
\tikz[baseline=0.05em] \fill [color={rgb,255:red,226; green,85; blue,8}] (0,0) rectangle (0.75em,0.75em); Orca  &
\tikz[baseline=0.05em] \fill [color={rgb,255:red,236; green,105; blue,28}] (0,0) rectangle (0.75em,0.75em); NuminaMath &
\tikz[baseline=0.05em] \fill [color={rgb,255:red,246; green,125; blue,48}] (0,0) rectangle (0.75em,0.75em); MathInstruct  &
\tikz[baseline=0.05em] \fill [color={rgb,255:red,255; green,145; blue,68}] (0,0) rectangle (0.75em,0.75em); Orca Math  \\
\tikz[baseline=0.05em] \fill [color={rgb,255:red,255; green,165; blue,88}] (0,0) rectangle (0.75em,0.75em); Magpie Pro(L3 MT)  &
\tikz[baseline=0.05em] \fill [color={rgb,255:red,255; green,185; blue,108}] (0,0) rectangle (0.75em,0.75em); Magpie Pro(L3 ST)  &
\tikz[baseline=0.05em] \fill [color={rgb,255:red,255; green,205; blue,128}] (0,0) rectangle (0.75em,0.75em); Others & \\

\makecell[l]{\cellcolor[RGB]{252,188,33} \textcolor{white}{Code/Math (8.3\%)}} &
\tikz[baseline=0.05em] \fill [color={rgb,255:red,252; green,188; blue,33}] (0,0) rectangle (0.75em,0.75em); MAVIS Geo  & 
\tikz[baseline=0.05em] \fill [color={rgb,255:red,252; green,194; blue,53}] (0,0) rectangle (0.75em,0.75em); MAVIS Metagen & 
\tikz[baseline=0.05em] \fill [color={rgb,255:red,252; green,200; blue,73}] (0,0) rectangle (0.75em,0.75em); Geometry3K &

\tikz[baseline=0.05em] \fill [color={rgb,255:red,252; green,206; blue,93}] (0,0) rectangle (0.75em,0.75em); GeomVerse  \\
\tikz[baseline=0.05em] \fill [color={rgb,255:red,253; green,212; blue,113}] (0,0) rectangle (0.75em,0.75em); Super-CLEVR  &
\tikz[baseline=0.05em] \fill [color={rgb,255:red,253; green,218; blue,133}] (0,0) rectangle (0.75em,0.75em); TabMWP  & 
\tikz[baseline=0.05em] \fill [color={rgb,255:red,254; green,224; blue,153}] (0,0) rectangle (0.75em,0.75em); VizWiz  &

\tikz[baseline=0.05em] \fill [color={rgb,255:red,252; green,206; blue,93}] (0,0) rectangle (0.75em,0.75em); Geo170K  &
\tikz[baseline=0.05em] \fill [color={rgb,255:red,253; green,212; blue,113}] (0,0) rectangle (0.75em,0.75em); MathVision  \\
\tikz[baseline=0.05em] \fill [color={rgb,255:red,253; green,218; blue,133}] (0,0) rectangle (0.75em,0.75em); GEOS  & 
\tikz[baseline=0.05em] \fill [color={rgb,255:red,254; green,224; blue,153}] (0,0) rectangle (0.75em,0.75em); GeoQA+  &

\tikz[baseline=0.05em] \fill [color={rgb,255:red,252; green,206; blue,93}] (0,0) rectangle (0.75em,0.75em); IconQA(Math)  &
\tikz[baseline=0.05em] \fill [color={rgb,255:red,253; green,212; blue,113}] (0,0) rectangle (0.75em,0.75em); PMC-VQA  &
\tikz[baseline=0.05em] \fill [color={rgb,255:red,253; green,218; blue,133}] (0,0) rectangle (0.75em,0.75em); UniGeo  \\
\tikz[baseline=0.05em] \fill [color={rgb,255:red,254; green,230; blue,173}] (0,0) rectangle (0.75em,0.75em); CLEVR-Math &

\tikz[baseline=0.05em] \fill [color={rgb,255:red,254; green,236; blue,193}] (0,0) rectangle (0.75em,0.75em); MapQA  &
\tikz[baseline=0.05em] \fill [color={rgb,255:red,254; green,242; blue,213}] (0,0) rectangle (0.75em,0.75em); RAVEN(M)  &
\tikz[baseline=0.05em] \fill [color={rgb,255:red,254; green,248; blue,232}] (0,0) rectangle (0.75em,0.75em); Design2Code &  \\

\makecell[l]{\cellcolor[RGB]{23,177,250} \textcolor{white}{Domain-specific (8.9\%)}} &
\tikz[baseline=0.05em] \fill [color={rgb,255:red,23; green,177; blue,250}] (0,0) rectangle (0.75em,0.75em); WIT  &
\tikz[baseline=0.05em] \fill [color={rgb,255:red,44; green,184; blue,250}] (0,0) rectangle (0.75em,0.75em); M3IT+FLAN  &
\tikz[baseline=0.05em] \fill [color={rgb,255:red,65; green,191; blue,251}] (0,0) rectangle (0.75em,0.75em); ScienceQA(Nona) &

\tikz[baseline=0.05em] \fill [color={rgb,255:red,86; green,198; blue,251}] (0,0) rectangle (0.75em,0.75em); Vision Flan  \\
\tikz[baseline=0.05em] \fill [color={rgb,255:red,107; green,205; blue,252}] (0,0) rectangle (0.75em,0.75em); PathVQA  &
\tikz[baseline=0.05em] \fill [color={rgb,255:red,127; green,212; blue,252}] (0,0) rectangle (0.75em,0.75em); TQA &
\tikz[baseline=0.05em] \fill [color={rgb,255:red,148; green,219; blue,253}] (0,0) rectangle (0.75em,0.75em); A-OKVQA  &

\tikz[baseline=0.05em] \fill [color={rgb,255:red,169; green,226; blue,253}] (0,0) rectangle (0.75em,0.75em); WebSight  &
\tikz[baseline=0.05em] \fill [color={rgb,255:red,190; green,233; blue,254}] (0,0) rectangle (0.75em,0.75em); ViQuAE \\
\tikz[baseline=0.05em] \fill [color={rgb,255:red,211; green,240; blue,254}] (0,0) rectangle (0.75em,0.75em); ShareGPT4V(Knowledge)  &
\tikz[baseline=0.05em] \fill [color={rgb,255:red,232; green,247; blue,255}] (0,0) rectangle (0.75em,0.75em); AI2D(4V) & & & \\

\makecell[l]{\cellcolor[RGB]{122,28,172} \textcolor{white}{Detection (3.2\%)}} & 
\tikz[baseline=0.05em] \fill [color={rgb,255:red,122; green,28; blue,172}] (0,0) rectangle (0.75em,0.75em); CLEVR  & 
\tikz[baseline=0.05em] \fill [color={rgb,255:red,162; green,96; blue,197}] (0,0) rectangle (0.75em,0.75em); VisualGenome  & 
\tikz[baseline=0.05em] \fill [color={rgb,255:red,202; green,164; blue,222}] (0,0) rectangle (0.75em,0.75em); TallyQA  &
\tikz[baseline=0.05em] \fill [color={rgb,255:red,202; green,194; blue,222}] (0,0) rectangle (0.75em,0.75em); VSR \\

\makecell[l]{\cellcolor[RGB]{85,138,214} \textcolor{white}{Multi-Image (5.8\%)}} &
\tikz[baseline=0.05em] \fill [color={rgb,255:red,85; green,138; blue,214}] (0,0) rectangle (0.75em,0.75em); NLVR2 &
\tikz[baseline=0.05em] \fill [color={rgb,255:red,100; green,149; blue,218}] (0,0) rectangle (0.75em,0.75em); Mimic CGD &
\tikz[baseline=0.05em] \fill [color={rgb,255:red,116; green,159; blue,221}] (0,0) rectangle (0.75em,0.75em); Coinstruct &

\tikz[baseline=0.05em] \fill [color={rgb,255:red,131; green,170; blue,225}] (0,0) rectangle (0.75em,0.75em); HQ-Edit \\
\tikz[baseline=0.05em] \fill [color={rgb,255:red,146; green,180; blue,229}] (0,0) rectangle (0.75em,0.75em); Raven &
\tikz[baseline=0.05em] \fill [color={rgb,255:red,162; green,191; blue,232}] (0,0) rectangle (0.75em,0.75em); IconQA &
\tikz[baseline=0.05em] \fill [color={rgb,255:red,177; green,201; blue,236}] (0,0) rectangle (0.75em,0.75em); VIST &

\tikz[baseline=0.05em] \fill [color={rgb,255:red,192; green,212; blue,240}] (0,0) rectangle (0.75em,0.75em); Contrast-Caption &
\tikz[baseline=0.05em] \fill [color={rgb,255:red,207; green,222; blue,244}] (0,0) rectangle (0.75em,0.75em); FlintstonesSV \\
\tikz[baseline=0.05em] \fill [color={rgb,255:red,223; green,233; blue,247}] (0,0) rectangle (0.75em,0.75em); PororoSV &
\tikz[baseline=0.05em] \fill [color={rgb,255:red,238; green,243; blue,251}] (0,0) rectangle (0.75em,0.75em); Others & & & \\

\makecell[l]{\cellcolor[RGB]{235,107,185} \textcolor{white}{Video (2.5\%)}} &
\tikz[baseline=0.05em] \fill [color={rgb,255:red,235; green,107; blue,185}] (0,0) rectangle (0.75em,0.75em); L-Video &
\tikz[baseline=0.05em] \fill [color={rgb,255:red,237; green,122; blue,192}] (0,0) rectangle (0.75em,0.75em); M4 Instruct Video &
\tikz[baseline=0.05em] \fill [color={rgb,255:red,239; green,137; blue,199}] (0,0) rectangle (0.75em,0.75em); L-Video-ActivityNetQA &

\tikz[baseline=0.05em] \fill [color={rgb,255:red,241; green,151; blue,206}] (0,0) rectangle (0.75em,0.75em); L-Hound \\
\tikz[baseline=0.05em] \fill [color={rgb,255:red,243; green,166; blue,213}] (0,0) rectangle (0.75em,0.75em); L-Video-NeXT-QA &
\tikz[baseline=0.05em] \fill [color={rgb,255:red,247; green,196; blue,227}] (0,0) rectangle (0.75em,0.75em); VideoChatGPT &

\tikz[baseline=0.05em] \fill [color={rgb,255:red,249; green,211; blue,234}] (0,0) rectangle (0.75em,0.75em); Video-MME &
\tikz[baseline=0.05em] \fill [color={rgb,255:red,251; green,225; blue,241}] (0,0) rectangle (0.75em,0.75em); L-Video-PerceptionTest &
\tikz[baseline=0.05em] \fill [color={rgb,255:red,253; green,240; blue,248}] (0,0) rectangle (0.75em,0.75em); EgoSchema  \\

\end{tabular}
\end{minipage}

\captionsetup{font={small}}
\caption{The data distribution of \textbf{\model-Instruct (12M).} Left: Category distribution. Right: Details of data sources.
}
\label{fig:single_image_all}
\end{figure*}

Our experimental results demonstrate the efficacy of this approach. By training an MLLM model, \model-8B, based on the LLaVA-OneVision architecture~\citep{li2024llava}, on our curated dataset, we achieve substantial improvements across multiple benchmarks, particularly in tasks requiring intricate reasoning and alignment between text and images. For example, \model-8B achieves an 8.1\% improvement on MathVerse~\citep{zhang2024mathversedoesmultimodalllm}, 7\% on MMMU-Pro~\citep{yue2024mmmu}, and 13.3\% on MuirBench~\cite{wang2024muirbenchcomprehensivebenchmarkrobust} compared to the open SoTA models. On other non-reasoning-based benchmarks, our model also delivers notable improvements. Ablation studies further reveal critical insights, including the importance of \textit{self-filtering} for mitigating hallucinations, the effectiveness of mixing rewritten and original data for enhancing task diversity, and the significant impact of scaling both training data size and rewrite model capacity on performance. In summary, our open-source, cost-effective methodology provides a scalable solution for the community to build high-quality rationale-enriched multimodal datasets.

    

\section{Method} \label{sec:Method}

While previous efforts have highlighted the potential of visual instruction tuning, many rely on resource-intensive methods such as human annotations or proprietary models. These approaches limit scalability and accessibility, particularly in open-source contexts. To address these challenges, we introduce a \textit{simple, scalable, and cost-effective data generation pipeline} that produces 12 million high-quality samples. Our pipeline involves three key steps: (1) open-source data collection and categorization, (2) task-specific data augmentation and rewriting using open models, and (3) quality filtering to remove hallucinated or irrelevant content.

\subsection{Dataset Collection and Categorization} \label{sec:data_collect}

To build a comprehensive and diverse multimodal dataset, we systematically collect and categorize data spanning a wide range of real-world tasks and scenarios. This foundational step ensures the dataset’s robustness, enabling MLLMs to acquire general capabilities suitable for a variety of applications.

\paragraph{Data Collection.} To achieve both scale and diversity while maintaining accessibility for open-source initiatives, we sourced data from 153 publicly available multimodal instruction datasets, expanding upon prior collection efforts~\citep{li2024llava,tong2024cambrian}. The raw data includes image-text pairs covering a broad spectrum of use cases such as OCR, charts, captioning, and domain-specific images (e.g., medical).

Based on MLLM training paradigms and common downstream tasks, we reorganized the training data into ten major categories: General, OCR, Chart, Caption, Domain-specific, Code\&Math, Language, Detection, Multi-Image, and Video, as illustrated in~\autoref{fig:single_image_all}. This structured categorization is critical for enhancing data quality during the rewriting stage, allowing us to apply tailored strategies to different scenarios. Moreover, this categorization facilitates fine-tuning models for specific applications, providing a valuable resource for the broader community.

\paragraph{Data Source Screening.}
The quality of the collected open-source datasets varies significantly, requiring careful assessment to ensure baseline quality and adequate coverage for task-specific rewriting. To address this, we conducted an initial manual screening of the 153 data sources. This screening process aimed to identify datasets containing rich textual information with strong modality correlations, which are essential for producing high-quality multimodal data. For each dataset, we randomly sampled 1,000 data points and performed a rapid evaluation to gauge overall quality. Based on this assessment, we categorized the datasets into three groups:

\begin{itemize}[leftmargin=*] 
\item \textbf{\textit{Group A (58 datasets).}} These datasets contain detailed, informative, and accurate responses that are well-structured and aligned with the desired task-oriented structure. Data from this group were retained in their original form as no further elaboration or rewriting was necessary.

\item \textbf{\textit{Group B (60 datasets).}} These datasets include responses that are brief or incomplete but have the potential for meaningful enhancement. To enrich their quality and utility, we rewrote the data into task-specific Q\&A pairs, simulating real-world applications by adding depth, context, and reasoning. The rewriting details will be discussed in the next section.

\item \textbf{\textit{Group C (35 datasets).}} These datasets contain responses that are overly brief, vague, or lacking in depth, making them unsuitable for meaningful improvement. These datasets often lack sufficient context, reasoning, or actionable content. We directly remove these datasets.

\end{itemize}

In addition, all images were standardized according to verified criteria~\citep{li2024llava}. This included resizing images to ensure dimensions fell between 224 and 4,096 pixels and adjusting extreme aspect ratios (e.g., greater than 7:1 or less than 1:7) to produce consistent formatting. Further details are illustrated in~\autoref{fig:group_abc_a}, ~\autoref{fig:group_abc_b} and~\autoref{fig:group_abc_c}.

\subsection{Instruction Data Rewriting}

To enhance the quality of instruction data in Group B, we implemented a task-aware rewriting process aimed at addressing two key shortcomings: 1) a lack of detailed intermediate rationales, as many datasets originate from academic visual question answering contexts where responses are typically concise (e.g., a single word or phrase); and 2) limited coverage of real-world tasks such as reasoning, data analysis, code generation, debugging, and other practical applications.

Our approach involves transforming the original multimodal data into \textit{diverse instruction-response pairs enriched with detailed rationales}. This process broadens the data’s coverage, ensuring it encompasses a wide range of real-world applications. Importantly, we leverage the original instruction-response associated with images, even when brief, to interpret the visual content and generate coherent, task-specific instructions.

To guide this transformation, we designed customized prompts tailored to each data category. These prompts are crafted to generate responses that align with real-world applications while encouraging critical thinking and reasoning. Each prompt undergoes an iterative refinement process: initial drafts are tested on sampled data points within a specific category, and the outputs are manually analyzed to identify areas for improvement. Once finalized, the optimized prompts are applied for rewriting.  The full prompts are provided in Appendix ~\autoref{app:prompts_fot_rewrite}.

For caption-based data, we employ a text-only model (Llama-3-70B-Instruct~\citep{grattafiori2024llama3herdmodels}) to generate task-oriented Q\&A pairs. Captions typically contain rich textual information, and we observe that text-only models are better suited for creating diverse and complex instructions compared to multimodal models. For all other types of data, we utilize a multimodal model (InternVL2-Llama3-76B~\citep{chen2023internvl}) to ensure strong alignment between visual content and generated instructions, effectively leveraging both text and images for coherent outputs.

This rewriting process results in a rich dataset of instruction-response pairs, characterized by detailed rationales and diverse real-world scenarios, all while maintaining high relevance to the visual content.


\begin{figure}[t]
    \centering
    \includegraphics[width=0.235\textwidth]{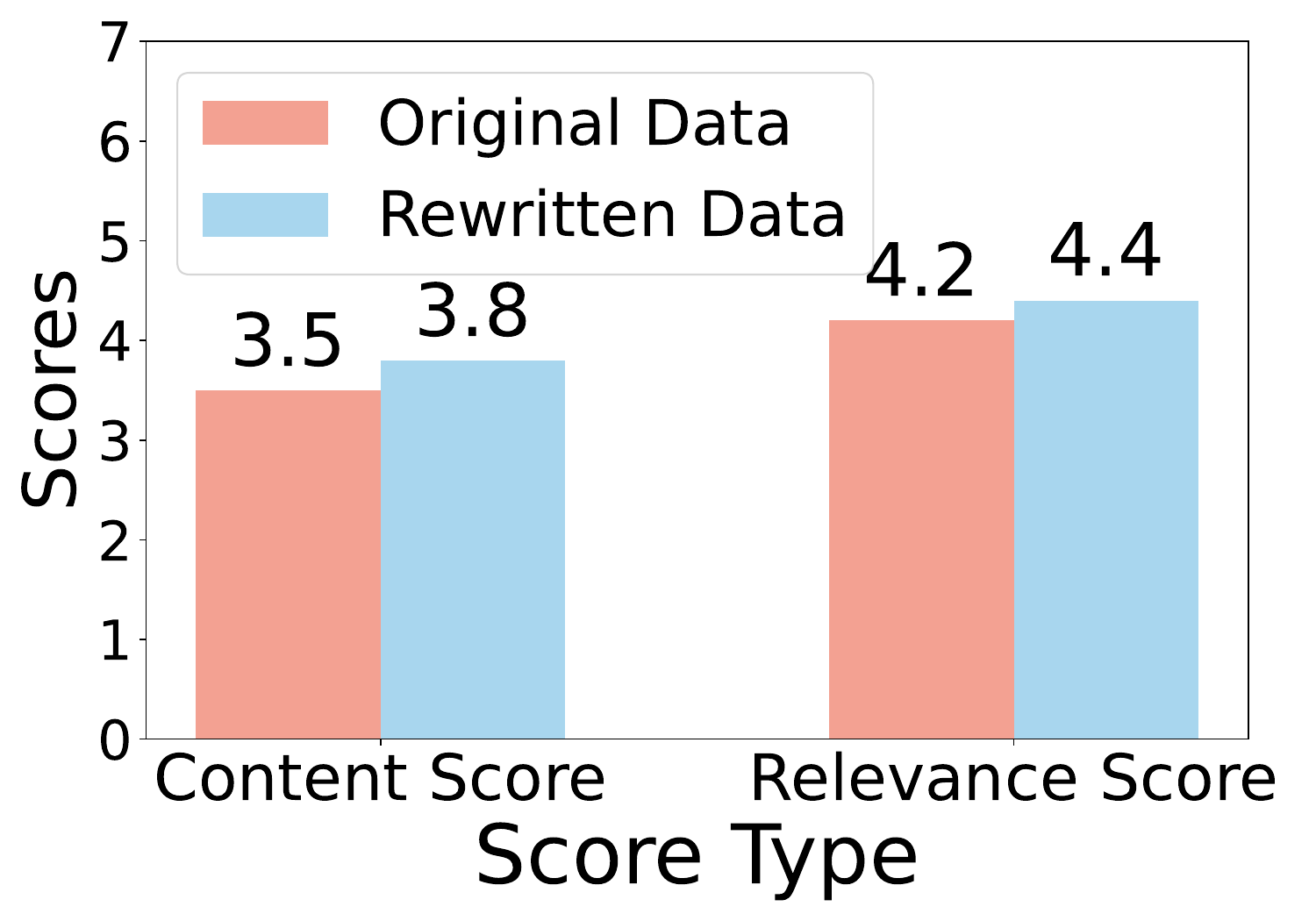}
    \hfill
    \includegraphics[width=0.235\textwidth]{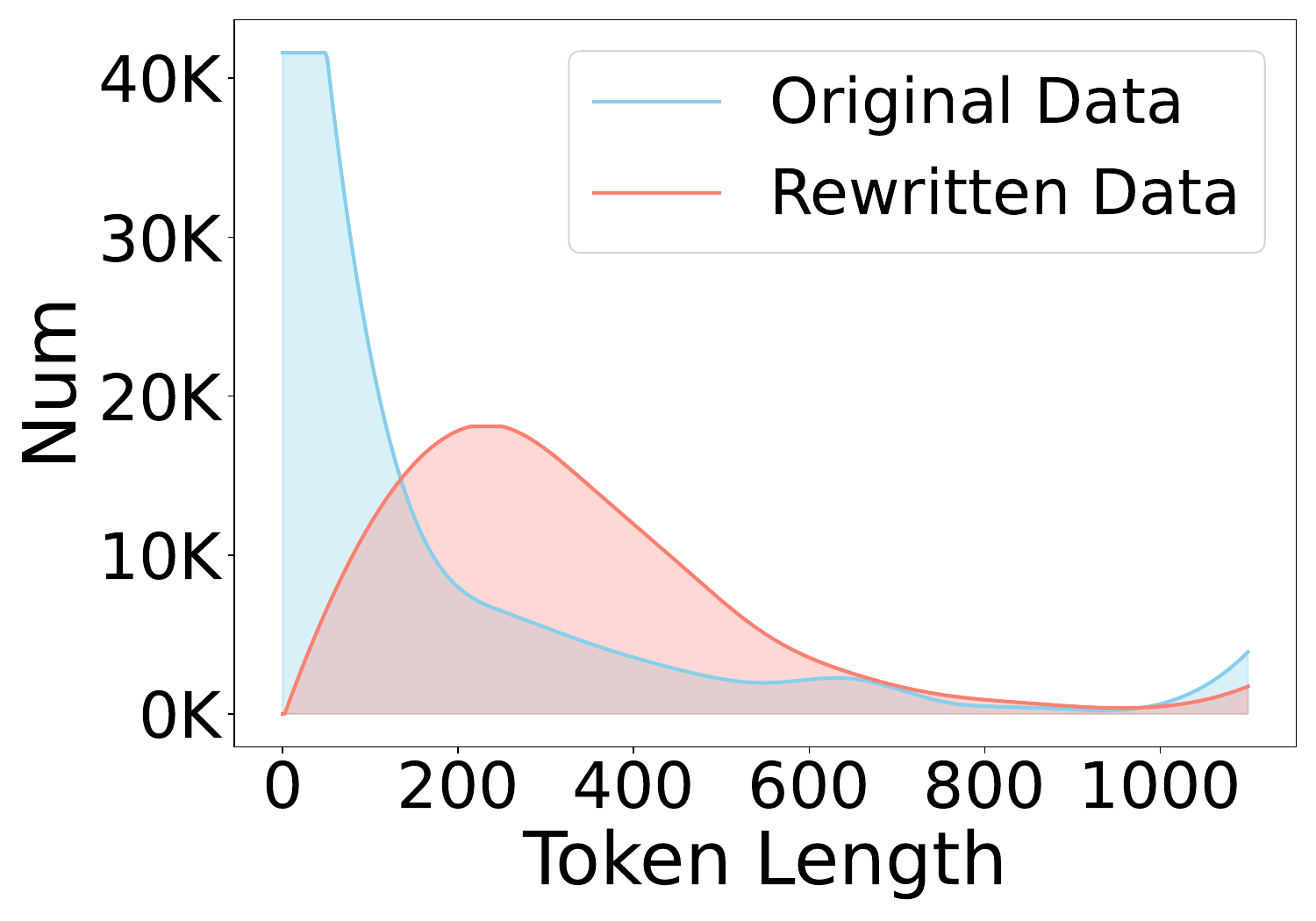}
    \caption{Comparison of original and rewritten data across two metrics: (1) Content and Relevance Scores judged by MLLMs show that rewritten data scores higher, indicating improved quality; (2) Token Length distribution suggests that rewritten data tends to be longer, including more tokens for rationales.}
    \label{fig:ablation_token_length}
\end{figure}

\subsection{Self-data Filtering}

A preliminary manual inspection of the rewritten data revealed instances of hallucinations, particularly in tasks such as OCR and chart interpretation. This underscores the necessity of a robust data filtering step to enhance the quality of the generated content. While human review is ideal for this purpose, it is both costly and impractical for large-scale applications. To address this challenge, we propose utilizing a "Model-as-Judge" approach to filter the data efficiently. Specifically, we leverage the InternVL2-Llama3-76B model, the same model used during the data rewriting process, to evaluate the logical consistency of each question-answer pair against the corresponding image. The assumption is that while the model may introduce inaccuracies during generation, it excels better in verification tasks. By implementing this filtering step, we ensure that the rewritten instructional data aligns closely with the visual information provided, minimizing hallucinations. 

\section{Analysis of \model-Instruct}
\subsection{Quality Verification}
To assess the overall quality of \model-Instruct, we randomly sample 1,000 data points from both the original and rewritten datasets in Group B. These samples are evaluated using the InternVL2-Llama3-76B model, focusing on two key quality dimensions: 1) Information Content Score (1–5): Evaluates the depth and richness of the conversation.
2) Relevance Score (1–5): Measures the alignment between the visual and textual components. The quality score prompt is included in Appendix \ref{app:prompts_fot_rewrite}.

The comparison presented in \autoref{fig:ablation_token_length} demonstrates that the rewritten dataset consistently surpasses the original dataset in both content and relevance scores. This indicates significant improvements in content quality and relevance after the rewriting process.

\subsection{Distribution Comparison}

\paragraph{Distribution of Data Length.}
\autoref{fig:ablation_token_length} illustrates the token length (instruction + response) distribution of the original data and rewritten data. The results show that the rewritten data demonstrates a broader and more evenly distributed token length, with a longer tail for larger token counts. This suggests that the rewriting process tends to generate longer texts, likely incorporating additional details or rationales to enhance explanation and clarity.

\begin{figure}[!t]
    \centering
    \includegraphics[width=1.0\linewidth]{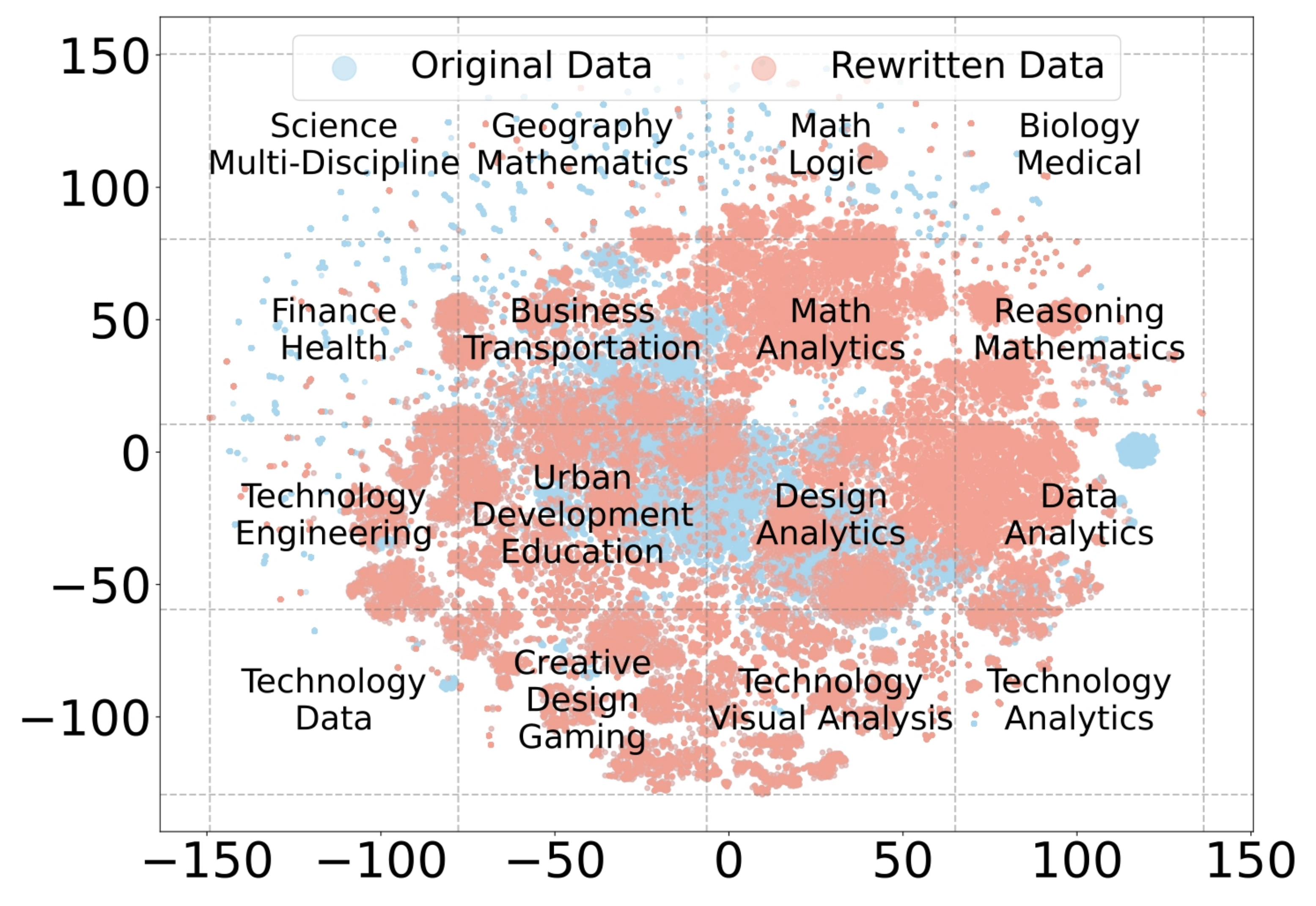}
    \caption{The t-SNE data distribution plot demonstrates how the rewritten data expands beyond the original dataset, increasing topic diversity and enhancing coverage of complex queries and reasoning. 
    }
    \label{fig:distribution_with_category}
\end{figure}

\paragraph{Distribution of Instructions.} To analyze the distributional differences between the original and rewritten data, we randomly sampled 80,000 examples from the dataset both before and after rewriting and visualized their distributions using t-SNE to project the instructions onto a two-dimensional plot \autoref{fig:distribution_with_category}. The resulting figure reveals two key takeaways: (1) The rewritten data exhibits significant overlap with the original data, indicating that it retains the core characteristics of the original distribution. This ensures that the rewritten data preserves the foundational structure of the dataset. (2) The rewritten data extends beyond the boundaries of the original distribution, demonstrating that it introduces new dimensions or variations, which shows that rewriting enhances the dataset by broadening its scope while maintaining its original essence. Based on this observation, during the experimental validation phase, we utilize a mixed dataset consisting of 70\% rewritten data and 30\% original data to train the model.

\subsection{Analysis of the Filtering Step} 

\paragraph{Model-based vs Human-based Filtering.} To evaluate the reliability of our model-based filtering approach, we conduct a comparative analysis between InternVL2-Llama3-76B and human evaluators. We sample 60 items and make them classified as either "good" or "bad" by both the model and three independent human evaluators. To measure inter-rater agreement, we employ Cohen's Kappa coefficient~\cite{mchugh2012interrater}, which accounts for chance agreement. The average Kappa value among the three human evaluators is 0.55. When substituting the model's classifications for one evaluator and calculating the average Kappa with the remaining two human evaluators, we obtained a higher average value of 0.64. This falls within the range consider to indicate good consistency (0.60-0.80), suggesting that our model-based filtering approach achieves reliable agreement with human judgment. Detailed results are presented in~\autoref{tab:consistency_of_model_user}.

\begin{table*}[!h]
\centering
\small
\begin{tabular}{@{}lccc@{}}
\toprule
 & \textbf{Stage-1} & \textbf{Stage-2} & \textbf{Stage-3} \\
 \midrule
\textbf{Resolution} & 384 & 384 $\times$ \{1$\times$1, ...\} & 384 $\times$ \{1$\times$1, ...\} \\
\textbf{\#Tokens} & 729 & Max 729$\times$5 & Max 729$\times$5 \\
\textbf{Dataset} & LCS & Single Image & Single, Multi-Image \& Video \\
\textbf{\#Samples} & 558K & 10M & 2M \\ \midrule
\textbf{Vision Tower} & \multicolumn{3}{c}{siglip-so400m-patch14-384} \\
\textbf{LLM Backbone} & \multicolumn{3}{c}{Qwen2.5-7B-Instruct} \\\midrule
\textbf{Trainable Model Parameters} & Projector: 20.0M & Full Model: 8.0B & Full Model: 8.0B \\
\textbf{Batch Size} & 512 & 256 & 256 \\
\textbf{Model Max Length} & 8192 & 8192 & 16384 \\
\textbf{Learning Rate: $\psi_{vision}$} & 1$\times$10$^{-3}$ & 2$\times$10$^{-6}$ & 2$\times$10$^{-6}$ \\
\textbf{Learning Rate: $\{\theta_{proj}, \Phi_{LLM}\}$} & 1$\times$10$^{-3}$ & 1$\times$10$^{-5}$ & 1$\times$10$^{-5}$ \\
\textbf{Epoch} & 1 & 1 & 1 \\ \bottomrule
\end{tabular}
\caption{Detailed configuration for each training stage of the \model-8B model.}
\label{tab:training_config}
\end{table*}

\begin{figure}[t]
    \centering
    \hspace{2pt}
    \includegraphics[width=\linewidth]{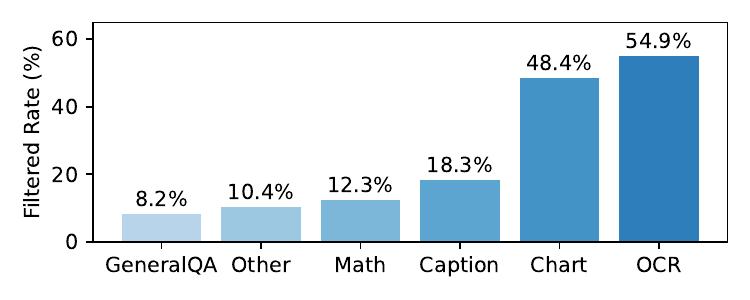}
    \caption{The filter rates of different data types after filtering, with a lower filtering rate seen in categories like GeneralQA and Math, while OCR and Chart data experience more extensive filtering. 
    }
    \label{fig:filter_rate}
\end{figure}

\paragraph{Category-specific Filtering Analysis.} Our filtering process reveals notable patterns across different data categories, as illustrated in~\autoref{fig:filter_rate}. Particularly high rates of hallucination are observed in visually complex categories such as OCR and Chart data, highlighting current MLLMs' insufficient capabilities of understanding OCR and chart contents. The importance of our filtering approach is further demonstrated through the ablation study in Appendix~\autoref{subsection:5.1}, which shows significant performance improvements in model training outcomes. We also show representative examples of correctly and incorrectly rewritten data in Appendix~\autoref{app:bad_case} and~\ref{app:good_case}.

\begin{table*}[t]
    \centering
\resizebox{1.0\linewidth}{!}{
\begin{tabular}{@{}l|cccccccc@{}}
\toprule
&\multicolumn{8}{c}{\textbf{Multi-Discipline Knowledge and Mathematical Reasoning}} \\
\cmidrule(lr){2-9}

\textbf{Model}
& \textbf{MMStar}
& \textbf{MMMU}
& \textbf{MMMU-Pro}
& \textbf{SeedBench}
& \textbf{MMBench}
& \textbf{MMVet}
& \textbf{MathVerse}
& \textbf{MathVista}\\
\cmidrule(lr){2-9}
 & test  & val & vision & test & en-test & test & mini-vision & testmini \\
\midrule
\rowcolor{gray!10}
GPT-4o~\citep{OpenAI2024} & 64.7  & 69.1 & 49.7  & 76.2 & 82.1 & 76.2 & 50.2 & 63.8 \\

\rowcolor{gray!10}
Gemini-1.5-Pro~\citep{team2023gemini} & 59.1
 & 65.8 & 44.4  & 76.0 & 73.9 & 64.0 & - & 63.9 \\
\rowcolor{gray!10}
Claude-3.5-Sonnet~\citep{Anthropic2024} & 62.2 & 68.3 & 48.0  & 72.2 & 79.7 & 75.4 & - & 67.7 \\
\midrule \midrule
\rowcolor{blue!10}
InternVL2-76B~\citep{chen2023internvl} & 67.1  & 58.2 & 38.0  & 77.6 & 86.5 & 64.4 & - & 65.5 \\
\rowcolor{blue!10}
Qwen2-VL-72B~\citep{wang2024qwen2} & 68.6  &64.5 & 37.1 & 	
77.9 & 86.9 & 73.9   & 37.3 &70.5 \\
\rowcolor{green!10}
LLaVA-OV-72B (SI)~\citep{li2024llava} & 65.2  & 57.4 & 26.0  & 77.6 & 86.6 & 60.0 & 37.7 & 66.5 \\
\rowcolor{green!10}
LLaVA-OV-72B~\citep{li2024llava} & 66.1 & 56.8 & 24.0 & 78.0 & 85.9 & 63.7 & 39.1 & 67.5 \\
\midrule
\rowcolor{blue!10}
MiniCPM-V-2.6-8B~\citep{yao2024minicpm} & 57.5  & 49.8 & 21.7   & 74.0 & 81.5 & 60.0 & - & 60.6 \\
\rowcolor{blue!10}
INXComp-2.5-7B~\citep{zhang2024internlm} & 59.9  & 42.9 & -   & 75.4 & 74.4 & 51.7 & 20.0 & 59.6 \\
\rowcolor{blue!10}
Llama-3.2-11B-Vision-Ins.~\citep{meta2024llama3.2} & 49.8 & 50.7 & 23.7 & 72.7 & 73.2 & 57.6 & 23.6 & 51.5 \\
\rowcolor{blue!10}
InternVL-2-8B~\citep{chen2023internvl} & 59.4  & 49.3 & 25.4 & 76.0 & 81.7 & 60.0 & 27.5 & 58.3 \\
\rowcolor{blue!10}
Qwen2-VL-7B-Ins.~\citep{wang2024qwen2} & 60.7 & 52.1 & 26.9  & 74.3 & 83.0 & 62.0 & 28.2 & 58.2 \\
\rowcolor{green!10}
Cambrian-1-8B~\citep{tong2024cambrian} & -   & 42.7 & 14.7 & 	
73.3 & 74.6 &  48.0 & - & 49.0 \\
\rowcolor{green!10}
Llava-CoT-11B~\citep{xu2024llavacotletvisionlanguage} & 57.6 & 48.9 & 18.5 & 75.2 & 75.0 & 60.3 & 24.2 & 54.8 \\
\rowcolor{green!10}
Molmo-8B-D~\citep{deitke2024molmo} & 50.5 & 45.3 & 18.9 & 74.1 & 73.6 & 58.0 & 21.5 & 51.6 \\
\rowcolor{green!10}
LLaVA-OV-7B (SI)~\citep{li2024llava} & 60.9 & 47.3 & 16.8 & 74.8 & 80.5 & 58.8 & 26.9 & 56.1 \\
\rowcolor{green!10}
LLaVA-OV-7B~\citep{li2024llava} & 61.7 & 48.8 & 18.7 & 75.4 & 80.8 & 58.6 & 26.2 & 63.2 \\
\midrule
\rowcolor{green!10}
\model-8B (SI) &  55.4 & 49.4 & \textbf{26.0} & 73.3 & 83.0 & 60.6 & \textbf{35.0} & 67.6 \\
\rowcolor{green!10}
\model-8B & \textbf{63.0}  & \textbf{50.8} & 25.3 & \textbf{76.0} & \textbf{83.4} & \textbf{62.3} & 34.2 & \textbf{67.6} \\
\rowcolor{cyan!20}
$\Delta$ \textbf{Over Best Open-Source ($\sim$10B Scale)} & +1.3 & +1.9 & +7.1 & +0.6 & +2.6 & +2.0 & +8.1 & +4.4 \\
\bottomrule
\end{tabular}}

\caption{Performance on multi-discipline knowledge and mathematical reasoning benchmarks. We highlight different groups of models with different colors: \colorbox{gray!10}{closed-source} models, \colorbox{blue!10}{open weights} but closed training details, and \colorbox{green!10}{fully open-source} models. Results are from official sources or running with lmms-eval package if unavailable.}
\label{tab:multi_discipline}
\end{table*}

\begin{table*}[t!]
    \centering
    \setlength{\tabcolsep}{8.5pt}
\resizebox{1.0\linewidth}{!}{
\begin{tabular}{@{}l|cccc|ccc@{}}
\toprule
 & \multicolumn{4}{c}{\textbf{Chart \& Doc Understanding}} & \multicolumn{3}{c}{\textbf{Multimodal Interactions \& Preferences}}\\

 \cmidrule(lr){2-5} \cmidrule(lr){6-8}

\textbf{Model}
& \textbf{AI2D}
& \textbf{ChartQA}
& \textbf{InfoVQA}
& \textbf{DocVQA}
& \textbf{RealWorldQA}
& \textbf{WildVision}
& \textbf{L-Wilder}\\
 \cmidrule(lr){2-5} \cmidrule(lr){6-8}

 &  test & test & test & test & test & 0617  & small \\

\midrule
\rowcolor{gray!10}
GPT-4o~\citep{OpenAI2024}  & 94.2 & 85.7 & 79.2 & 92.8 & 76.5 & 89.4  &  85.9 \\
\rowcolor{gray!10}
Gemini-1.5-Pro~\citep{team2023gemini}   & 94.4 & 87.2 & 81.0 & 93.1 & 70.4 & -  & - \\
\rowcolor{gray!10}
Claude-3.5-Sonnet~\citep{Anthropic2024}   & 94.7 & 90.8 & 49.7 & 95.2 & 60.1 &  50.0 & 83.1 \\
\midrule \midrule
\rowcolor{blue!10}
InternVL2-76B~\citep{chen2023internvl}  & 88.4 & 88.4 & 82.0 & 94.1  & 72.7 & -  & - \\
\rowcolor{blue!10}
Qwen2-VL-72B~\citep{wang2024qwen2}  & 88.1 & 88.3 & 84.5 & 96.5  & 77.8 & 52.3  & 53.6  \\
\rowcolor{green!10}
LLaVA-OV-72B (SI)~\citep{li2024llava}  & 85.1 & 84.9 & 74.6 & 91.8 & 73.8 & 49.5  & 72.9 \\
\rowcolor{green!10}
LLaVA-OV-72B~\citep{li2024llava}   & 85.6 & 83.7 & 74.9 & 91.3 & 71.9 & 52.3  & 72.0 \\
\midrule
\rowcolor{blue!10}
MiniCPM-V-2.6-7B~\citep{yao2024minicpm}  & 82.1 & 82.4 & - & 90.8 & 65.0  &  11.7  & - \\
\rowcolor{blue!10}
INXComp-2.5-7B~\citep{zhang2024internlm}   & 81.5 & 82.2 & 70.0 & 90.9 & 67.8 &   - & 61.4 \\
\rowcolor{blue!10}
Llama-3.2-11B-Vision-Ins~\citep{meta2024llama3.2} & 77.3 & 83.4 & 65.0 & 88.4 & 63.3 & 49.7 & 62.0 \\
\rowcolor{blue!10}
InternVL-2-8B~\citep{chen2023internvl}  & 83.8 & 83.3 & 74.8 & 91.6 & 64.4 & 51.5 & 62.5  \\
\rowcolor{blue!10}
Qwen2-VL-7B-Ins~\citep{wang2024qwen2}  & 83.0 & 83.0 & 76.5 & 94.5  & 70.1 & 44.0  & 66.3 \\
\rowcolor{green!10}
Cambrian-1-8B~\citep{tong2024cambrian}  & 73.3 & 73.3 & 41.6 & 77.8  & 64.2 &  - & 34.1 \\
\rowcolor{green!10}
Llava-CoT-11B~\citep{xu2024llavacotletvisionlanguage} & - & 67.0 & 44.8 & - & - & - & 65.3 \\
\rowcolor{green!10}
Molmo-7B-D~\citep{deitke2024molmo}   & 81.0 & 84.1 & 72.6 & 92.2  & 70.7 & 40.0 & - \\
\rowcolor{green!10}
LLaVA-OV-7B (SI)~\citep{li2024llava}  & 81.6 & 78.8 & 65.3 & 86.9 & 65.5 & 39.2  & 69.1  \\
\rowcolor{green!10}
LLaVA-OV-7B~\citep{li2024llava} & 81.4 & 80.0 & 68.8 & 87.5 & 66.3 & \textbf{53.8}  & 67.8  \\
\midrule
\rowcolor{green!10}
\model-8B (SI)  & 83.4 & 85.9 & \textbf{74.8} & \textbf{93.8} & \textbf{71.3} & 51.9  & \textbf{71.3} \\
\rowcolor{green!10}
\model-8B  & \textbf{84.0} & \textbf{86.2} & 73.1 & 93.7 & 69.9 & 51.1  & 70.8  \\
\rowcolor{cyan!20}
$\Delta$ \textbf{Over Best Open-Source ($\sim$10B Scale)} & +2.4 & +2.1 & +2.2 & +1.6 & +0.6 & -1.9 & +2.2 \\
\bottomrule
    \end{tabular}}
    \caption{Main results on Chart, Diagram, and Document Understanding, and Real-world Multimodal Interactions and Human Preferences benchmarks. Follow the same settings as in~\autoref{tab:multi_discipline}. }
    \vspace{-5pt}
    \label{tab:chart_document_realworld}
\end{table*}

 \section{Experiments}
\subsection{Model Training}
To demonstrate the effectiveness of \model-12M, we train an MLLM following the architecture of Llava-OneVision~\citep{li2024llava}, comprising a language tower, a vision tower, and a projector. We use Qwen2.5-7B-Instruct~\citep{qwen2.5} as the LLM backbone, Siglip-so400m-patch14-384~\citep{zhai2023sigmoid} as the vision tower, and a two-layer MLP as the projector. Following Llava-OneVision~\citep{li2024llava}, we divide the training into three stages, as illustrated in \autoref{tab:training_config}.
\begin{itemize}[leftmargin=*]
    \item \textbf{Stage-1: Language-Image Alignment.} The goal is to align the visual features well into the word embedding space of LLMs. We used the same pre-training corpus as LLaVA~\cite{liu2024llava}.
    \item \textbf{Stage-2: Visual Instruction Tuning (Single Image, SI).} The model is first trained on 10M single-image instructions randomly sampled from \model-12M, resulting in a model with strong performance at following a diverse set of instructions to complete visual tasks using a single image.
    \item \textbf{Stage-3: Visual Instruction Tuning (One Vision).} The model is then trained on a mixture of single-image, multi-image, and video data (2M). In this phase, the model expands its capabilities from single-image scenarios to diverse scenarios. The single-image dataset used in this stage is around 1M, which does not have any overlaps with Stage-2 training data.
\end{itemize}

\subsection{Model Evaluation Setup}
To reveal the generality and effectiveness of the model, we comprehensively evaluate it across different scenarios, including single-image, multi-image, and video benchmarks. Detailed results are presented in~\autoref{tab:multi_discipline},~\autoref{tab:chart_document_realworld} and~\autoref{tab:multi_image_and_video}, respectively. We denote the model checkpoint that completed the single-image stage and one-vision stage as \model-8B (SI) and \model-8B.

We conduct standardized, reproducible evaluations of our model across all 23 benchmarks using LMMs-Eval~\citep{zhang2024lmmsevalrealitycheckevaluation}.
To ensure a fair comparison with other MLLMs, we primarily report results from the original papers. When results are unavailable, we onboard the models in LMMs-Eval and evaluate them using consistent settings. All results are reported using greedy decoding and zero-shot settings unless specified.

\begin{table*}[t!]
    \centering
    \setlength{\tabcolsep}{3.5pt}
\resizebox{\linewidth}{!}{
\begin{tabular}{@{}l|cccccccc@{}}
\toprule
&\multicolumn{8}{c}{\textbf{Multi-Image and Video}} \\
\cmidrule(lr){2-9}

\textbf{Model}
& \textbf{MuirBench}
& \textbf{MEGABench}
& \textbf{EgoSchema}
& \textbf{PerceptionTest}
& \textbf{SeedBench}
& \textbf{MLVU}
& \textbf{MVBench}
& \textbf{VideoMME} \\
\cmidrule(lr){2-9}
 & test & test & test & test & video & dev & test & w/o subs  \\
\midrule
\rowcolor{gray!10}
GPT-4o~\citep{OpenAI2024} & 68.0 & 54.2 & - & - & - & 64.6 & - & 71.9 \\
\rowcolor{gray!10}
GPT-4V~\citep{OpenAI2023} & 62.3 & - & - & - & 60.5 & 49.2 & 43.5 & 59.9 \\
\midrule \midrule
\rowcolor{green!10}
LLaVA-OV-72B (SI)~\citep{li2024llava} & 33.2 & - & 58.6 & 62.3 & 60.9 & 60.9 & 57.1 & 64.8  \\
\rowcolor{green!10}
LLaVA-OV-72B~\citep{li2024llava} & 54.8 & 33.8 & 62.0 & 66.9 & 62.1 & 66.4 & 59.4 & 66.2  \\
\midrule

\rowcolor{blue!10}
InternVL-2-8B~\citep{chen2023internvl} & 59.4 & 27.7 & 54.2 & 57.4 & 54.9 & 30.2 & 66.4 & 54.0 \\
\rowcolor{blue!10}
Qwen2-VL-7B-Ins.~\citep{wang2024qwen2} & 41.6 & 36.0 & 66.7 & 62.3 & 55.3 & 58.6 & 67.0 & 63.3 \\
\rowcolor{green!10}
LLaVA-OV-7B (SI)~\citep{li2024llava} & 32.7 & 22.1 & 52.9 & 54.9 & 51.1 & 60.2 & 51.2 & 55.0 \\
\rowcolor{green!10}
LLaVA-OV-7B~\citep{li2024llava} & 41.8 & 23.9 & \textbf{60.1} & 57.1 & 56.9 & 64.7 & 56.7 & 58.2 \\
\midrule
\rowcolor{green!10}
\model-8B & \textbf{55.1} & \textbf{28.2} & 58.5 &\textbf{ 59.3} & \textbf{57.1}  &\textbf{64.7} & \textbf{59.1} & \textbf{58.8} \\
\rowcolor{cyan!20}
$\Delta$ \textbf{Over Best Open-Source $\sim$10B Scale)} & +13.3 & +4.3 & -1.6 & +2.2 & +0.2 & +0 & +2.4 & +0.6 \\
\bottomrule
\end{tabular}}
\caption{Main results on Multi-Image and Video benchmarks. Follow the same settings as in~\autoref{tab:multi_discipline}.}

\label{tab:multi_image_and_video}
\end{table*}

\subsection{Single-Image Performance}

\paragraph{Evaluation Benchmarks} To validate the performance for single-image tasks in real-world scenarios, we consider a comprehensive set of benchmarks in~\autoref{tab:multi_discipline} and~\autoref{tab:chart_document_realworld}. It can be categorized into three classes:
\begin{itemize}[leftmargin=*]
    \item \textbf{Chart, Diagram, and Document Understanding.} To evaluate the performance of models on OCR-based tasks, we assess it on four benchmarks: AI2D~\cite{kembhavi2016diagram},  ChartQA~\cite{masry2022chartqa}, DocVQA~\cite{mathew2020docvqa}, and InfoVQA~\cite{mathew2021infographicvqa}.
    \item \textbf{Multi-discipline Knowledge and Mathematical Reasoning.} To highlight the strong capabilities of our model in multi-disciplinary knowledge and mathematical reasoning, we evaluate it on eight benchmarks: MMStar~\cite{chen2024we}, MMMU~\cite{yue2023mmmu}, MMMU-Pro~\cite{yue2024mmmu}, SeedBench~\cite{li2023seedbenchbenchmarkingmultimodalllms}, MMBench~\cite{liu2024mmbenchmultimodalmodelallaround}, MMvet~\cite{yu2023mmvetevaluatinglargemultimodal}, Mathverse~\cite{zhang2024mathversedoesmultimodalllm}, Mathvista~\cite{lu2024mathvistaevaluatingmathematicalreasoning}.
    \item \textbf{Real-world Multimodal Interactions and Human Preferences.} Focusing solely on task-specific evaluations, without considering real-world interactions, provides a limited perspective of a model's true capabilities. To assess our model's performance in more realistic settings, we test it on three benchmarks: RealworldQA~\cite{grok15v}, WildVision~\cite{lu2024wildvisionevaluatingvisionlanguagemodels}, Llava-Wilder-Small~\cite{li2024llavanext-strong}.
\end{itemize}

\paragraph{Results Analysis.} (1) The experimental results show that \model-8B achieves state-of-the-art performance among open-source multimodal models across diverse benchmarks, approaching the performance of leading open-weight models. (2) Specifically, \model-8B achieves state-of-the-art performance on 9 benchmarks among both open-source and open-weight models particularly in mathematical reasoning. (3) In comparing \model-8B with its single-image variant \model-8B (SI), we observe that despite minor performance decreases on some benchmarks due to the introduction of multi-image and video, the overall performance remains robust.

\subsection{Multi-Image and Video Performance}

\paragraph{Evaluation Benchmarks} After completing Stage 3 of training, \model-8B has been trained to process both multi-image and video data. To evaluate its performance, we assess it on eight benchmarks: MuirBench~\cite{wang2024muirbenchcomprehensivebenchmarkrobust}, Megabench~\cite{chen2024megabenchscalingmultimodalevaluation}, EgoSchema~\cite{mangalam2023egoschema}, PerceptionTest~\cite{pătrăucean2023perceptiontestdiagnosticbenchmark}, SeedBench(Video)\cite{li2023seedbenchbenchmarkingmultimodalllms}, 
MLVU~\cite{zhou2024mlvucomprehensivebenchmarkmultitask}, MVBench~\cite{li2024mvbenchcomprehensivemultimodalvideo}, and VideoMME~\cite{fu2024video}. 

\paragraph{Results Analysis.} The experimental results demonstrate that \model-8B significantly advances the state of the art among open-source models, surpassing the previous leader LLaVA-One-Vision-7B, with a particularly striking 13-point improvement on MuirBench. While \model-8B shows promising results, a performance gap persists relative to Qwen2-VL-7B. This gap likely stems from our current data limitations - specifically, our training dataset includes only 1M samples of multi-image and video content due to computational constraints. This suggests substantial headroom for performance gains through expanded training data coverage of these modalities.

\section{Ablation Study}
In this section, we employ Qwen2.5-1.5B-Instruct as the backbone model for conducting ablation studies. These ablation studies evaluate the impact of key factors in our dataset creation and training process. We examine the effects of data filtering, data mixing ratio, training data scale, and the impact of rewrite model size, providing insights into the most important elements for optimizing multimodal learning. The following subsections detail these experiments.

\subsection{Effect of Data Filtering}
\label{subsection:5.1}

\begin{figure}
    \centering
    \includegraphics[width=0.9\linewidth]{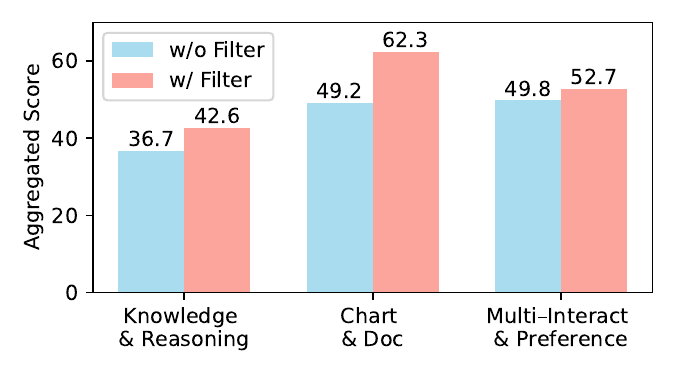}
    \caption{Data filtering significantly improves the quality of generated data, particularly in chart and document understanding, where hallucinations are more frequent.}
           \vspace{-10pt} 
    \label{fig:ablation_filter}
\end{figure}

To verify the necessity of filtering, we randomly sample 1M data from both the dataset before filtering and the dataset after filtering and conduct experiments on the same model. The experimental results in ~\autoref{fig:ablation_filter} demonstrate that, the model trained on the filtered dataset outperforms that trained on the no-filtering dataset, particularly in the benchmark for Chart, Diagram, and Document Understanding. This aligns with our previous observation that OCR and Chart data constitute the majority of the data filtered. This finding effectively validates the necessity and significance of hallucination mitigation. It also indirectly highlights the persistent issue of hallucinations in current MLLMs. See~\autoref{tab: effect_of_hallucinatory} for more details.

\subsection{Effect of Data Mixing Ratio}
\label{subsection:5.2}

\begin{figure}[t]
    \centering
    \includegraphics[width=1.0\linewidth]{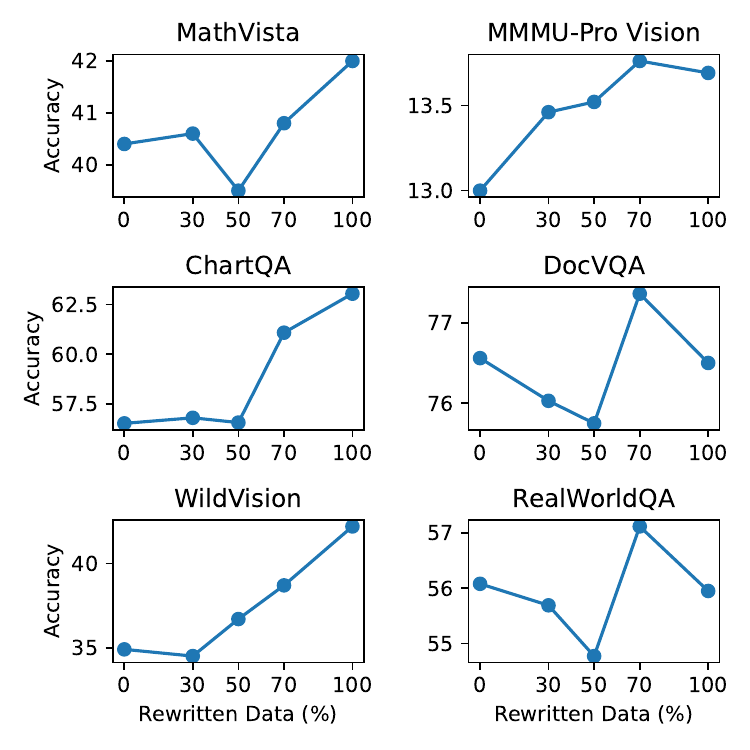}
    \caption{Effect of the mixing ratio of rewritten and original data. Performance generally improves with increased data rewriting. We select the 70\% rewritten data level as a balanced choice. }
    \label{fig:ablation_mix}
\end{figure}

To validate the effectiveness of merging original and rewritten datasets, we train five models using different combinations of original and rewritten data: one using only the original dataset, another using only the rewritten dataset, and three using combined datasets with original:rewritten ratios of 3:7, 7:3, and 5:5 respectively. As shown in~\autoref{fig:ablation_mix}, the model trained exclusively on rewritten data demonstrated superior average performance compared to the one trained on original data alone. Furthermore, the model trained on the combined dataset with a 3:7 ratio of original to rewritten data achieved marginally better performance than the rewritten-only model. These results suggest that merging the datasets is beneficial, likely due to the increased diversity and comprehensiveness of the training data. Detailed performance metrics can be found in~\autoref{tab:ablation_mix_main_result}.

\begin{figure*}[t]
    \centering
    \includegraphics[width=0.8\linewidth]{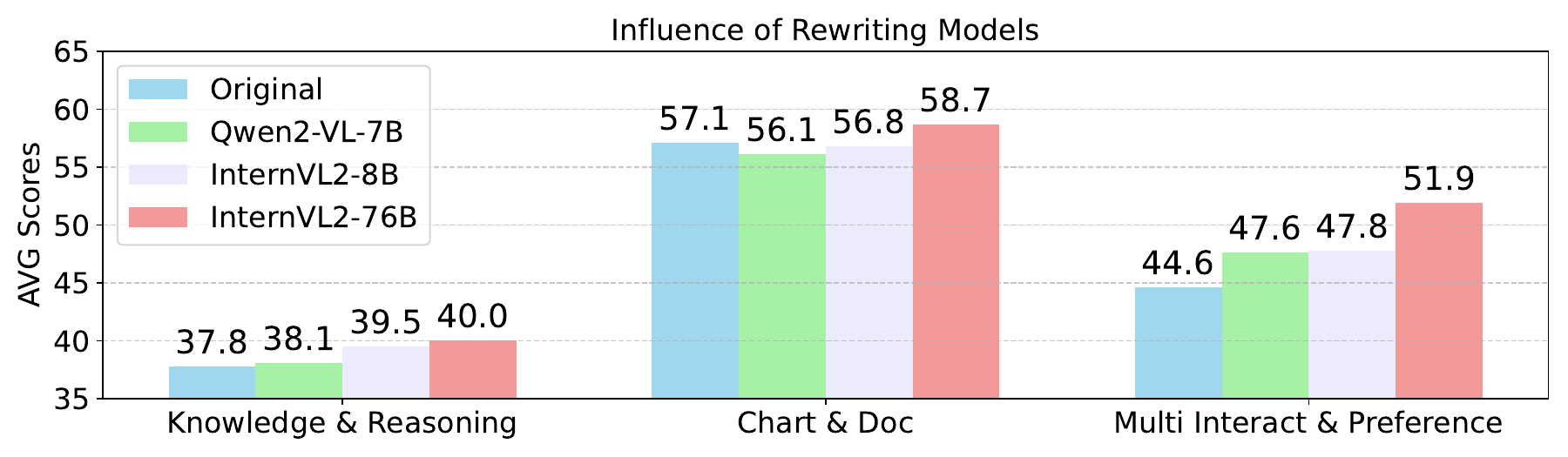}
    \caption{Performance of data rewritten by different models on three benchmark subsets}
    \label{fig:ablation_rewrite_model}
\end{figure*}

\subsection{Effect of Training Data Scale}
\label{subsection:5.3}
As shown in~\autoref{fig:enter-label} and \autoref{fig:ablation_scale_appendix_combined}, performance is tracked across benchmarks with training dataset size increasing in 2-million-sample intervals. Results are compared to three leading models: LLaVA-OneVision-72B, Llava-CoT-11B \citep{xu2024llavacotletvisionlanguage}, and LLaVA-OneVision-7B. The findings demonstrate a positive correlation between training data scale and performance, indicating that diverse instruction data improves the model's ability to handle complex tasks. 
Interestingly, performance on the MMStar and SeedBench tends to decline in late stage-2 training. Investigation shows that this decline primarily stems from decreased performance in Scene Understanding and Instance Attributes subsets. We suppose that as models develop capabilities for processing more complex images and questions, they may prioritize understanding object relationships over detailed attribute recognition.

\subsection{Impact of Rewrite Model Size}
To assess the effect of model size on the quality of rewritten data, we conduct experiments using four models trained on a dataset of 500K samples. The first model is trained on the original dataset. The second model use data rewritten by InternVL2-Llama3-76B and Meta-Llama-3-70B-Instruct. The third model is trained on data rewritten by Qwen2-VL-7B-Instruct and Qwen2.5-7B-Instruct. The fourth model is trained on data rewritten by InternVL2-8B and InternLM2.5-7B. Among these, Qwen2.5-72B-Instruct, Qwen2.5-7B-Instruct, and InternLM2.5-7B are employed solely for rewriting caption data, while InternVL2-Llama3-76B, Qwen2-VL-7B-Instruct, and InternVL2-8B are used for data filtering.

As shown in~\autoref{fig:ablation_rewrite_model}, our analysis reveals distinct patterns in model performance across different task categories. For knowledge \& reasoning tasks, models trained on data rewritten by smaller models (approximately 7B parameters) achieve performance comparable to those using larger model rewrites. However, the impact of data rewriting varies significantly by task type. For chart and document-related tasks, rewriting with smaller models actually leads to performance degradation, while larger models provide modest improvements. This suggests that sophisticated visual understanding capabilities of larger models are crucial for effectively rewriting such data. In contrast, Multi Interact \& Preference tasks demonstrate a clear correlation with model scale, where larger models excel in handling these complex scenarios that demand subtle understanding and nuanced preference modeling. Comprehensive performance metrics across all task categories are detailed in~\autoref{tab:ablation_different_rewrite_model_detail}.



\subsection{Qualitative Analysis}
\label{sec:qualitative_analysis}
In~\autoref{app:real_examples}, we present a diverse collection of question-and-answer pairs drawn from real-world scenarios to evaluate our model's practical capabilities. These examples showcase the model's ability to handle complex, unconstrained queries while providing meaningful, contextually appropriate responses. The analysis of these Q\&A pairs reveals not only the model's comprehension and response generation capabilities but also offers concrete evidence of its real-world applicability. Through these examples, we gain valuable insights into both the model's current strengths and areas for potential improvement in future iterations.


\section{Related Work}


\paragraph{Multimodal Instruction Data.}
The construction of high-quality, large-scale multimodal datasets is a cornerstone for advancing MLLMs. However, this process presents significant challenges, particularly in open-source settings where resource constraints are pronounced. Existing methods for dataset generation typically fall into three categories. First, human annotation provides precise and contextually rich datasets~\citep{xu2024vision, deitke2024molmo, mckinzie2024mm1,sun2023aligning}, but it is prohibitively expensive and labor-intensive, limiting scalability. Second, repurposing academic datasets such as visual question-answering collections offers cost-effective alternatives and can be enhanced through automated or manual instruction templates~\citep{tong2024cambrian, liu2024gpt}, but these datasets often lack diversity and fail to support nuanced reasoning tasks. Third, querying proprietary models like GPT-4 enables the generation of high-quality and diverse datasets~\citep{chen2023sharegpt4v, luo2024mmevol,liu2024llava,wang2023see,chen2024allava}. However, this approach incurs substantial computational costs and raises licensing issues, making it less feasible for open-source initiatives. Our work addresses these challenges by introducing a simple, scalable, and cost-effective methodology for constructing multimodal datasets exclusively using open-source models, combining task-specific data augmentation and rigorous quality filtering to provide a robust foundation for open-source MLLMs without relying on proprietary resources.

\paragraph{Multimodal Large Language Models.}
MLLMs have advanced AI by seamlessly integrating text and visual processing, with proprietary systems like GPT-4o and Gemini~\citep{team2023gemini} achieving state-of-the-art performance. However, these models remain inaccessible, leaving open-source alternatives at a disadvantage due to limited resources and data. To address this gap, connector-based approaches~\citep{li2023blip,NEURIPS2023_9a6a435e} like LLaVA~\citep{li2024llava} have emerged as efficient solutions, linking visual encoders to language models using lightweight projection modules. Despite these innovations, the primary challenge for open-source MLLMs is the scarcity of high-quality supervised fine-tuning data~\citep{bai2024survey}, which is essential for advancing their capabilities. Our work tackles this bottleneck by scaling and improving SFT datasets while building on the connector-training paradigm. Through these efforts, we aim to bridge the gap between proprietary and open-source MLLMs, enabling open-source competitive multimodal systems.

\section{Conclusion}

In this paper, we introduce a streamlined yet scalable approach to enhancing MLLM performance through the strategic use of open-source models to generate diverse, high-quality training data that reflects human preferences and real-world complexity. At the core of our contribution is the \model-Instruct dataset, containing 12 million multimodally enriched entries, which serves as the foundation for our \model-8B architecture. This model achieves state-of-the-art performance across various multimodal tasks while reducing reliance on costly proprietary models. Our experimental results confirm that this approach substantially improves MLLMs' practical capabilities, particularly in handling diverse, realistic scenarios. Beyond advancing multimodal understanding, our method democratizes access to advanced AI development, making sophisticated modeling techniques accessible to a broader range of researchers and organizations. This work opens promising avenues for future research into expanding the methodology across different modalities and datasets.


\bibliography{custom}

\appendix

\clearpage
\setcounter{page}{1}

\DoToC

\newpage
\onecolumn
\appendix


\section{Additional Results of Ablation Study}
\label{app:ablation_study_supply}

\lstset{
    basicstyle=\small
}

\definecolor{darkorange}{RGB}{255, 140, 0}
\definecolor{darkblue}{RGB}{84, 112, 198}
\definecolor{lightgreen}{RGB}{145, 204, 117}
\definecolor{lightyellow}{RGB}{250, 200, 88}
\definecolor{lightred}{RGB}{238, 102, 102}
\definecolor{lightblue}{RGB}{115, 192, 222}

\newtcolorbox{promptbox}[2][Prompt]{
colback=black!5!white,
arc=5pt, 
boxrule=0.5pt,
fonttitle=\bfseries,
title=#1, 
before upper={\small}, fontupper=\fontfamily{ptm}\selectfont,
colframe=#2, 
}

\renewcommand{\thetable}{A\arabic{table}} 
\setcounter{table}{0}
\renewcommand{\thefigure}{A\arabic{figure}} %
\setcounter{figure}{0}

\subsection{Breakdown performance on each benchmark before and after filtering}

\begin{figure*}[t]
    \centering
    \begin{minipage}{0.3\linewidth}
        \centering
        \includegraphics[width=\linewidth]{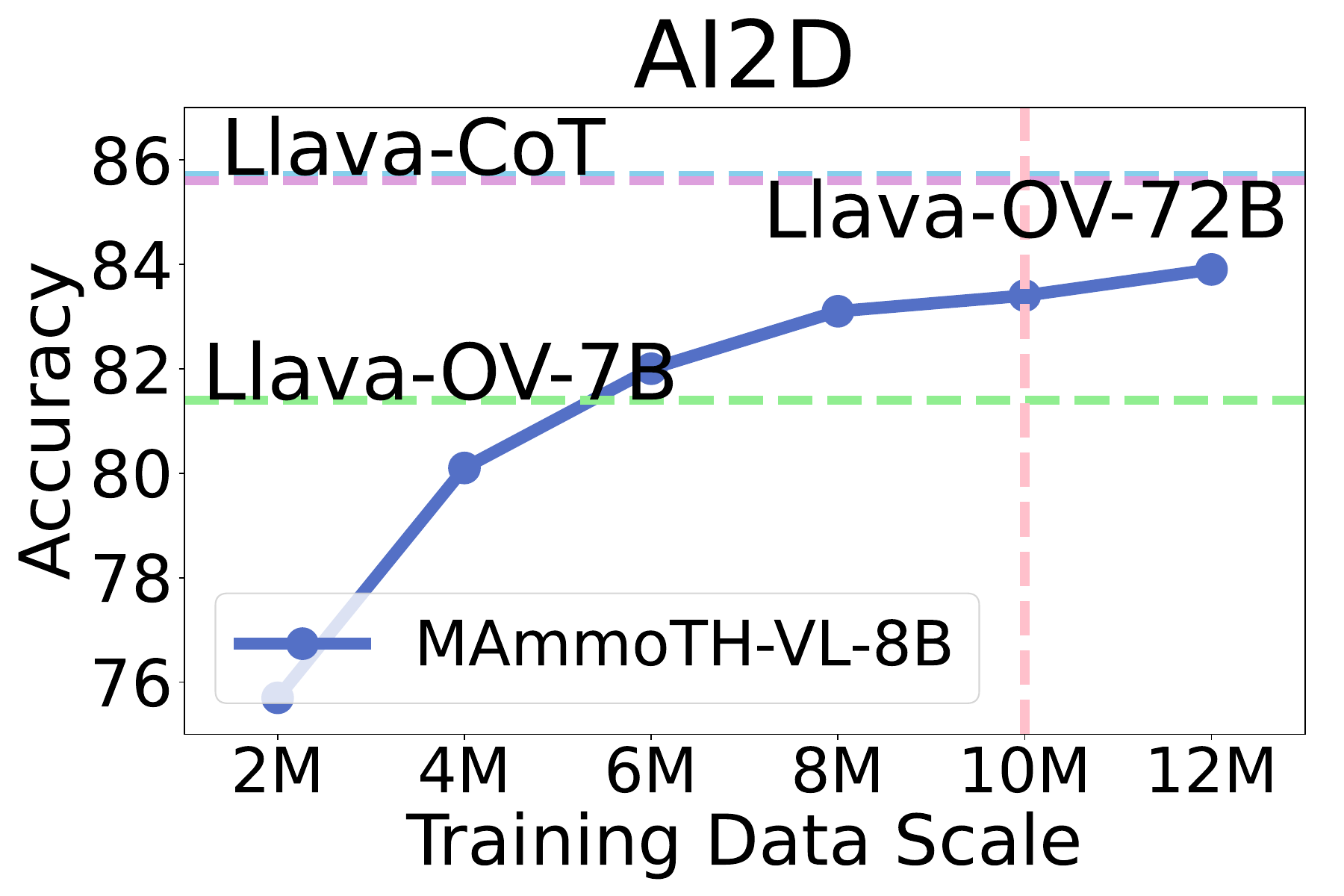}
    \end{minipage} \hfill
    \begin{minipage}{0.3\linewidth}
        \centering
        \includegraphics[width=\linewidth]{scale_pdf_cot/chartqa.pdf}
    \end{minipage} \hfill
    \begin{minipage}{0.3\linewidth}
        \centering
        \includegraphics[width=\linewidth]{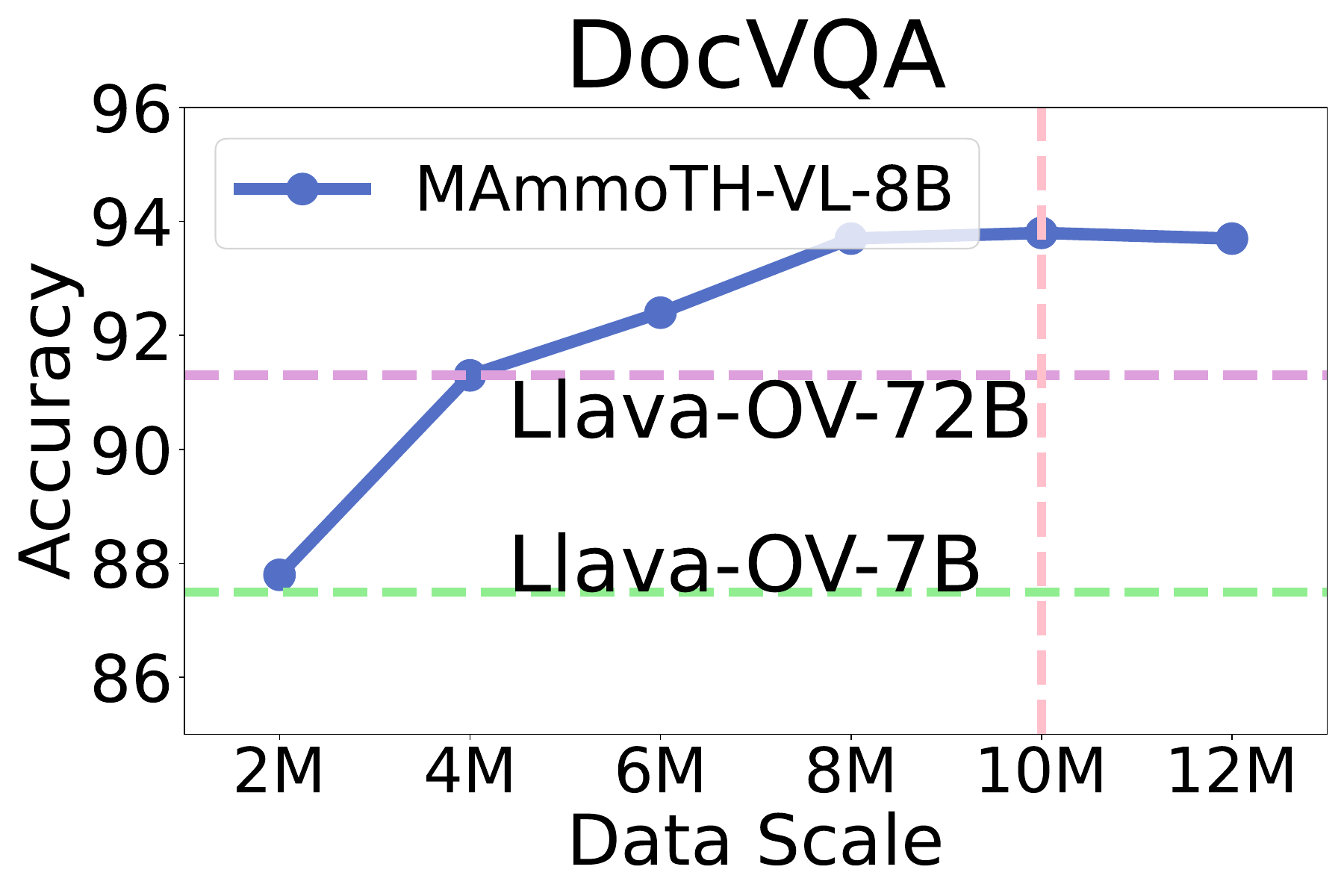}
    \end{minipage}
    
    \vspace{4pt} 

    \begin{minipage}{0.3\linewidth}
        \centering
        \includegraphics[width=\linewidth]{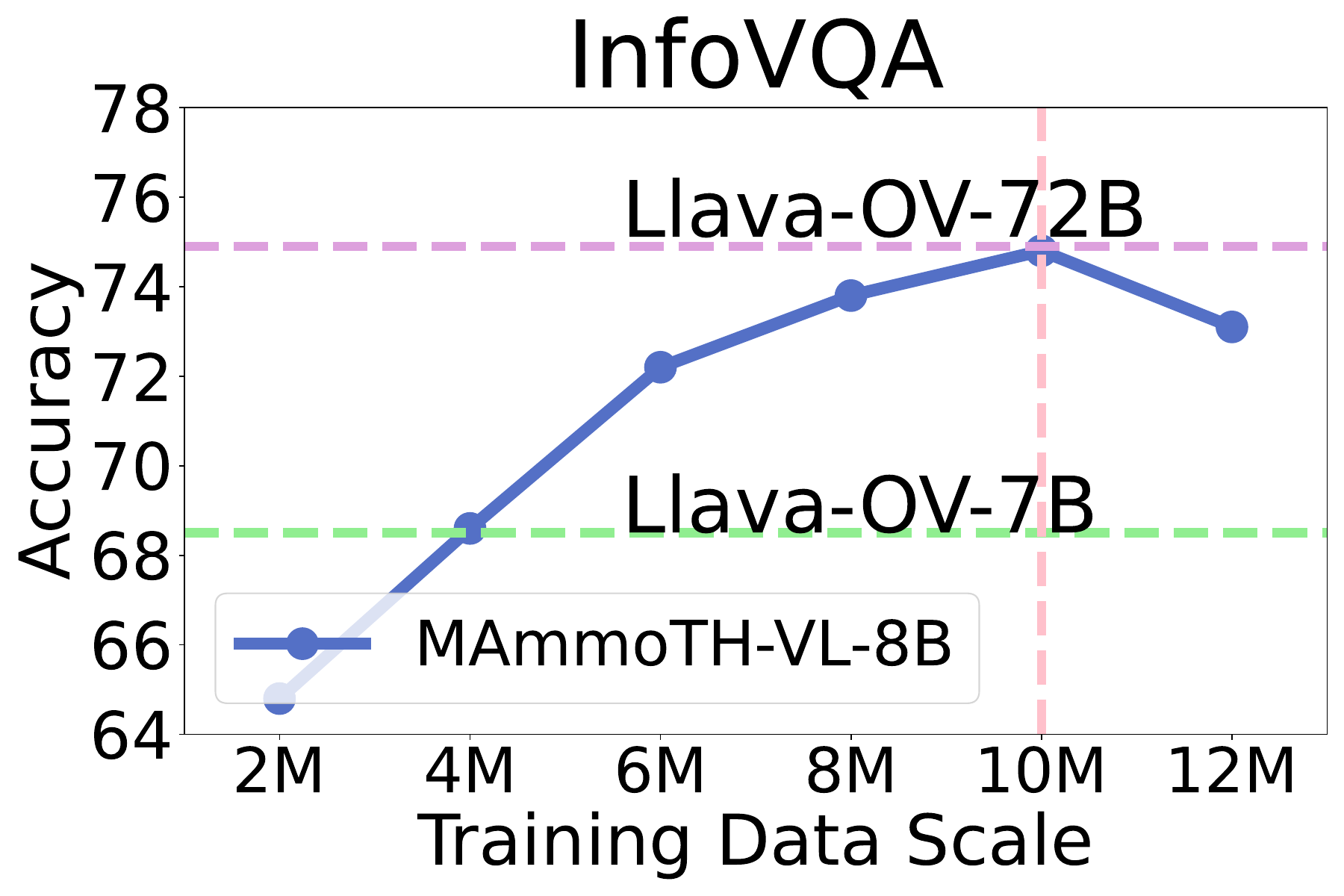}
    \end{minipage} \hfill
    \begin{minipage}{0.3\linewidth}
        \centering
        \includegraphics[width=\linewidth]{scale_pdf_cot/realworldqa.pdf}
    \end{minipage} \hfill
    \begin{minipage}{0.3\linewidth}
        \centering
        \includegraphics[width=\linewidth]{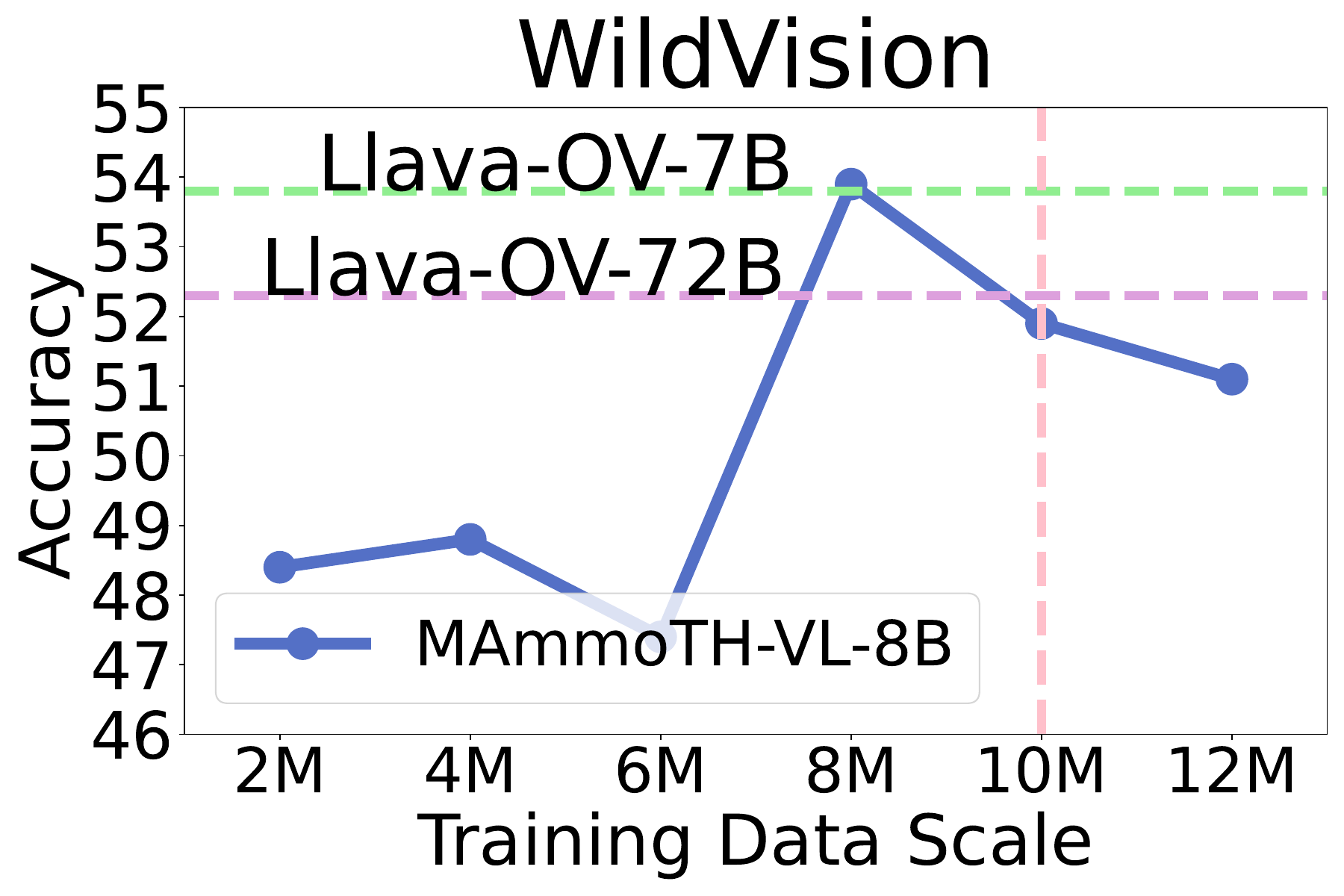}
    \end{minipage}

    \vspace{4pt}

    \begin{minipage}{0.3\linewidth}
        \centering
        \includegraphics[width=\linewidth]{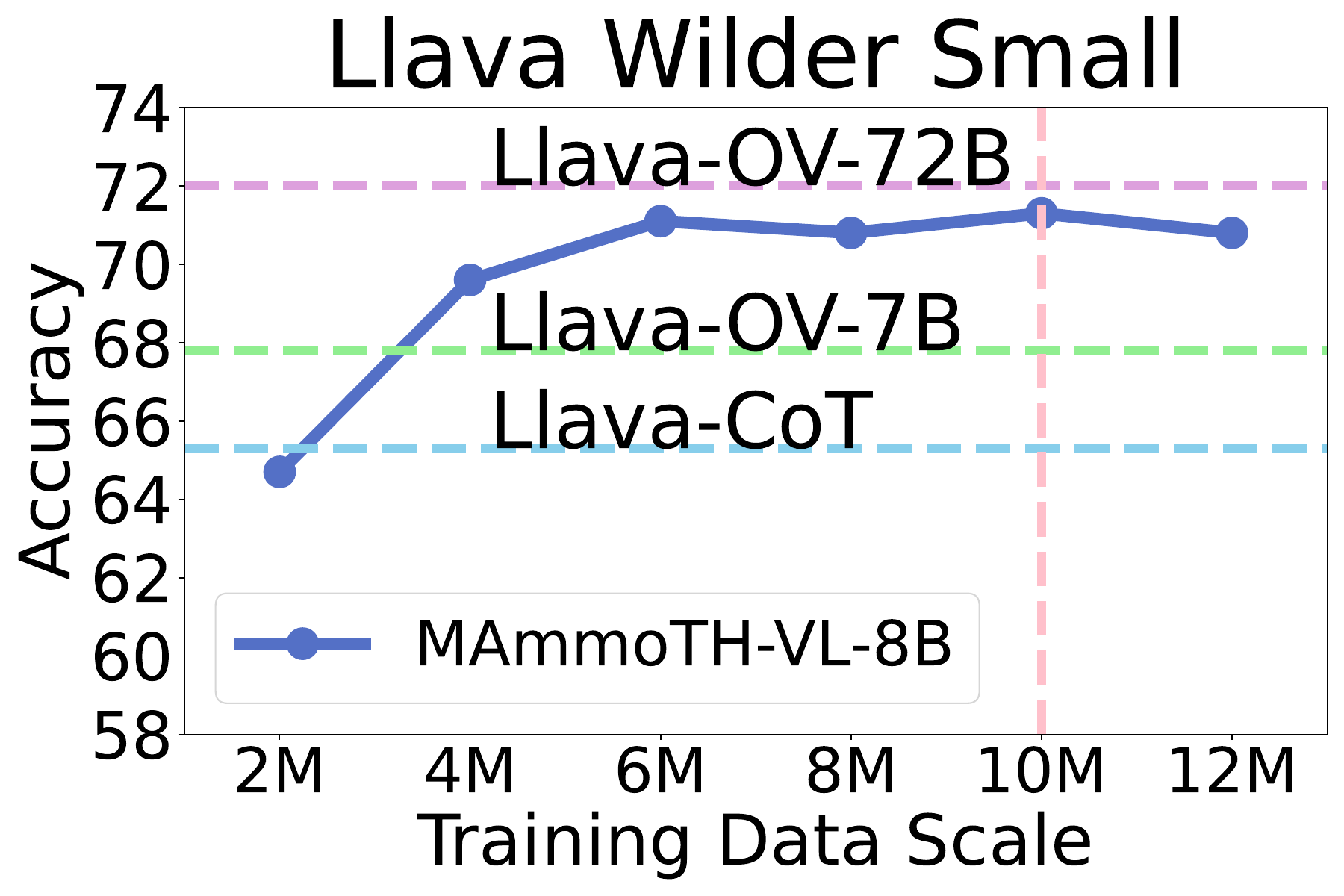}
    \end{minipage} \hfill
    \begin{minipage}{0.3\linewidth}
        \centering
        \includegraphics[width=\linewidth]{scale_pdf_cot/mmmu.pdf}
    \end{minipage} \hfill
    \begin{minipage}{0.3\linewidth}
        \centering
        \includegraphics[width=\linewidth]{scale_pdf_cot/mmmu_pro.pdf}
    \end{minipage}

    \vspace{4pt}

    \begin{minipage}{0.3\linewidth}
        \centering
        \includegraphics[width=\linewidth]{scale_pdf_cot/mmvet.pdf}
    \end{minipage} \hfill
    \begin{minipage}{0.3\linewidth}
        \centering
        \includegraphics[width=\linewidth]{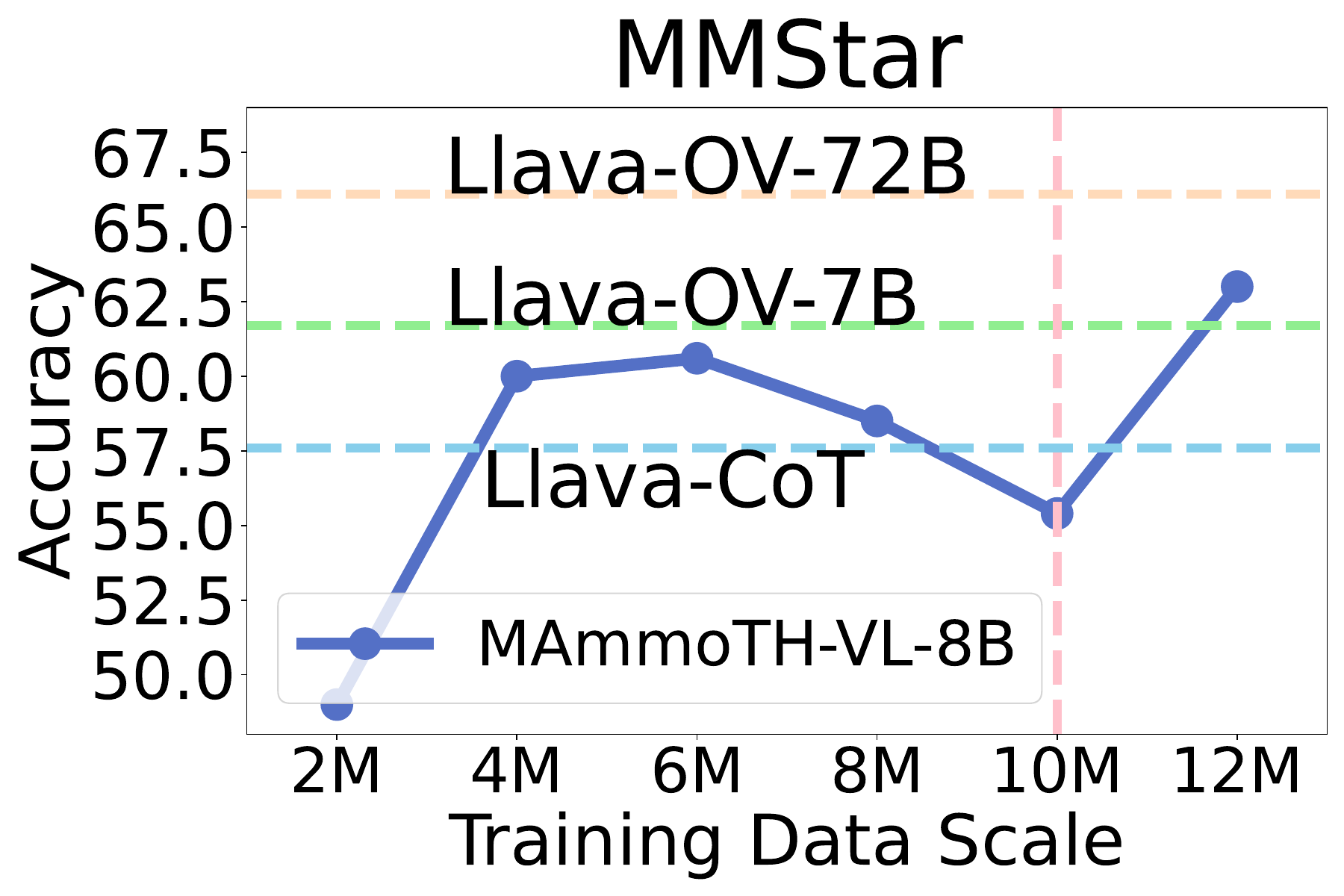}
    \end{minipage} \hfill
    \begin{minipage}{0.3\linewidth}
        \centering
        \includegraphics[width=\linewidth]{scale_pdf_cot/mmbench_en.pdf}
    \end{minipage}

    \vspace{4pt}

    \begin{minipage}{0.3\linewidth}
        \centering
        \includegraphics[width=\linewidth]{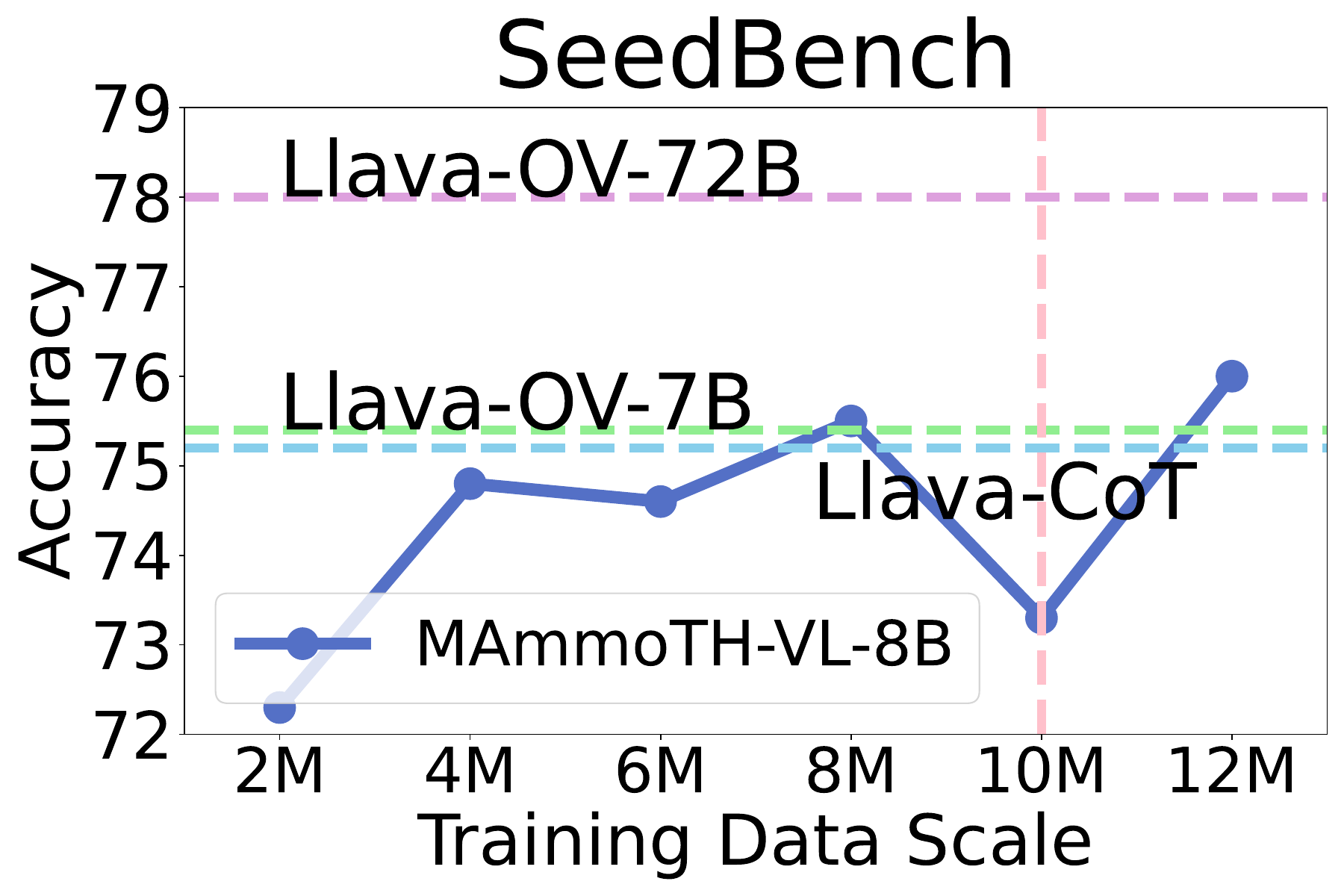}
    \end{minipage} \hfill
    \begin{minipage}{0.3\linewidth}
        \centering
        \includegraphics[width=\linewidth]{scale_pdf_cot/mathverse.pdf}
    \end{minipage} \hfill
    \begin{minipage}{0.3\linewidth}
        \centering
        \includegraphics[width=\linewidth]{scale_pdf_cot/mathvista.pdf}
    \end{minipage}

    \vspace{-1pt}
    \caption{\model-8B's overall performance across various benchmarks as the training dataset size increases.}
    \label{fig:ablation_scale_appendix_combined}
    \vspace{-1pt}
\end{figure*}

To evaluate the impact of our filtering approach, we conduct experiments using the Qwen2.5-1.5B-Instruct model under identical experimental conditions. We randomly sample 1M instances from the same dataset in both its pre-filtered and post-filtered versions to ensure a fair comparison.
The experimental results in~\autoref{tab: effect_of_hallucinatory} demonstrate that the model trained on the after filtering dataset outperforms that trained on the before filtering dataset, particularly in the benchmark for Chart, Diagram, and Document Understanding. This aligns with our previous observation that OCR and Chart data constitute the majority of the data filtered. This finding effectively validates the necessity and significance of hallucination mitigation. It also indirectly highlights the persistent issue of hallucinations in current MLLMs.

~\autoref{tab: num_of_filter} shows the specific numbers and retention rates for each type of data before and after filtering.

\begin{table*}[ht]
\centering
\begin{tabular}{lcccc}
\midrule
\textbf{Bench Name} & \textbf{Before Filter} & \textbf{After Filter} \\
\midrule
MMMU             & 39.6  & \textbf{40.9}      \\
MMStar                & 14.0  & \textbf{44.6}  \\
SeedBench             & 66.4  & \textbf{67.9}   \\
MMMU-Pro Vision & \textbf{15.5}  & 13.7  \\
MathVista   & 39.5   & \textbf{42.0}   \\
MMBench EN   &  58.6  & \textbf{65.1}   \\
MMVet   &   40.5 & \textbf{43.9}   \\
MathVerse &  19.3   & \textbf{22.6}   \\
AI2D          & 56.9   & \textbf{61.8}   \\
ChartQA       & 26.8  & \textbf{63.0}   \\
InfoVQA       & 41.5  & \textbf{48.0}   \\
DocVQA        & 71.7   & \textbf{76.5}   \\
L-Wilder Small   &  58.8  & \textbf{59.8}   \\
WildVision      & 40.2   & \textbf{42.2}    \\
RealWorldQA       & 50.3   & \textbf{56.0}  \\
Avg & 42.6 & \textbf{49.9}	\\
\midrule
\end{tabular}
\caption{Performance Comparison of Models Trained on Filtered versus Unfiltered Data Across Multiple Benchmarks.}
\label{tab: effect_of_hallucinatory}
\end{table*}

\begin{table*}[!h]\centering
\begin{tabular}{@{}lccc@{}}\toprule
\textbf{Data Type} & Before Filter & After Filter & Filter Rate \\\midrule
OCR & 1104960 & 498337 & 54.9 \\
Chart & 7326189 & 3782029 & 48.4 \\
GeneralQA & 1726180 & 1584308 & 8.2\\
Caption & 244874 & 199853 & 18.3 \\
Math & 590894 & 518393 & 12.3 \\
Other & 1315039 & 1178275 & 10.4\\
\bottomrule
\end{tabular}
\caption{Filter Rates Of Different Data Types After Data Filtering.}
\label{tab: num_of_filter}
\end{table*}


\clearpage

\subsection{Specific performance on each benchmark for different mix ratio}

To validate the effectiveness of merging original and rewritten datasets, we conduct experiments by training the model in five data schedules: first trained on the original dataset, second on the rewritten dataset, third on a combined dataset with a 3:7 ratio of original to rewritten data, fourth with a 7:3 ratio and fifth with a 5:5 ratio. The results in~\autoref{tab:ablation_mix_main_result} indicate that the model trained on the rewritten dataset achieved better average performance compared to the model trained on the original dataset. In addition, the model trained on the combined dataset with the merge ratio of 3:7 shows a slight performance gain which is better than rewritten. This demonstrates the effectiveness of merging datasets, as it enables the construction of a more diverse and comprehensive dataset.

\begin{table*}[!h]
    \centering
    \begin{tabular}{lccccc}
        \toprule
        \textbf{Bench Name} & \textbf{Rewrite} & \textbf{Original} & \textbf{Mix 3:7} & \textbf{Mix 7:3} & \textbf{Mix 5:5} \\
        \midrule
        MMMU & 40.9 & \textbf{41.9} & 41.5 & 41.3 & 41.7 \\
        MMStar & \textbf{44.6} & 43.3 & 43.4 & 42.3 & 43.7 \\
        SeedBench & 67.9 & \textbf{69.9} & 68.7 & 69.3 & 68.9 \\
        MMMU-Pro Vision & 13.7 & 13.0 & \textbf{13.8} & 13.5 & 13.5 \\
        MathVista & \textbf{42.0} & 40.4 & 41.8 & 40.6 & 39.5 \\
        MMBench EN & 65.1 & 67.8 & 66.1 & \textbf{67.9} & 66.4 \\
        MMVet & 43.9 & 37.3 & \textbf{45.5} & 40.7 & 38.9 \\
        MathVerse & \textbf{22.6} & 19.8 & 21.4 & 21.0 & 20.4 \\
        AI2D & 61.8 & \textbf{63.1} & 62.9 & 62.5 & 62.8 \\
        ChartQA & \textbf{63.1} & 56.5 & 61.1 & 56.8 & 56.6 \\
        InfoVQA & 48.0 & 47.3 & \textbf{49.0} & 45.7 & 45.6 \\
        DocVQA & 76.5 & 76.6 & \textbf{77.4} & 76.0 & 75.7 \\
        L-Wilder Small & 59.8 & 56.4 & \textbf{60.9} & 56.8 & 57.4 \\
        WildVision & \textbf{42.2} & 34.9 & 38.7 & 34.5 & 36.7 \\
        RealworldQA & 56.0 & 56.1 & \textbf{57.1} & 55.7 & 54.8 \\
        Avg & 49.9 & 48.3 & \textbf{50.0} & 48.3 & 48.2 \\
        \bottomrule
    \end{tabular}
    \caption{Benchmark Performance Of Models Trained On Data With Different Mix Ratios.}
    \label{tab:ablation_mix_main_result}
\end{table*}

\subsection{Specific performance on each benchmark as trained data increased}

~\autoref{fig:ablation_scale_appendix_combined} show 
\model-8B's overall performance across all single image benchmarks as the training dataset size increased, with each node representing 2 million data points. The results are compared against those of LLaVA-OneVision-7B, Llava-CoT-11B, and LLaVA-OneVision-72B. This trend clearly demonstrates that expanding the scale of instruction data has a significant positive effect on model performance. This observation suggests that as more diverse instruction data is introduced, the model’s ability to handle complex tasks is enhanced.

\newpage
\subsection{Additional Details of different models for data rewriting}

\begin{table*}[!h]
    \centering
\small
    \begin{tabular}{lcccc}
        \toprule
        \textbf{Bench Name} & \textbf{Original} & \textbf{Rewrite (Qwen2-VL-7B)} & \textbf{Rewrite (InternVL2-8B)} & \textbf{Rewrite (InternVL2-76B)} \\
        \midrule
            MMMU & 40.4 & 40.6 & \textbf{40.9} & 40.78 \\
            MMStar & 40.9 & 41.7 & \textbf{41.7} & 37.9 \\
            SeedBench & 50.6 & 52.1 & 65.0 & \textbf{67.0} \\
            MMMU-Pro Vision & 12.3 & 12.9 & 12.9 & \textbf{15.3} \\
            MathVista & 36.4 & 38.8 & 37.4 & \textbf{39.0} \\
            MMBench EN & \textbf{65.8} & 59.1 & 60.1 & 58.3 \\
            MMVet & 38.6 & 38.1 & 38.6 & \textbf{41.1} \\
            MathVerse & 17.6 & \textbf{21.6} & 19.8 & 20.6 \\
            AI2D & 61.8 & \textbf{62.3} & 61.7 & 59.6 \\
            ChartQA & 49.4 & 48.1 & 50.6 & \textbf{58.7} \\
            InfoVQA & 43.8 & 43.1 & 43.7 & \textbf{44.3} \\
            DocVQA & \textbf{73.4} & 70.8 & 71.3 & 72.2 \\
            L-Wilder Small & 44.5 & 55.7 & 55.7 & \textbf{60.5} \\
            WildVision & 32.7 & 32.0 & 30.8 & \textbf{41.7} \\
            RealWorldQA & 56.5 & 55.1 & \textbf{56.8} & 53.5 \\
            Avg & 46.8 & 47.3 &	48.4 &	\textbf{50.0} \\
        \bottomrule
    \end{tabular}
    \caption{Performance On Different Benchmarks Of Models Trained On Data Rewritten By Different Models}
    \label{tab:ablation_different_rewrite_model_detail}
\end{table*}

\begin{figure*}[t]
    \centering
    \includegraphics[width=0.7\textwidth]{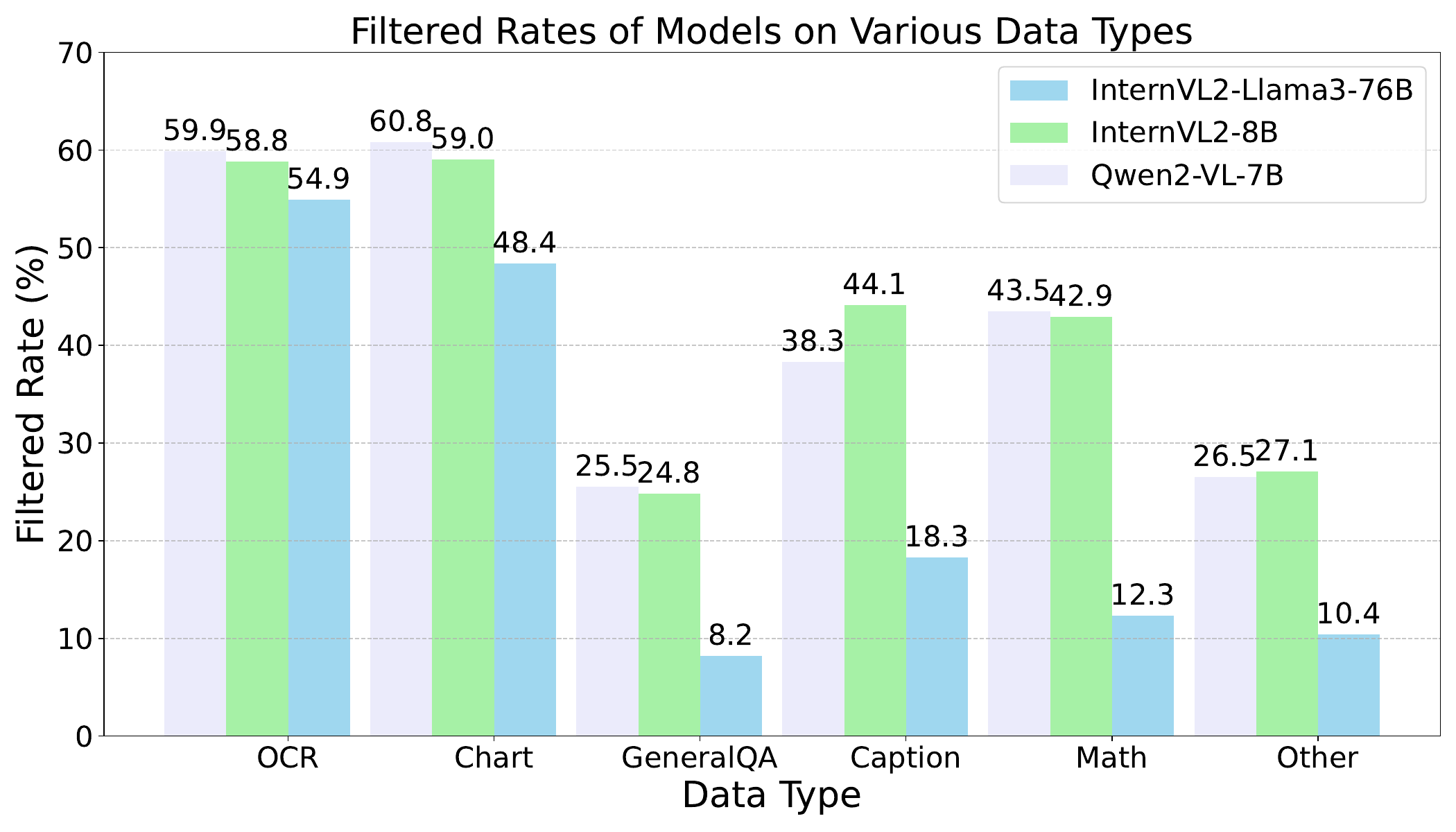}
    \caption{Different Filter Rates For Rewritten Data Filtered By Different Models On Different Data Types.}
    \label{fig:filter_rate_models}
\end{figure*}

~\autoref{tab:ablation_different_rewrite_model_detail} displays the performance on different benchmarks of models trained on data rewritten by different models.
~\autoref{fig:filter_rate_models} shows the filter rate of models on various data types.It can be seen that the retention rate of the data rewritten by larger model is much higher than that rewritten by models with smaller parameters, but it is worth noting that the retention rate of all three models in terms of ocr data is very low, which highlights the current problems facing multimodal model.

\newpage
\subsection{Additional Details of Verifying the consistency of model filter and manual filter}

\begin{table*}[!h]\centering
\begin{tabular}{@{}ccccc@{}}\toprule
\textbf{$\backslash$}  & Model & Evaluator1 & Evaluator2 & Evaluator3 \\
\midrule
Model  & -    & 0.73 & 0.70 & 0.63 \\
Evaluator1 & 0.73 & -    & 0.70 & 0.42 \\
Evaluator2 & 0.70 & 0.70 & -    & 0.53 \\
Evaluator3 & 0.63 & 0.42 & 0.53 & -    \\
\bottomrule
\end{tabular}
\caption{Kappa Value Between Any Two.}
\label{tab:consistency_of_model_user}
\end{table*}

~\autoref{tab:consistency_of_model_user} indicate that by averaging the Kappa values\cite{mchugh2012interrater}, we obtain an average pairwise Kappa value of 0.55 among three evaluators. When the model's filtering decision is used to replace one of the evaluators, the average Kappa of the three new derived evaluator combinations is 0.64 (\textit{eg}., calculate among Model, Evaluator1 and Evaluator2). Based on the Kappa value range (0.60 to 0.80 for good consistency), the agreement between model and manual judge is quite strong, suggesting that the process of model filtering is reliable and consistent with human judgment in this work.

\newpage
\section{Additional Details of Rewriting Step}

\subsection{Prompts For Rewriting}
\label{app:prompts_fot_rewrite}

\label{sec:prompt_used_in_data_filtering}

\begin{promptbox}[Prompt for Score]{darkblue}
You are given a Q\&A conversation and a corresponding image.
- Conversations: {}
Your goal is to:

1. Evaluate Information Content and Complexity: Rate the combined information richness and complexity of the Q\&A conversation and image on a scale from 1 to 5. Consider factors like the amount of detail, depth of content, and how well the conversation and image complement each other in conveying comprehensive information.

1: Minimal detail, shallow content, limited interaction between conversation and image.\\
2: Some detail, moderate depth, a basic connection between the conversation and image.\\
3: Good amount of detail, some complexity, moderate complementarity between conversation and image.\\
4: High level of detail, deep content, the conversation and image work well together to convey a thorough understanding.\\
5: Very rich in detail, highly complex, the conversation and image are seamlessly integrated to provide a comprehensive, insightful picture.\\

2. Evaluate Relevance: Rate the relevance of the conversation to the image on a scale from 1 to 5.

1: Very low relevance, the conversation and image are almost unrelated.\\
2: Low relevance, the conversation and image share some overlap, but one is mostly independent of the other.\\
3: Moderate relevance, there is some connection, but the conversation could stand alone without the image or vice versa.\\
4: High relevance, the conversation and image are closely tied, and both contribute significantly to each other's meaning.\\
5: Very high relevance, the conversation and image are inseparable, and one cannot fully be understood without the other.\\

Output Format: \\
1. Information Content Score (1-5): \\
2. Relevance (1-5): \\

Only provide the final result in the above structured format without any additional explanations.
\end{promptbox}

\begin{promptbox}[Rewrite Prompt for GeneralQA]{darkblue}
Your objective is to craft a series of advanced \textless instruction, response \textgreater pairs derived from a specific image and its associated question-answer pair. First, choose an appropriate task type from the list below as the basis for your rewrite:\\
  - 1. Coding \& Debugging\\
  - 2. Information Seeking\\
  - 3. Creative Writing\\
  - 4. Reasoning\\
  - 5. Planning\\
  - 6. Math\\
  - 7. Editing\\
  - 8. Data Analysis\\
  - 9. Role Playing\\
  - 10. Brainstorming\\
  - 11. Advice Seeking\\
\\
**Guidelines for New Instructions:**\\
- Each instruction should be substantially more intricate than the original, introducing complex and thought-provoking elements.\\
- Focus on deeper aspects of the original topic, incorporating critical thinking, detailed analysis, or creative problem-solving.\\
- Where possible, connect the topic with other related fields or practical applications.\\
\\
**Guidelines for Responses:**\\
- Responses must be detailed and comprehensive, covering all aspects of the instruction.\\
- Include diverse perspectives or approaches, and integrate examples, case studies, or hypothetical scenarios.\\
- Discuss potential implications, challenges, or future developments in relation to the topic.\\
\\
**Format:**\\
Each \textless instruction, response\textgreater pair should be formatted as follows:\\
  - \textless Instruction: [In-depth, sophisticated instruction linked to the initial Q\&A]\textgreater\\
  - \textless Response: [Extensive, expert-level response detailing the instruction's complexities]\textgreater\\
\\
Aim to generate at least {pair\_num} pairs, ensuring each set is unique and progressively builds upon the knowledge from the initial Q\&A. The goal is to elevate a simple concept into a sequence of intricate, expert-level discussions.\\
**Based on the following assets:**\\
- **Image:** \textless image\textgreater\\
- **Initial Q\&A:** {qa\_text}
\end{promptbox}

\begin{promptbox}[Rewrite Prompt for Chart]{darkblue}
Your objective is to craft a series of advanced \textless instruction, response\textgreater pairs derived from a specific image that is associated question-answer pair. The image could be a screenshot of table, scenetext images or other charts. First, choose an appropriate task type from the list below as the basis for your rewrite:\\
  - 1. Data Analysis\\
  
  - 2. Information Seeking\\
  
  - 3. Planning\\
  
  - 4. Editing\\
  
  - 5. Reasoning\\

Then you need to generate new \textless instruction, response\textgreater pairs should be progressively built upon the knowledge from the initial Q\&A. The goal is to elevate a simple concept into a sequence of intricate, expert-level discussions.

Guidelines for New Instructions:\\
- Each instruction should be substantially more intricate than the original, introducing complex and thought-provoking elements.\\
- Focus on deeper aspects of the original topic, incorporating critical thinking, detailed analysis, or creative problem-solving.\\
- Where possible, connect the topic with other related fields or practical applications.\\

Guidelines for Responses:\\
- Responses must be detailed and comprehensive, covering all aspects of the instruction.\\

- Include diverse perspectives or approaches, and integrate examples, case studies, or hypothetical scenarios.\\

- Discuss potential implications, challenges, or future developments in relation to the topic.\\

Format:\\
Each \textless instruction, response\textgreater pair should be formatted as follows and use the starting "\#\#Instruction\#\#:" and "\#\#Response\#\#:":\\
  - \#\#Instruction\#\#: [In-depth, sophisticated instruction linked to the initial Q\&A]\\
  - \#\#Response\#\#: [Extensive, expert-level response detailing the instruction's complexities]

Based on the following assets:\\
- **Image:** \textless image\textgreater\\
- **Initial Q\&A:** {qa\_text}
\end{promptbox}

\begin{promptbox}[Rewrite Prompt for Math]{darkblue}
Your task is to generate a detailed, step-by-step solution for a given math problem that includes an image. The answer provided is simple, but the goal is to fill in all the missing intermediate steps required to reach the final answer.\\

Please adhere to the following guidelines:\\

1. Response Crafting:\\
  - Provide a comprehensive response that includes all intermediate steps, explanations of the reasoning process, and any relevant mathematical concepts or operations used to solve the problem.\\
  - Ensure the response is exhaustive, covering each stage required to reach the final answer, while considering all details from the image.\\

2. Content Requirements:\\
  - The response should be well-structured, with clear steps and explanations that thoroughly address the instruction.\\
  - Do not include additional text or explanations outside of the required \textless response\textgreater.\\

\#\#\# Example:\\
Question: What is the distance between points B and C?\\
Simple Answer: The answer is 9.\\
Revised Answer: \textless response: To find the distance between points B and C', we start by analyzing the given geometric configuration... [continue with detailed steps].\textgreater\\

Question: {}\\
Simple Answer: {}\\
\end{promptbox}

\begin{promptbox}[Rewrite Prompt for Caption]{darkblue}
Your task is to generate a series of \textless instruction, response\textgreater pairs based on a given image caption while selecting one suitable task type from the following list for rewriting:\\
  - 1. Coding \& Debugging

  ......
  
  - 11. Advice Seeking\\

The goal is to create challenging tasks that require a deep understanding of the visual content described in the caption, without explicitly mentioning too many details from the caption itself.\\

Please adhere to the following guidelines:\\

1. Instruction Generation:\\
  - Select one task type from the above list and generate at least {} \textless instruction, response\textgreater pairs based on the chosen task.\\
  - Ensure each instruction is unique, complex, and related to the given caption and task type.\\
  - Avoid directly repeating specific details from the caption.\\
  - Instructions should require critical thinking, analysis, or creative problem-solving based on a clear understanding of the visual scene.\\

2. Response Crafting:\\
  - Provide detailed, comprehensive responses that demonstrate accurate interpretation of the visual information.\\
  - Include in-depth explanations, multiple perspectives, or detailed steps where appropriate.\\
  - Ensure responses clearly show an accurate grasp of the visual information implied by the instruction.\\

3. Content Requirements:\\
  - Instructions should be varied, challenging, and explore different advanced aspects or potential sophisticated interpretations of the visual scene.\\
  - Instructions should require the responder to infer and utilize visual information that may not be explicitly stated in the instruction.\\
  - Responses should demonstrate expertise in both understanding the implied visual content and providing substantial, valuable information related to the task.\\
  - Responses should be well-structured with clear sections or paragraphs.\\

4. Output Clarification:\\
  - Do not include additional text or explanations outside of the required \textless instruction, response\textgreater pairs.\\
  - Your output should only consist of the generated \textless instruction, response\textgreater pairs.\\

\#\#\# Example:\\
Caption: a street scene with construction scaffolding, three individuals, a shopping cart filled with personal belongings, street signs, and a sidewalk. The construction scaffolding is blue and has text about the construction company and contact details. One individual is walking by, another person is standing and looking at something in their hands, and the third person is behind a shopping cart going through the items inside. There are no vehicles visible in the image. The shopping cart filled with personal belongings might suggest that the individual using it is homeless or transient. The time of day appears to be daytime, but no direct indicators of weather conditions are present, although the presence of personal umbrellas might suggest rain.\\
Task Category: Information Seeking\\
\textless Instruction: Analyze the social dynamics at play in this street scene, taking into account the individuals' body language, interactions, and apparent activities. How do their actions and postures reveal or conceal their relationships with each other and their environment?\textgreater\\
\textless Response: Upon close examination, it appears that the individual walking by is intentionally avoiding eye contact with the other two individuals, suggesting a sense of detachment or disengagement. The person standing and looking at something in their hands seems engrossed in their own world, oblivious to the surroundings. Meanwhile, the person behind the shopping cart is intensely focused on rummaging through the items, indicating a sense of urgency or desperation. The lack of eye contact or acknowledgement among the three individuals implies a lack of community or social bonding in this urban setting. Furthermore, the absence of vehicles and the presence of construction scaffolding create an atmosphere of disruption and transition, which may be contributing to the individuals' isolation. The fact that one individual is potentially homeless or transient adds a layer of complexity to the social dynamics, as they may be grappling with issues of poverty, displacement, or marginalization.\textgreater\\

Caption: \{\}
\end{promptbox}

\begin{promptbox}[Rewrite Prompt for OCR]{darkblue}
I have an image and several corresponding Visual Question Answering (VQA) pairs. I'd like you to imagine a scenario and create a richer set of dialogues based on the information known from the VQA pairs. This dialogue set should include question-answering and instruction-based tasks. \\

The dialogue should revolve around the image, but the questions should not "spoil" too much information about the image content. This is because I intend to use this dialogue set to train a vision-language model, and excessive spoilers would make the learning process too easy. \\

In your output, please first provide a randomly imagined scenario (try to make it relevant to the image information), and then list the set of dialogues you've created. \\

When creating the dialogue: \\
1. Vary the types of questions and tasks (e.g., open-ended questions, specific queries, instructions) \\
2. Maintain a natural conversational flow \\
3. Include some challenging question or instruction that requires complex reasoning or analysis \\
4. Avoid directly stating key visual elements in the questions \\
5. Provide detailed and comprehensive responses for ALL answers \\
6. The AI assistant should act as if it's unaware of the scenario at the beginning of the conversation \\
7. Minimize the use of conversational fillers or emotional language in questions. The AI assistant's tone should be neutral and professional \\
8. The AI assistant should strive to be as helpful as possible, answering questions to the best of its ability based on the given information \\
9. The human should not ask questions that go beyond the scope of the information provided in the VQA pairs \\

The questioner is a curious human seeking help, and the answerer is an AI assistant capable of perceiving and analyzing images. \\

Format requirement: \\
Please format the dialogue as a series of alternating messages, with at most 5 rounds of conversation: \\

Scenario: [The scenario related to the image.] \\
Human: [First question, without conversational fillers] \\
Assistant: [Detailed and comprehensive answer to the first question, based only on VQA information] \\
Human: [Second question or instruction, without conversational fillers] \\
Assistant: [Detailed and comprehensive response to the second question or instruction, based only on VQA information] \\
...

Each message should be on a new line, clearly indicating the speaker. 
Do not mention the human or assistant's name in the dialogue. 
Ensure that all responses from the Assistant are detailed and comprehensive, providing as much relevant information as possible based on the VQA pairs.\\

Please create this rich dialogue set based on the brief VQA pairs I will provide. Do not include any additional content or explanations outside of the specified output format.\\

VQA: {vqa} \\

\end{promptbox}

\begin{promptbox}[Prompt for Data Filtering]{darkblue}
\label{promptbox:data_filtering}
**Q\&A:** \{\}\\\\
Evaluate the provided question and answer based on the given image. Specifically:

1. **Relevance**: Determine whether the question is directly related to the content of the image.  \\
2. **Consistency**: Verify if the answer is logically coherent and aligns with the question.  \\
3. **Accuracy**: Confirm whether both the question and answer accurately reflect the details or facts depicted in the image.  \\

Provide your response as follows: \\ 
- Respond with **'Yes'** if the question and answer are relevant, consistent, and factually accurate. \\ 
- Respond with **'No'** if there are any issues with relevance, logical coherence, or factual accuracy.\\

Only respond with a single word Yes or No.
\end{promptbox}

\subsection{Score Details Before and After Rewriting}

~\autoref{table:scores} shows the comparison of Original and Rewritten data on Content and Relevance Scores. Rewritten data scores higher, indicating improved quality.

\begin{table*}[h]
\centering
\setlength{\tabcolsep}{4pt}
\scalebox{0.9}{
\begin{tabular}{@{}lcc@{}}
\toprule
\textbf{Source} & \makecell{\textbf{Original Content\&}\\\textbf{Relevance Score}} & \makecell{\textbf{Rewrite Content\&}\\\textbf{Relevance Score}}  \\
\midrule
chrome-writting	& 3.0/3.4 & 4.0/4.7 \\
OCRVQA &	3.9/4.6 &	3.9/4.5 \\
screen-qa &	3.4/4.2	& 4.0/4.3 \\
hme &	3.3/4.3 &	4.0/4.8\\
textvqa &	3.2/4.1 &	4.0/4.5 \\
docvqa & 3.9/4.8 & 4.0/4.8 \\
st-vqa(cauldron,llava-format) &	3.0/3.8 & 4/4.4 \\
Ureader Chart &	4.0/4.8 &	4.0/4.8 \\
infographic-vqa & 4.4/4.9 & 4.2/4.8 \\
finqa  & 4.0/4.3 & 3.9/4.1 \\
ureader-kg & 4.0/4.7 &	4.1/4.7 \\
chartqa(cauldron,llava-format) &	4.0/4.9 & 4.0/4.9\\
vistext(cauldron) &	4.0/4.8 &	4.0/4.8 \\
chart2text(cauldron) &	4.0/4.7 &4.0/4.7 \\
ureader-qa &	4.0/4.7 &	4.1/4.7 \\
Irv-chart &	4.0/4.9	 & 4.0/4.8 \\
idefics375k	& 3.6/4.4 & 3.9/4.5 \\
cambrian(filtered) &	3.2/3.6 &	3.8/4.1 \\
GQA &	3.2/4.0 &	3.8/4.1 \\
AlfWorld &	3.0/3.6 & 3.2/3.8 \\
IDK &	3.0/3.6 & 3.8/4.2\\
CLlava Instruct &	3.7/4.1 &	3.9/4.1 \\
llava-zh &	3.5/4.1 &	3.9/4.1 \\
SVITCore &	3.7/4.1 &	3.9/4.1 \\
SVITCore mix &	3.4/4.1	& 3.9/4.1 \\
visual7w(cauldron,llava-format) &	3.1/4.0 &3.9/4.2\\
sharegpt4v &	3.9/4.2	 & 3.7/3.7 \\
infographic(gpt4v) &	4.3/5.0 & 4.1/4.5\\
sharegpt4o & 4.0/4.4 &	3.8/3.9 \\
sharegpt4v(coco) &	3.9/4.0 &	3.7/3.7 \\
sharegpt4v(llava) &	3.7/3.9 &	3.7/3.8 \\
sharegpt4v(sam) &	3.9 / 3.9 &	3.8/3.7 \\
Geo170K &	3.9/4.7 & 4.0/4.8 \\
MathVision &	3.7/4.7 & 4.0/4.9 \\
CLEVR-Math(MathV360K) &	3.0/3.8 &	3.1/3.8\\
GEOS(MathV360K) &	3.8/4.8	& 4.0/4.9 \\
GeoQA+(MathV360K)&	3.9/4.9 & 4.0/4.9 \\
Geometry3K(MathV360K)&	3.9/4.9 &	4/4.9 \\
IconQA(MathV360K) &	2.7/4.2 &	3.0/4.2 \\
PMC-VQA(MathV360K) &	3.9/4.3	& 4.0/4.6 \\
Super-CLEVR(MathV360K)&	3.4/4.0  &	3.3/4.0\\
TabMWP(MathV360K)&	3.3/4.2 &	3.7/4.7 \\
UniGeo(MathV360K) &	3.9/4.9 &	4.0/4.9 \\
VizWiz(MathV360K) &	3.0/4.0 &	3.5/4.3\\
MapQA(MathV360K) &	3.0/2.4 & 3.9/4.3 \\
raven(cauldron) &	3.3/4.2 &	3.9/4.6 \\
M3IT+FLAN &	3.2/4.0 & 3.4/4.1\\
WIT &	3.9/4.5	& 4.1/4.7 \\
ViQuAE &	3.3/3.8 &	3.9/4.0 \\
aokvqa(cauldron,llava-format) &	3.0/3.7 & 3.6/4.1 \\
vision-flan(filtered) &	3.0/3.7	& 3.4/4.0 \\
websight(cauldron) &	3.7/4.4 & 3.8/4.4\\
vsr(cauldron,llava-format) &	3.0/3.5 & 3.0/3.7 \\
CLEVR &	3.0/3.8 & 3.2/3.9 \\
tallyqa(cauldron,llava-format) & 2.9/3.7 & 3.1/4.0 \\
scienceqa(nona-context)& 3.4/4.0 & 3.5/4.1 \\
PathVQA (164) &	3.3/3.6 & 4.0/4.2\\
tqa(cauldron,llava-format) & 2.6/3.6 &2.8/3.9 \\
vqarad(cauldron,llava-format) &	4.0/4.7 &	4.0/4.7 \\
\midrule
\textbf{Average} & 3.5 / 4.2 & 3.8 / 4.4 \\
\bottomrule
\end{tabular}
}
\caption{Comparison of Original and Rewrite Average Content and Relevance Scores}
\label{table:scores}
\end{table*}




\clearpage

\begin{figure*}[htp]
\centering
\begin{minipage}{.01\textwidth}
\end{minipage}%
\begin{minipage}{.99\textwidth}
\centering
\renewcommand{\arraystretch}{1.1}
\setlength\tabcolsep{1pt}
\fontsize{9pt}{12pt}\selectfont
\begin{tabular}{lll}
\makecell[l]{\cellcolor[RGB]{234,142,178} \textcolor{white}{Group A(37.9\%)}} &
\tikz[baseline=0.05em] \fill [color={rgb,255:red,234; green,142; blue,178}] (0,0) rectangle (0.75em,0.75em); \begin{tabular}{l} IAM(cauldron) \\ {\footnotesize \cite{Marti2002IAM}} \end{tabular} &
\tikz[baseline=0.05em] \fill [color={rgb,255:red,234; green,141; blue,168}] (0,0) rectangle (0.75em,0.75em); \begin{tabular}{l} Ureader OCR \\ {\footnotesize \cite{ye2023ureader}} \end{tabular} \\

\tikz[baseline=0.05em] \fill [color={rgb,255:red,234; green,140; blue,158}] (0,0) rectangle (0.75em,0.75em); \begin{tabular}{l} ReCTs \\ {\footnotesize \cite{liu2019icdar}} \end{tabular} &
\tikz[baseline=0.05em] \fill [color={rgb,255:red,234; green,139; blue,148}] (0,0) rectangle (0.75em,0.75em); \begin{tabular}{l} IIIT 5K \\ {\footnotesize \cite{MishraBMVC12}} \end{tabular} &
\tikz[baseline=0.05em] \fill [color={rgb,255:red,234; green,138; blue,138}] (0,0) rectangle (0.75em,0.75em); \begin{tabular}{l} llavar-gpt4-20k \\ {\footnotesize \cite{li2024llava}} \end{tabular} \\

\tikz[baseline=0.05em] \fill [color={rgb,255:red,234; green,137; blue,128}] (0,0) rectangle (0.75em,0.75em); \begin{tabular}{l} RenderedText(cauldron) \\ {\footnotesize \cite{WendlerRenderedText}} \end{tabular} &
\tikz[baseline=0.05em] \fill [color={rgb,255:red,234; green,136; blue,118}] (0,0) rectangle (0.75em,0.75em); \begin{tabular}{l} TextOCR(cleaned) \\ {\footnotesize \cite{singh2021textocr}} \end{tabular} &
\tikz[baseline=0.05em] \fill [color={rgb,255:red,234; green,126; blue,108}] (0,0) rectangle (0.75em,0.75em); \begin{tabular}{l} Visualmrc(cauldron) \\ {\footnotesize \cite{tanaka2021visualmrc}} \end{tabular} \\

\tikz[baseline=0.05em] \fill [color={rgb,255:red,234; green,116; blue,98}] (0,0) rectangle (0.75em,0.75em); \begin{tabular}{l} ArxivQA \\ {\footnotesize \cite{li-etal-2024-multimodal-arxiv}} \end{tabular} &
\tikz[baseline=0.05em] \fill [color={rgb,255:red,234; green,106; blue,88}] (0,0) rectangle (0.75em,0.75em); \begin{tabular}{l} mPLUG-DocOwlchart \\ {\footnotesize \cite{hu-etal-2024-mplug}} \end{tabular} &
\tikz[baseline=0.05em] \fill [color={rgb,255:red,234; green,96; blue,88}] (0,0) rectangle (0.75em,0.75em); \begin{tabular}{l} Diagram-Image-to-Text(cauldron) \\ {\footnotesize \cite{li2024llava}} \end{tabular} \\

\tikz[baseline=0.05em] \fill [color={rgb,255:red,234; green,86; blue,88}] (0,0) rectangle (0.75em,0.75em); \begin{tabular}{l} DVQA \\ {\footnotesize \cite{kafle2018dvqa}} \end{tabular} &
\tikz[baseline=0.05em] \fill [color={rgb,255:red,234; green,76; blue,78}] (0,0) rectangle (0.75em,0.75em); \begin{tabular}{l} RobuT-WTQ(cauldron,llava-format) \\ {\footnotesize \cite{zhao-etal-2023-robut}} \end{tabular} &
\tikz[baseline=0.05em] \fill [color={rgb,255:red,234; green,66; blue,68}] (0,0) rectangle (0.75em,0.75em); \begin{tabular}{l} RobuT-SQA(cauldron) \\ {\footnotesize \cite{zhao-etal-2023-robut}} \end{tabular} \\

\tikz[baseline=0.05em] \fill [color={rgb,255:red,234; green,56; blue,58}] (0,0) rectangle (0.75em,0.75em); \begin{tabular}{l} HiTab(cauldron,llava-format) \\ {\footnotesize \cite{cheng-etal-2022-hitab}} \end{tabular} &
\tikz[baseline=0.05em] \fill [color={rgb,255:red,234; green,46; blue,48}] (0,0) rectangle (0.75em,0.75em); \begin{tabular}{l} TobuT-WikiSQL(cauldron) \\ {\footnotesize \cite{zhao-etal-2023-robut}} \end{tabular} &
\tikz[baseline=0.05em] \fill [color={rgb,255:red,234; green,105; blue,38}] (0,0) rectangle (0.75em,0.75em); \begin{tabular}{l} TAT-QA \\ {\footnotesize \cite{zhu2021tat}} \end{tabular} \\

\tikz[baseline=0.05em] \fill [color={rgb,255:red,234; green,125; blue,28}] (0,0) rectangle (0.75em,0.75em); \begin{tabular}{l} PlotQA \\ {\footnotesize \cite{methani2020plotqa}} \end{tabular} &
\tikz[baseline=0.05em] \fill [color={rgb,255:red,234; green,85; blue,18}] (0,0) rectangle (0.75em,0.75em); \begin{tabular}{l} WildVision-Chat \\ {\footnotesize \cite{lu2024wildvisionevaluatingvisionlanguagemodels}} \end{tabular} &
\tikz[baseline=0.05em] \fill [color={rgb,255:red,234; green,105; blue,8}] (0,0) rectangle (0.75em,0.75em); \begin{tabular}{l} ALLaVA-Laion \\ {\footnotesize \cite{chen2024allava}} \end{tabular} \\

\tikz[baseline=0.05em] \fill [color={rgb,255:red,234; green,125; blue,48}] (0,0) rectangle (0.75em,0.75em); \begin{tabular}{l} ALLaVA-Vflan \\ {\footnotesize \cite{chen2024allava}} \end{tabular} &
\tikz[baseline=0.05em] \fill [color={rgb,255:red,234; green,85; blue,8}] (0,0) rectangle (0.75em,0.75em); \begin{tabular}{l} LVIS-InstructV4 \\ {\footnotesize \cite{wang2023see}} \end{tabular} &
\tikz[baseline=0.05em] \fill [color={rgb,255:red,234; green,105; blue,28}] (0,0) rectangle (0.75em,0.75em); \begin{tabular}{l} ShareGPT4V(new) \\ {\footnotesize (\textcolor{blue}{Self Collection})} \end{tabular} \\

\tikz[baseline=0.05em] \fill [color={rgb,255:red,234; green,85; blue,8}] (0,0) rectangle (0.75em,0.75em); \begin{tabular}{l} Laion-GPT4V \\ {\footnotesize \cite{chen2024allava}} \end{tabular} &
\tikz[baseline=0.05em] \fill [color={rgb,255:red,234; green,105; blue,28}] (0,0) rectangle (0.75em,0.75em); \begin{tabular}{l} Sherlock \\ {\footnotesize \cite{hessel2022abduction}} \end{tabular} &
\tikz[baseline=0.05em] \fill [color={rgb,255:red,234; green,125; blue,48}] (0,0) rectangle (0.75em,0.75em); \begin{tabular}{l} ShareGPT4V \\ {\footnotesize (\textcolor{blue}{Self Collection})} \end{tabular} \\

\tikz[baseline=0.05em] \fill [color={rgb,255:red,234; green,85; blue,8}] (0,0) rectangle (0.75em,0.75em); \begin{tabular}{l} ALLaVA-4V-Chinese(vflan) \\ {\footnotesize \cite{chen2024allava}} \end{tabular} &
\tikz[baseline=0.05em] \fill [color={rgb,255:red,234; green,105; blue,28}] (0,0) rectangle (0.75em,0.75em); \begin{tabular}{l} Irv-Normal(filtered) \\ {\footnotesize \cite{liu2023aligning}} \end{tabular} &
\tikz[baseline=0.05em] \fill [color={rgb,255:red,234; green,125; blue,48}] (0,0) rectangle (0.75em,0.75em); \begin{tabular}{l} ALLaVA-4V-Chinese(Laion) \\ {\footnotesize \cite{chen2024allava}} \end{tabular} \\

\tikz[baseline=0.05em] \fill [color={rgb,255:red,234; green,85; blue,8}] (0,0) rectangle (0.75em,0.75em); \begin{tabular}{l} Mavis-Math-Rule-Geo \\ {\footnotesize \cite{zhang2024mavismathematicalvisualinstruction}} \end{tabular} &
\tikz[baseline=0.05em] \fill [color={rgb,255:red,234; green,105; blue,28}] (0,0) rectangle (0.75em,0.75em); \begin{tabular}{l} Geo3k \\ {\footnotesize \cite{lu2021inter}} \end{tabular} &
\tikz[baseline=0.05em] \fill [color={rgb,255:red,234; green,125; blue,48}] (0,0) rectangle (0.75em,0.75em); \begin{tabular}{l} Mavis-Math-Metagen \\ {\footnotesize \cite{zhang2024mavismathematicalvisualinstruction}} \end{tabular} \\

\tikz[baseline=0.05em] \fill [color={rgb,255:red,234; green,85; blue,8}] (0,0) rectangle (0.75em,0.75em); \begin{tabular}{l} TabMWP(cauldron) \\ {\footnotesize \cite{lu2023dynamic}} \end{tabular} &
\tikz[baseline=0.05em] \fill [color={rgb,255:red,234; green,105; blue,28}] (0,0) rectangle (0.75em,0.75em); \begin{tabular}{l} AI2D(gpt4v) \\ {\footnotesize \cite{kembhavi2016diagram}} \end{tabular} &
\tikz[baseline=0.05em] \fill [color={rgb,255:red,234; green,125; blue,48}] (0,0) rectangle (0.75em,0.75em); \begin{tabular}{l} Geomverse(cauldron) \\ {\footnotesize \cite{kazemi2024geomverse}} \end{tabular} \\

\tikz[baseline=0.05em] \fill [color={rgb,255:red,234; green,125; blue,48}] (0,0) rectangle (0.75em,0.75em); \begin{tabular}{l} ShareGPT4V(knowledge) \\ {\footnotesize \cite{chen2025sharegpt4v}} \end{tabular} &
\tikz[baseline=0.05em] \fill [color={rgb,255:red,234; green,85; blue,8}] (0,0) rectangle (0.75em,0.75em); \begin{tabular}{l} Design2Code \\ {\footnotesize \cite{si2024design2code}} \end{tabular} & 
\tikz[baseline=0.05em] \fill [color={rgb,255:red,234; green,125; blue,48}] (0,0) rectangle (0.75em,0.75em); \begin{tabular}{l} MultiUI \\ {\footnotesize \cite{liu2024harnessing}} \end{tabular} \\

\tikz[baseline=0.05em] \fill [color={rgb,255:red,234; green,125; blue,48}] (0,0) rectangle (0.75em,0.75em); \begin{tabular}{l} Arxiv-Chart-4o \\ {\footnotesize (\textcolor{blue}{Self Collection})} \end{tabular} &
\tikz[baseline=0.05em] \fill [color={rgb,255:red,234; green,125; blue,48}] (0,0) rectangle (0.75em,0.75em); \begin{tabular}{l} NLVR2 \\ {\footnotesize \cite{suhr2019corpus}} \end{tabular} &
\tikz[baseline=0.05em] \fill [color={rgb,255:red,234; green,85; blue,8}] (0,0) rectangle (0.75em,0.75em); \begin{tabular}{l} Mimic CGD \\ {\footnotesize \cite{li2023mimic}} \end{tabular} \\

\tikz[baseline=0.05em] \fill [color={rgb,255:red,234; green,125; blue,48}] (0,0) rectangle (0.75em,0.75em); \begin{tabular}{l} Coinstruct \\ {\footnotesize \cite{wu2025towards}} \end{tabular} &
\tikz[baseline=0.05em] \fill [color={rgb,255:red,234; green,125; blue,48}] (0,0) rectangle (0.75em,0.75em); \begin{tabular}{l} HQ-Edit \\ {\footnotesize \cite{hui2024hq}} \end{tabular} &
\tikz[baseline=0.05em] \fill [color={rgb,255:red,234; green,85; blue,8}] (0,0) rectangle (0.75em,0.75em); \begin{tabular}{l} Raven \\ {\footnotesize \cite{zhang2019raven}} \end{tabular} \\

\tikz[baseline=0.05em] \fill [color={rgb,255:red,234; green,125; blue,48}] (0,0) rectangle (0.75em,0.75em); \begin{tabular}{l} IconQA \\ {\footnotesize \cite{lu2021iconqa}} \end{tabular} &
\tikz[baseline=0.05em] \fill [color={rgb,255:red,234; green,125; blue,48}] (0,0) rectangle (0.75em,0.75em); \begin{tabular}{l} VIST \\ {\footnotesize \cite{huang2016visual}} \end{tabular} &
\tikz[baseline=0.05em] \fill [color={rgb,255:red,234; green,85; blue,8}] (0,0) rectangle (0.75em,0.75em); \begin{tabular}{l} Contrast-Caption \\ {\footnotesize \cite{Jiang2024MANTISIM}} \end{tabular} \\ 

\tikz[baseline=0.05em] \fill [color={rgb,255:red,234; green,125; blue,48}] (0,0) rectangle (0.75em,0.75em); \begin{tabular}{l} FlintstonesSV \\ {\footnotesize \cite{gupta2018imagine}} \end{tabular} &
\tikz[baseline=0.05em] \fill [color={rgb,255:red,234; green,125; blue,48}] (0,0) rectangle (0.75em,0.75em); \begin{tabular}{l} PororoSV \\ {\footnotesize \cite{li2019storygan}} \end{tabular} &
\tikz[baseline=0.05em] \fill [color={rgb,255:red,234; green,85; blue,8}] (0,0) rectangle (0.75em,0.75em); \begin{tabular}{l} LLaVA-Video \\ {\footnotesize \cite{zhang2024videoinstructiontuningsynthetic}} \end{tabular} \\ 

\tikz[baseline=0.05em] \fill [color={rgb,255:red,234; green,125; blue,48}] (0,0) rectangle (0.75em,0.75em); \begin{tabular}{l} M4 Instruct Video \\ {\footnotesize \cite{li2024llavanextinterleavetacklingmultiimagevideo}} \end{tabular} &
\tikz[baseline=0.05em] \fill [color={rgb,255:red,234; green,125; blue,48}] (0,0) rectangle (0.75em,0.75em); \begin{tabular}{l} LLaVA-Video-ActivityNetQA \\ {\footnotesize \cite{zhang2024videoinstructiontuningsynthetic}} \end{tabular} &
\tikz[baseline=0.05em] \fill [color={rgb,255:red,234; green,85; blue,8}] (0,0) rectangle (0.75em,0.75em); \begin{tabular}{l} LLaVA-Hound \\ {\footnotesize \cite{zhang2024direct}} \end{tabular} \\

\tikz[baseline=0.05em] \fill [color={rgb,255:red,234; green,125; blue,48}] (0,0) rectangle (0.75em,0.75em); \begin{tabular}{l} LLaVA-Video-NeXT-QA \\ {\footnotesize \cite{zhang2024videoinstructiontuningsynthetic}} \end{tabular} &
\tikz[baseline=0.05em] \fill [color={rgb,255:red,234; green,125; blue,48}] (0,0) rectangle (0.75em,0.75em); \begin{tabular}{l} VideoChatGPT \\ {\footnotesize \cite{maaz-etal-2024-video}} \end{tabular} &
\tikz[baseline=0.05em] \fill [color={rgb,255:red,234; green,85; blue,8}] (0,0) rectangle (0.75em,0.75em); \begin{tabular}{l} Video-MME \\ {\footnotesize \cite{fu2024video}} \end{tabular} \\ 

\tikz[baseline=0.05em] \fill [color={rgb,255:red,234; green,125; blue,48}] (0,0) rectangle (0.75em,0.75em); \begin{tabular}{l} LLaVA-Video-PerceptionTest \\ {\footnotesize \cite{zhang2024videoinstructiontuningsynthetic}} \end{tabular} &
\tikz[baseline=0.05em] \fill [color={rgb,255:red,234; green,125; blue,48}] (0,0) rectangle (0.75em,0.75em); \begin{tabular}{l} EgoSchema \\ {\footnotesize \cite{mangalam2023egoschema}} \end{tabular} & \\

\end{tabular}
\end{minipage}

\captionsetup{font={small}}
\caption{
\model data source group. Group A: datasets that are directly kept without modification.
}
\vspace{-5pt}
\label{fig:group_abc_a}
\end{figure*}

\begin{figure*}[htp]
\centering
\begin{minipage}{.01\textwidth}
\end{minipage}%
\begin{minipage}{.99\textwidth}
\centering
\renewcommand{\arraystretch}{1.1}
\setlength\tabcolsep{1pt}
\fontsize{9pt}{12pt}\selectfont
\begin{tabular}{lll}
\makecell[l]{\cellcolor[RGB]{193,149,245} \textcolor{white}{Group B(39.2\%)}} &
\tikz[baseline=0.05em] \fill [color={rgb,255:red,193; green,149; blue,245}] (0,0) rectangle (0.75em,0.75em); \begin{tabular}{l} Chrome-Writing \\ {\footnotesize \cite{li2024llava}} \end{tabular} &
\tikz[baseline=0.05em] \fill [color={rgb,255:red,199; green,159; blue,234}] (0,0) rectangle (0.75em,0.75em); \begin{tabular}{l} OCRVQA \\ {\footnotesize \cite{OCR-VQA}} \end{tabular} \\

\tikz[baseline=0.05em] \fill [color={rgb,255:red,204; green,168; blue,247}] (0,0) rectangle (0.75em,0.75em); \begin{tabular}{l} ScreenQA \\ {\footnotesize \cite{hsiao2024screenqalargescalequestionanswerpairs}} \end{tabular} &
\tikz[baseline=0.05em] \fill [color={rgb,255:red,184; green,132; blue,234}] (0,0) rectangle (0.75em,0.75em); \begin{tabular}{l} HME \\ {\footnotesize \cite{yuan2022syntax}} \end{tabular} &
\tikz[baseline=0.05em] \fill [color={rgb,255:red,209; green,167; blue,241}] (0,0) rectangle (0.75em,0.75em); \begin{tabular}{l} TextVQA \\ {\footnotesize \cite{textvqa}} \end{tabular} \\

\tikz[baseline=0.05em] \fill [color={rgb,255:red,198; green,156; blue,236}] (0,0) rectangle (0.75em,0.75em); \begin{tabular}{l} DocVQA(llava) \\ {\footnotesize \cite{DocVQA}} \end{tabular} &
\tikz[baseline=0.05em] \fill [color={rgb,255:red,210; green,173; blue,242}] (0,0) rectangle (0.75em,0.75em); \begin{tabular}{l} ST-VQA(cauldron,llava-format) \\ {\footnotesize \cite{STVQA}} \end{tabular} &
\tikz[baseline=0.05em] \fill [color={rgb,255:red,215; green,180; blue,243}] (0,0) rectangle (0.75em,0.75em); \begin{tabular}{l} Ureader Chart \\ {\footnotesize \cite{ye-etal-2023-ureader}} \end{tabular} \\

\tikz[baseline=0.05em] \fill [color={rgb,255:red,221; green,189; blue,248}] (0,0) rectangle (0.75em,0.75em); \begin{tabular}{l} Infographic-VQA \\ {\footnotesize \cite{mathew2021infographicvqa}} \end{tabular} &
\tikz[baseline=0.05em] \fill [color={rgb,255:red,199; green,161; blue,235}] (0,0) rectangle (0.75em,0.75em); \begin{tabular}{l} FinQA \\ {\footnotesize \cite{FinQA}} \end{tabular} &
\tikz[baseline=0.05em] \fill [color={rgb,255:red,190; green,150; blue,233}] (0,0) rectangle (0.75em,0.75em); \begin{tabular}{l} Ureader-KG \\ {\footnotesize \cite{ye-etal-2023-ureader}} \end{tabular} \\

\tikz[baseline=0.05em] \fill [color={rgb,255:red,211; green,175; blue,244}] (0,0) rectangle (0.75em,0.75em); \begin{tabular}{l} ChartQA(cauldron,llava-format) \\ {\footnotesize \cite{ChartQA}} \end{tabular} &
\tikz[baseline=0.05em] \fill [color={rgb,255:red,209; green,173; blue,241}] (0,0) rectangle (0.75em,0.75em); \begin{tabular}{l} VisText(cauldron) \\ {\footnotesize \cite{VisText}} \end{tabular} &
\tikz[baseline=0.05em] \fill [color={rgb,255:red,214; green,177; blue,243}] (0,0) rectangle (0.75em,0.75em); \begin{tabular}{l} Chart2Text(cauldron) \\ {\footnotesize \cite{Chart2Text}} \end{tabular} \\

\tikz[baseline=0.05em] \fill [color={rgb,255:red,220; green,187; blue,247}] (0,0) rectangle (0.75em,0.75em); \begin{tabular}{l} Ureader-QA \\ {\footnotesize \cite{ye-etal-2023-ureader}} \end{tabular} &
\tikz[baseline=0.05em] \fill [color={rgb,255:red,205; green,168; blue,246}] (0,0) rectangle (0.75em,0.75em); \begin{tabular}{l} Irv-chart \\ {\footnotesize \cite{liu2024mitigatinghallucinationlargemultimodal}} \end{tabular} &
\tikz[baseline=0.05em] \fill [color={rgb,255:red,203; green,160; blue,240}] (0,0) rectangle (0.75em,0.75em); \begin{tabular}{l} Idefics \\ {\footnotesize \cite{tong2024cambrian}} \end{tabular} \\

\tikz[baseline=0.05em] \fill [color={rgb,255:red,202; green,163; blue,239}] (0,0) rectangle (0.75em,0.75em); \begin{tabular}{l} Cambrian(filtered) \\ {\footnotesize \cite{tong2024cambrian}} \end{tabular} &
\tikz[baseline=0.05em] \fill [color={rgb,255:red,198; green,157; blue,234}] (0,0) rectangle (0.75em,0.75em); \begin{tabular}{l} GQA \\ {\footnotesize \cite{hudson2019gqa}} \end{tabular} &
\tikz[baseline=0.05em] \fill [color={rgb,255:red,196; green,153; blue,232}] (0,0) rectangle (0.75em,0.75em); \begin{tabular}{l} AlfWorld \\ {\footnotesize \cite{tong2024cambrian}} \end{tabular} \\

\tikz[baseline=0.05em] \fill [color={rgb,255:red,212; green,178; blue,245}] (0,0) rectangle (0.75em,0.75em); \begin{tabular}{l} IDK \\ {\footnotesize \cite{cha2024visually}} \end{tabular} &
\tikz[baseline=0.05em] \fill [color={rgb,255:red,219; green,184; blue,248}] (0,0) rectangle (0.75em,0.75em); \begin{tabular}{l} CLlava Instruct \\ {\footnotesize} Self Collection \end{tabular} &
\tikz[baseline=0.05em] \fill [color={rgb,255:red,222; green,191; blue,250}] (0,0) rectangle (0.75em,0.75em); \begin{tabular}{l} LLaVA-zh \\ {\footnotesize \cite{openbmb2023llava}} \end{tabular} \\

\tikz[baseline=0.05em] \fill [color={rgb,255:red,223; green,192; blue,251}] (0,0) rectangle (0.75em,0.75em); \begin{tabular}{l} SVITCore \\ {\footnotesize \cite{zhao2023svit}} \end{tabular} &
\tikz[baseline=0.05em] \fill [color={rgb,255:red,218; green,183; blue,247}] (0,0) rectangle (0.75em,0.75em); \begin{tabular}{l} SVITCore-mix \\ {\footnotesize \cite{zhao2023svit}} \end{tabular} &
\tikz[baseline=0.05em] \fill [color={rgb,255:red,215; green,177; blue,244}] (0,0) rectangle (0.75em,0.75em); \begin{tabular}{l} Visual7W(cauldron,llava-format) \\ {\footnotesize \cite{Visual7w}} \end{tabular} \\

\tikz[baseline=0.05em] \fill [color={rgb,255:red,211; green,172; blue,241}] (0,0) rectangle (0.75em,0.75em); \begin{tabular}{l} ShareGPT4v \\ {\footnotesize \cite{chen2023sharegpt4v}} \end{tabular} &
\tikz[baseline=0.05em] \fill [color={rgb,255:red,222; green,188; blue,249}] (0,0) rectangle (0.75em,0.75em); \begin{tabular}{l} Infographic \\ {\footnotesize \cite{mathew2021infographicvqa}} \end{tabular} &
\tikz[baseline=0.05em] \fill [color={rgb,255:red,221; green,183; blue,248}] (0,0) rectangle (0.75em,0.75em); \begin{tabular}{l} ShareGPT4o \\ {\footnotesize \cite{sharegpt4o2023}} \end{tabular} \\

\tikz[baseline=0.05em] \fill [color={rgb,255:red,216; green,179; blue,245}] (0,0) rectangle (0.75em,0.75em); \begin{tabular}{l} ShareGPT4V(COCO) \\ {\footnotesize \cite{li2024llava}} \end{tabular} &
\tikz[baseline=0.05em] \fill [color={rgb,255:red,219; green,182; blue,247}] (0,0) rectangle (0.75em,0.75em); \begin{tabular}{l} ShareGPT4V(LLAVA) \\ {\footnotesize \cite{li2024llava}} \end{tabular} &
\tikz[baseline=0.05em] \fill [color={rgb,255:red,215; green,178; blue,243}] (0,0) rectangle (0.75em,0.75em); \begin{tabular}{l} ShareGPT4V(SAM) \\ {\footnotesize \cite{li2024llava}} \end{tabular} \\

\tikz[baseline=0.05em] \fill [color={rgb,255:red,209; green,173; blue,242}] (0,0) rectangle (0.75em,0.75em); \begin{tabular}{l} Geo \\ {\footnotesize \cite{gao2023gllavasolvinggeometricproblem}} \end{tabular} &
\tikz[baseline=0.05em] \fill [color={rgb,255:red,204; green,169; blue,240}] (0,0) rectangle (0.75em,0.75em); \begin{tabular}{l} MathVision \\ {\footnotesize \cite{wang2024measuringmultimodalmathematicalreasoning}} \end{tabular} &
\tikz[baseline=0.05em] \fill [color={rgb,255:red,213; green,177; blue,243}] (0,0) rectangle (0.75em,0.75em); \begin{tabular}{l} CLEVR-Math(MathV360K) \\ {\footnotesize \cite{CLEVR-Math}} \end{tabular} \\

\tikz[baseline=0.05em] \fill [color={rgb,255:red,202; green,163; blue,239}] (0,0) rectangle (0.75em,0.75em); \begin{tabular}{l} GEOS(MathV360K) \\ {\footnotesize \cite{seo-etal-2015-solving}} \end{tabular} &
\tikz[baseline=0.05em] \fill [color={rgb,255:red,220; green,184; blue,248}] (0,0) rectangle (0.75em,0.75em); \begin{tabular}{l} GeoQA+(MathV360K) \\ {\footnotesize \cite{chen-etal-2021-geoqa}} \end{tabular} &
\tikz[baseline=0.05em] \fill [color={rgb,255:red,193; green,149; blue,245}] (0,0) rectangle (0.75em,0.75em); \begin{tabular}{l} Geometry3K (MathV360K) \\ {\footnotesize \cite{lu2021inter}} \end{tabular} \\

\tikz[baseline=0.05em] \fill [color={rgb,255:red,199; green,159; blue,234}] (0,0) rectangle (0.75em,0.75em); \begin{tabular}{l} IconQA (MathV360K) \\ {\footnotesize \cite{lu2021iconqa}} \end{tabular} &
\tikz[baseline=0.05em] \fill [color={rgb,255:red,204; green,168; blue,247}] (0,0) rectangle (0.75em,0.75em); \begin{tabular}{l} PMC-VQA (MathV360K) \\ {\footnotesize \cite{zhang2024pmcvqavisualinstructiontuning}} \end{tabular} &
\tikz[baseline=0.05em] \fill [color={rgb,255:red,210; green,178; blue,250}] (0,0) rectangle (0.75em,0.75em); \begin{tabular}{l} Super-CLEVR (MathV360K) \\ {\footnotesize \cite{li2023super}} \end{tabular} \\

\tikz[baseline=0.05em] \fill [color={rgb,255:red,215; green,188; blue,252}] (0,0) rectangle (0.75em,0.75em); \begin{tabular}{l} TabMWP (MathV360K) \\ {\footnotesize \cite{lu2023dynamic}} \end{tabular} &
\tikz[baseline=0.05em] \fill [color={rgb,255:red,220; green,198; blue,255}] (0,0) rectangle (0.75em,0.75em); \begin{tabular}{l} UniGeo (MathV360K) \\ {\footnotesize \cite{chen2022unigeounifyinggeometrylogical}} \end{tabular} &
\tikz[baseline=0.05em] \fill [color={rgb,255:red,225; green,208; blue,255}] (0,0) rectangle (0.75em,0.75em); \begin{tabular}{l} VizWiz (MathV360K) \\ {\footnotesize \cite{gurari2018vizwizgrandchallengeanswering}} \end{tabular} \\

\tikz[baseline=0.05em] \fill [color={rgb,255:red,230; green,218; blue,255}] (0,0) rectangle (0.75em,0.75em); \begin{tabular}{l} MapQA (MathV360K) \\ {\footnotesize \cite{chang2022mapqa}} \end{tabular} &
\tikz[baseline=0.05em] \fill [color={rgb,255:red,235; green,178; blue,255}] (0,0) rectangle (0.75em,0.75em); \begin{tabular}{l} Raven (cauldron) \\ {\footnotesize \cite{zhang2019raven}} \end{tabular} &
\tikz[baseline=0.05em] \fill [color={rgb,255:red,240; green,168; blue,255}] (0,0) rectangle (0.75em,0.75em); \begin{tabular}{l} M3IT+FLAN \\ {\footnotesize \cite{li2023m3it}} \end{tabular} \\

\tikz[baseline=0.05em] \fill [color={rgb,255:red,245; green,158; blue,255}] (0,0) rectangle (0.75em,0.75em); \begin{tabular}{l} WIT \\ {\footnotesize \cite{srinivasan2021wit}} \end{tabular} &
\tikz[baseline=0.05em] \fill [color={rgb,255:red,250; green,145; blue,255}] (0,0) rectangle (0.75em,0.75em); \begin{tabular}{l} ViQuAE \\ {\footnotesize \cite{lerner:hal-03650618}} \end{tabular} &
\tikz[baseline=0.05em] \fill [color={rgb,255:red,255; green,135; blue,255}] (0,0) rectangle (0.75em,0.75em); \begin{tabular}{l} A-OKVQA (cauldron, llava-format) \\ {\footnotesize \cite{schwenk2022aokvqabenchmarkvisualquestion}} \end{tabular} \\

\tikz[baseline=0.05em] \fill [color={rgb,255:red,255; green,125; blue,255}] (0,0) rectangle (0.75em,0.75em); \begin{tabular}{l} Vision-Flan (filtered) \\ {\footnotesize \cite{xu2024vision}} \end{tabular} &
\tikz[baseline=0.05em] \fill [color={rgb,255:red,255; green,225; blue,255}] (0,0) rectangle (0.75em,0.75em); \begin{tabular}{l} WebSight (cauldron) \\ {\footnotesize \cite{laurençon2024unlockingconversionwebscreenshots}} \end{tabular} &
\tikz[baseline=0.05em] \fill [color={rgb,255:red,255; green,215; blue,255}] (0,0) rectangle (0.75em,0.75em); \begin{tabular}{l} VSR (cauldron, llava-format) \\ {\footnotesize \cite{Liu2022VisualSR}} \end{tabular} \\

\tikz[baseline=0.05em] \fill [color={rgb,255:red,255; green,205; blue,255}] (0,0) rectangle (0.75em,0.75em); \begin{tabular}{l} CLEVR \\ {\footnotesize \cite{johnson2017clevr}} \end{tabular} &
\tikz[baseline=0.05em] \fill [color={rgb,255:red,255; green,195; blue,255}] (0,0) rectangle (0.75em,0.75em); \begin{tabular}{l} TallyQA (cauldron, llava-format) \\ {\footnotesize \cite{TallyQA}} \end{tabular} &
\tikz[baseline=0.05em] \fill [color={rgb,255:red,255; green,185; blue,255}] (0,0) rectangle (0.75em,0.75em); \begin{tabular}{l} ScienceQA (nona-context) \\ {\footnotesize \cite{ScienceQA}} \end{tabular} \\

\tikz[baseline=0.05em] \fill [color={rgb,255:red,255; green,175; blue,255}] (0,0) rectangle (0.75em,0.75em); \begin{tabular}{l} PathVQA \\ {\footnotesize \cite{he2020pathvqa30000questionsmedical}} \end{tabular} &
\tikz[baseline=0.05em] \fill [color={rgb,255:red,255; green,165; blue,255}] (0,0) rectangle (0.75em,0.75em); \begin{tabular}{l} TQA (cauldron, llava-format) \\ {\footnotesize \cite{TQA}} \end{tabular} &
\tikz[baseline=0.05em] \fill [color={rgb,255:red,255; green,155; blue,255}] (0,0) rectangle (0.75em,0.75em); \begin{tabular}{l} VQA-RAD (cauldron, llava-format) \\ {\footnotesize \cite{lau2018dataset}} \end{tabular} \\

\tikz[baseline=0.05em] \fill [color={rgb,255:red,255; green,175; blue,255}] (0,0) rectangle (0.75em,0.75em); \begin{tabular}{l} RCTW-17 \\ {\footnotesize \cite{shi2018icdar2017competitionreadingchinese}} \end{tabular} & & \\
\end{tabular}
\end{minipage}

\captionsetup{font={small}}
\caption{
\model data source group. Group B: datasets that are used for rewriting.
}
\vspace{-5pt}
\label{fig:group_abc_b}
\end{figure*}

\begin{figure*}[htp]
\centering
\begin{minipage}{.01\textwidth}
\end{minipage}%
\begin{minipage}{.99\textwidth}
\centering
\renewcommand{\arraystretch}{1.1}
\setlength\tabcolsep{1pt}
\fontsize{9pt}{12pt}\selectfont
\begin{tabular}{lll}
\makecell[l]{\cellcolor[RGB]{38,172,204} \textcolor{white}{Group C(22.9\%)}} &
\tikz[baseline=0.05em] \fill [color={rgb,255:red,38; green,172; blue,204}] (0,0) rectangle (0.75em,0.75em); \begin{tabular}{l} Iconqa (cauldron,llava-format) \\ {\footnotesize \cite{lu2021iconqa}} \end{tabular} &
\tikz[baseline=0.05em] \fill [color={rgb,255:red,51; green,177; blue,207}] (0,0) rectangle (0.75em,0.75em); \begin{tabular}{l} CN-OCR-1 \\ {\footnotesize \cite{russakovsky2015imagenetlargescalevisual}} \end{tabular} \\

\tikz[baseline=0.05em] \fill [color={rgb,255:red,64; green,182; blue,210}] (0,0) rectangle (0.75em,0.75em); \begin{tabular}{l} K12-Printing \\ {\footnotesize \cite{li2024llava}} \end{tabular} &
\tikz[baseline=0.05em] \fill [color={rgb,255:red,38; green,165; blue,199}] (0,0) rectangle (0.75em,0.75em); \begin{tabular}{l} Orand-Car-A \\ {\footnotesize \cite{russakovsky2015imagenetlargescalevisual}} \end{tabular} &
\tikz[baseline=0.05em] \fill [color={rgb,255:red,51; green,170; blue,202}] (0,0) rectangle (0.75em,0.75em); \begin{tabular}{l} Sroie \\ {\footnotesize \cite{huang2019icdar2019}} \end{tabular} \\

\tikz[baseline=0.05em] \fill [color={rgb,255:red,64; green,175; blue,205}] (0,0) rectangle (0.75em,0.75em); \begin{tabular}{l} Ureader-IE \\ {\footnotesize \cite{ye-etal-2023-ureader}} \end{tabular} &
\tikz[baseline=0.05em] \fill [color={rgb,255:red,38; green,160; blue,195}] (0,0) rectangle (0.75em,0.75em); \begin{tabular}{l} Inter-GPs (cauldron,llava-format) \\ {\footnotesize \cite{lu2021inter}} \end{tabular} &
\tikz[baseline=0.05em] \fill [color={rgb,255:red,51; green,167; blue,199}] (0,0) rectangle (0.75em,0.75em); \begin{tabular}{l} Q-Align \\ {\footnotesize \cite{wu2023qalignteachinglmmsvisual}} \end{tabular} \\

\tikz[baseline=0.05em] \fill [color={rgb,255:red,64; green,172; blue,205}] (0,0) rectangle (0.75em,0.75em); \begin{tabular}{l} Q-Instruct \\ {\footnotesize \cite{wu2023qinstructimprovinglowlevelvisual}} \end{tabular} &
\tikz[baseline=0.05em] \fill [color={rgb,255:red,38; green,155; blue,190}] (0,0) rectangle (0.75em,0.75em); \begin{tabular}{l} VizWiz \\ {\footnotesize \cite{gurari2018vizwizgrandchallengeanswering}} \end{tabular} &
\tikz[baseline=0.05em] \fill [color={rgb,255:red,51; green,162; blue,193}] (0,0) rectangle (0.75em,0.75em); \begin{tabular}{l} OODVQA \\ {\footnotesize \cite{tu2023unicornsimagesafetyevaluation}} \end{tabular} \\

\tikz[baseline=0.05em] \fill [color={rgb,255:red,64; green,167; blue,200}] (0,0) rectangle (0.75em,0.75em); \begin{tabular}{l} SketchyVQA \\ {\footnotesize \cite{tu2023unicornsimagesafetyevaluation}} \end{tabular} &
\tikz[baseline=0.05em] \fill [color={rgb,255:red,38; green,148; blue,185}] (0,0) rectangle (0.75em,0.75em); \begin{tabular}{l} CLEVR (cauldron,llava-format) \\ {\footnotesize \cite{johnson2017clevr}} \end{tabular} &
\tikz[baseline=0.05em] \fill [color={rgb,255:red,51; green,155; blue,188}] (0,0) rectangle (0.75em,0.75em); \begin{tabular}{l} FigureQA (cauldron,llava-format) \\ {\footnotesize \cite{FigureQA}} \end{tabular} \\

\tikz[baseline=0.05em] \fill [color={rgb,255:red,64; green,160; blue,190}] (0,0) rectangle (0.75em,0.75em); \begin{tabular}{l} Hatefulmemes (cauldron,llava-format) \\ {\footnotesize \cite{hatefulmeme}} \end{tabular} &
\tikz[baseline=0.05em] \fill [color={rgb,255:red,38; green,140; blue,175}] (0,0) rectangle (0.75em,0.75em); \begin{tabular}{l} Screen2Words (cauldron) \\ {\footnotesize \cite{screen2words}} \end{tabular} &
\tikz[baseline=0.05em] \fill [color={rgb,255:red,51; green,147; blue,180}] (0,0) rectangle (0.75em,0.75em); \begin{tabular}{l} COCOQA \\ {\footnotesize \cite{ren2015exploring}} \end{tabular} \\

\tikz[baseline=0.05em] \fill [color={rgb,255:red,64; green,152; blue,185}] (0,0) rectangle (0.75em,0.75em); \begin{tabular}{l} VQAv2 \\ {\footnotesize \cite{goyal2017making}} \end{tabular} &
\tikz[baseline=0.05em] \fill [color={rgb,255:red,38; green,135; blue,170}] (0,0) rectangle (0.75em,0.75em); \begin{tabular}{l} LNQA \\ {\footnotesize \cite{changpinyo2022needvqaimagecaptions}} \end{tabular} &
\tikz[baseline=0.05em] \fill [color={rgb,255:red,51; green,142; blue,175}] (0,0) rectangle (0.75em,0.75em); \begin{tabular}{l} COCOCaption-Train \\ {\footnotesize \cite{lin2015microsoftcococommonobjects}} \end{tabular} \\

\tikz[baseline=0.05em] \fill [color={rgb,255:red,64; green,147; blue,180}] (0,0) rectangle (0.75em,0.75em); \begin{tabular}{l} image-textualization (filtered) \\ {\footnotesize \cite{pi2024imagetextualizationautomaticframework}} \end{tabular} &
\tikz[baseline=0.05em] \fill [color={rgb,255:red,38; green,125; blue,160}] (0,0) rectangle (0.75em,0.75em); \begin{tabular}{l} COCOCaption-Val \\ {\footnotesize \cite{lin2015microsoftcococommonobjects}} \end{tabular} &
\tikz[baseline=0.05em] \fill [color={rgb,255:red,51; green,132; blue,165}] (0,0) rectangle (0.75em,0.75em); \begin{tabular}{l} MSCOCO \\ {\footnotesize \cite{lin2015microsoftcococommonobjects}} \end{tabular} \\

\tikz[baseline=0.05em] \fill [color={rgb,255:red,64; green,137; blue,170}] (0,0) rectangle (0.75em,0.75em); \begin{tabular}{l} CAP2QA \\ {\footnotesize \cite{cha2024visually}} \end{tabular} &
\tikz[baseline=0.05em] \fill [color={rgb,255:red,38; green,120; blue,150}] (0,0) rectangle (0.75em,0.75em); \begin{tabular}{l} Macaw-LLM \\ {\footnotesize \cite{lyu2023macawllmmultimodallanguagemodeling}} \end{tabular} &
\tikz[baseline=0.05em] \fill [color={rgb,255:red,51; green,127; blue,160}] (0,0) rectangle (0.75em,0.75em); \begin{tabular}{l} DetGPT \\ {\footnotesize \cite{pi2023detgptdetectneedreasoning}} \end{tabular} \\

\tikz[baseline=0.05em] \fill [color={rgb,255:red,64; green,132; blue,175}] (0,0) rectangle (0.75em,0.75em); \begin{tabular}{l} TextCaps \\ {\footnotesize \cite{sidorov2019textcaps}} \end{tabular} &
\tikz[baseline=0.05em] \fill [color={rgb,255:red,38; green,110; blue,140}] (0,0) rectangle (0.75em,0.75em); \begin{tabular}{l} Ureader-Cap \\ {\footnotesize \cite{ye-etal-2023-ureader}} \end{tabular} &
\tikz[baseline=0.05em] \fill [color={rgb,255:red,51; green,117; blue,150}] (0,0) rectangle (0.75em,0.75em); \begin{tabular}{l} Ureader-Caption \\ {\footnotesize \cite{ye-etal-2023-ureader}} \end{tabular} \\

\tikz[baseline=0.05em] \fill [color={rgb,255:red,64; green,122; blue,160}] (0,0) rectangle (0.75em,0.75em); \begin{tabular}{l} LocalizedNarratives \\ {\footnotesize  \cite{PontTuset_eccv2020}} \end{tabular} &
\tikz[baseline=0.05em] \fill [color={rgb,255:red,38; green,105; blue,130}] (0,0) rectangle (0.75em,0.75em); \begin{tabular}{l} FigureQA (MathV360K) \\ {\footnotesize \cite{FigureQA}} \end{tabular} &
\tikz[baseline=0.05em] \fill [color={rgb,255:red,51; green,112; blue,140}] (0,0) rectangle (0.75em,0.75em); \begin{tabular}{l} Hand-Written-Arith \\ {\footnotesize (\textcolor{blue}{Self Collection}) } \end{tabular} \\

\tikz[baseline=0.05em] \fill [color={rgb,255:red,64; green,117; blue,150}] (0,0) rectangle (0.75em,0.75em); \begin{tabular}{l} MapQA (cauldron,llava-format) \\ {\footnotesize \cite{chang2022mapqa}} \end{tabular} &
\tikz[baseline=0.05em] \fill [color={rgb,255:red,38; green,90; blue,120}] (0,0) rectangle (0.75em,0.75em); \begin{tabular}{l} VisualGenome \\ {\footnotesize \cite{Krishna2017}} \end{tabular} &
\tikz[baseline=0.05em] \fill [color={rgb,255:red,51; green,97; blue,130}] (0,0) rectangle (0.75em,0.75em); \begin{tabular}{l} RefCOCO \\ {\footnotesize \cite{yu2016modelingcontextreferringexpressions}} \end{tabular} \\

\end{tabular}
\end{minipage}

\captionsetup{font={small}}
\caption{
\model data source group. Group C: data sources that are not included.
}
\vspace{-5pt}
\label{fig:group_abc_c}
\end{figure*}

\clearpage
\section{Case Study}



\subsection{Incorrect Cases}

\label{app:bad_case}

\begin{promptbox}[OCR Incorrect Case]{darkblue}
\begin{minipage}{\textwidth}
    \includegraphics[height=8cm]{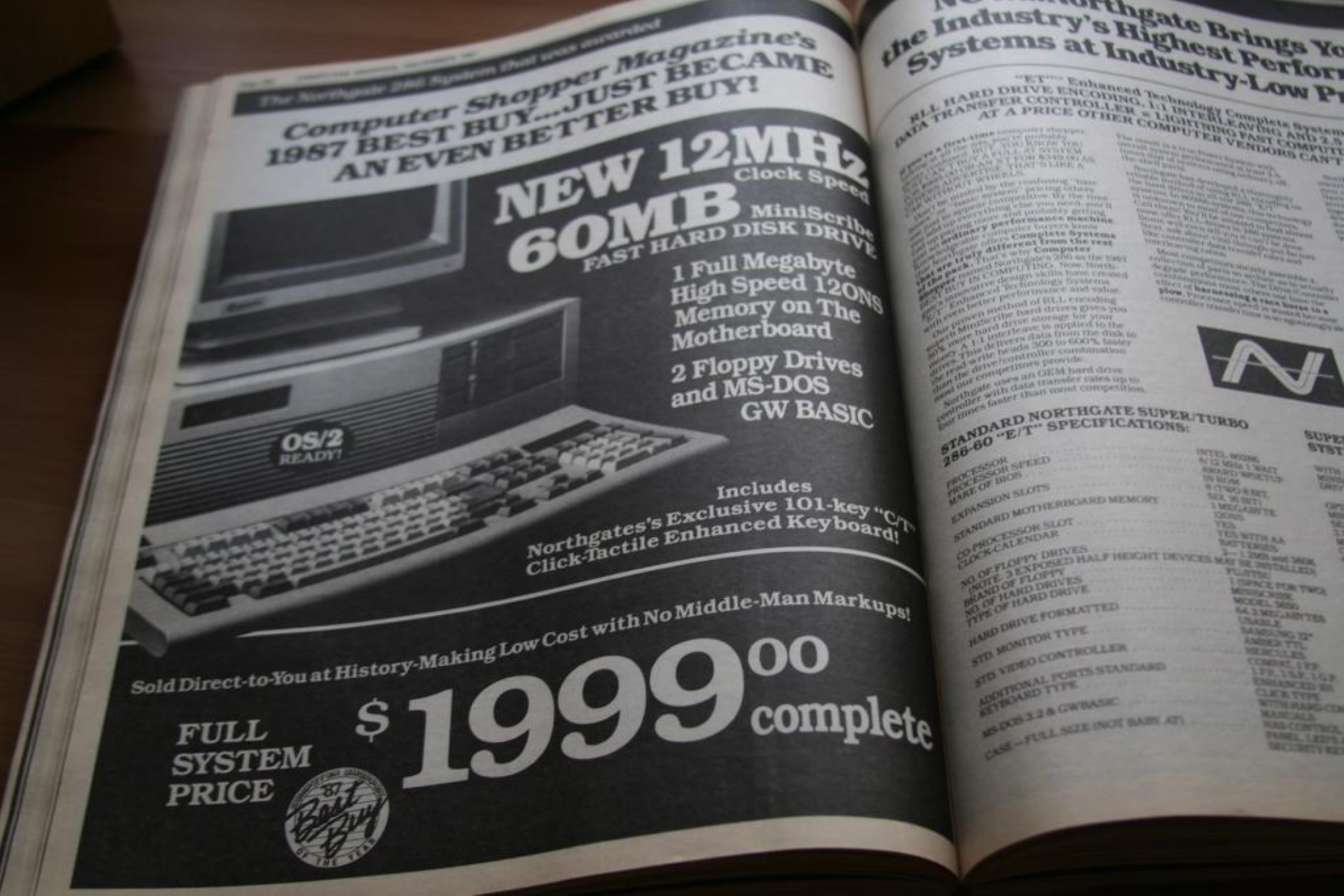}
\end{minipage}

\vspace{1em}

\textbf{Original Question}: whats the amount on the page?

\vspace{1em}

\textbf{Original Answer}: 1999

\vspace{1em}

\textbf{Incorrect Rewrite Question}: A tech-savvy customer is interested in purchasing a computer system that can support high-performance graphics. Based on the advertisement, what specific features of the Northgate 286 system would you highlight to convince them to make a purchase, and how do these features compare to the industry standards of 1987?
\vspace{1em}
\textbf{Incorrect Rewrite Answer}: To convince the customer to purchase the Northgate 286 system, I would highlight the following features that are relevant to high-performance graphics:\\
The system's "Super VGA Graphics" capability, which suggests that it can support high-resolution graphics.The inclusion of a "Northgate Graphics Card", which is specifically designed to enhance graphics performance. The "101-key Enhanced Keyboard" with its "OS/2 READY!" label, implying that the system is compatible with Operating System/2 (OS/2), which was known for its graphics capabilities.In 1987, these features would have been considered impressive, especially considering the system's competitive pricing. The Super VGA Graphics capability would have provided high-quality graphics performance, while the Northgate Graphics Card would have further enhanced the system's graphics capabilities. The OS/2 compatibility would have also been seen as a significant advantage, given the operating system's reputation for graphics support. It's worth noting that the system's 12MHz clock speed, 1MB of RAM, and 60MB hard disk drive would have also contributed to its overall performance, making it an attractive option for customers requiring high-performance graphics. By highlighting these features, I would be able to demonstrate the Northgate 286 system's capabilities and value proposition to the customer, making a compelling case for its purchase.

\vspace{1em}

\textbf{Analysis}: Some details in the answer do not fully match the information in the image. The analysis is as follows:\\
1.Super VGA Graphics: The image does not mention the "Super VGA Graphics" functionality, so this point is inaccurate.
2. Northgate Graphics Card: The image does not mention the "Northgate Graphics Card" or any specific graphical capabilities, so this point is also incorrect.\\
3. OS/2 READY: The image indeed has an "OS/2 READY!" label, which is correct. However, this only indicates system compatibility with OS/2 and does not necessarily relate to high-performance graphical capabilities.\\
4. 101-key Enhanced Keyboard: The image indeed mentions that the system comes with a "101-key Click-Tactile Enhanced Keyboard," and this description is accurate.\\
5. 12MHz processor, 1MB memory, and 60MB hard drive: These configurations do align with the information in the image.
\end{promptbox}

\newpage

\begin{promptbox}[Chart Incorrect Case]{darkblue}
\begin{minipage}{\textwidth}
    \includegraphics[height=8cm]{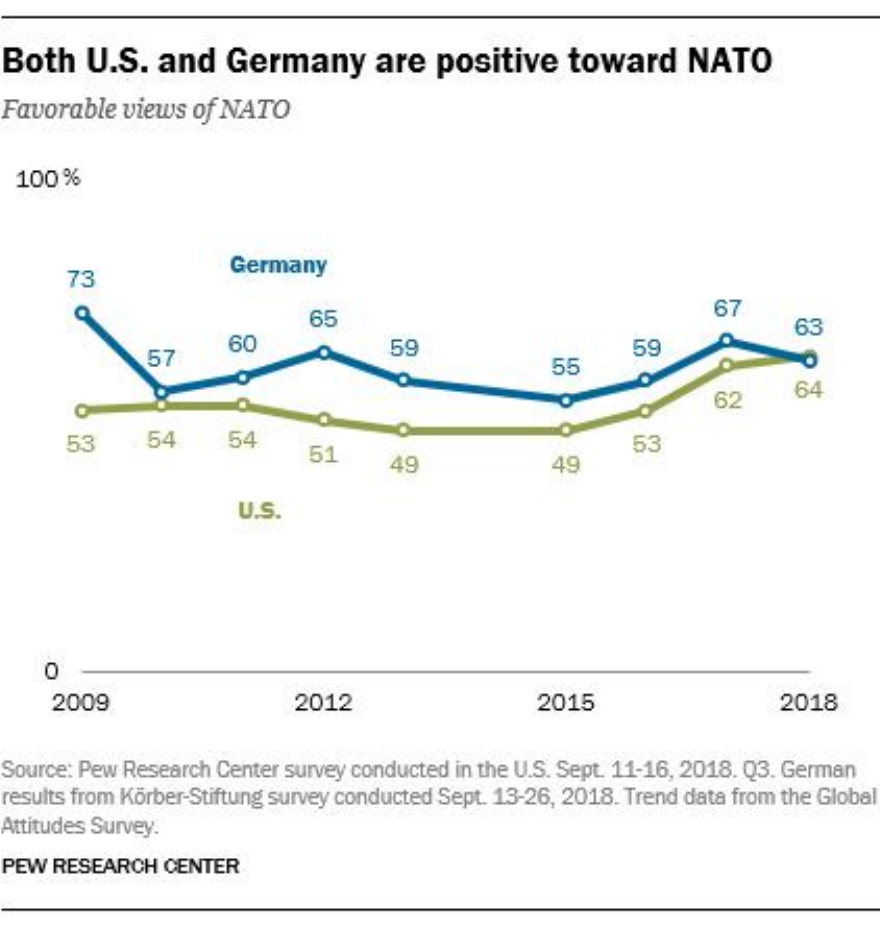}
\end{minipage}

\vspace{1em}

\textbf{Original Question}: What's the rightmost value of green graph? Answer the question using a single word or phrase

\vspace{1em}

\textbf{Original Answer}: 64

\vspace{1em}

\textbf{Incorrect Rewrite Question}: Calculate the average favorability rating for the U.S. and Germany from 2009 to 2018. Compare the results and discuss the significance of any differences.

\vspace{1em}

\textbf{Incorrect Rewrite Answer}: To calculate the average favorability rating for the U.S., we add up the values for each year and divide by the number of years: (73 + 54 + 49 + 64) / 4 = 240 / 4 = 60\%For Germany, we follow the same process: (60 + 65 + 55 + 67) / 4 = 247 / 4 = 61.75\% 
The average favorability rating for the U.S. is 60\%, while for Germany, it is 61.75\%. The difference is relatively small, indicating that both countries have similar levels of support for NATO over the years. However, the slightly higher average for Germany could suggest a more consistent and stable support for NATO compared to the U.S. This difference could be significant in the context of NATO's decision-making process, where member countries' opinions and support play a crucial role. A more consistent support from Germany could influence the alliance's policies and strategies, potentially leading to a stronger and more unified NATO.

\vspace{1em}

\textbf{Analysis}: Here is a detailed analysis of the errors:
\\
1. Calculation Error for Averages: 
   - For the U.S., the values in the graph are 73, 57, 49, and 64, but the calculation uses 73, 54, 49, and 64. This mismatch leads to an incorrect average. The actual average for the U.S. should be \( (73 + 57 + 49 + 64) / 4 = 243 / 4 = 60.75\%\).
   - For Germany, the values in the graph are 60, 59, 55, and 67, but the calculation uses 60, 65, 55, and 67. The actual average for Germany should be \( (60 + 59 + 55 + 67) / 4 = 241 / 4 = 60.25\%\).
\\
2. Incorrect Comparison:
   - The answer claims that the U.S. average is 60\% and Germany's average is 61.75\%, but the recalculated averages (60.75\% for the U.S. and 60.25\% for Germany) reveal the opposite trend. The U.S. has a slightly higher average, not Germany.
\\
3. Misinterpretation of Significance:
   - The conclusion states that Germany has "more consistent and stable support for NATO" compared to the U.S., based on the incorrect averages. However, the actual data shows that the averages are nearly the same, and there is no significant evidence of Germany having more consistent support than the U.S. The fluctuations in favorability ratings over the years are similar for both countries.
\\
4. Failure to Acknowledge Trend Differences:
   - The analysis does not highlight the differing trends in the graph. For example, U.S. favorability starts high, dips, and then rises again, while Germany's favorability remains relatively stable but does not show the same sharp recovery seen in the U.S. This is a critical observation that the answer omits.
\\
5. Use of Rounded Averages:
   - The response uses rounded averages (e.g., 60\% for the U.S. and 61.75\% for Germany) without clarifying that this is an approximation, which can lead to a misrepresentation of the actual differences.

\end{promptbox}

\begin{promptbox}[Math Incorrect Case]{darkblue}
\begin{minipage}{\textwidth}
    \includegraphics[height=4cm]{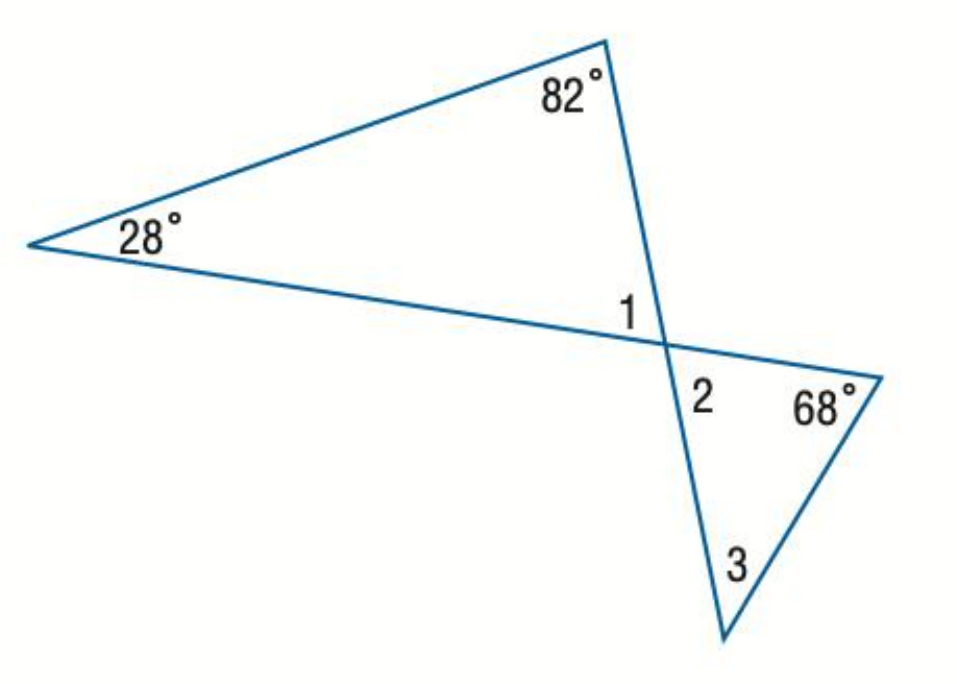}
\end{minipage}

\vspace{1em}

\textbf{Original Question}: Hint: Please answer the question and provide the correct option letter, e.g., A, B, C, D, at the end. Question: Find the measure of angle 3. Choices: (A) 28 (B) 32 (C) 38 (D) 42

\vspace{1em}

\textbf{Original Answer}: The answer is D.

\vspace{1em}

\textbf{Incorrect Rewrite Question}: Hint: Please answer the question and provide the correct option letter, e.g., A, B, C, D, at the end. Question: Find the measure of angle 3. Choices: (A) 28 (B) 32 (C) 38 (D) 42

\vspace{1em}

\textbf{Incorrect Rewrite Answer}: To find the measure of \( \angle 3 \), we need to use the properties of triangles and the fact that the sum of the angles in a triangle is always \( 180^\circ \).
1. First, consider the triangle on the left with angles \( 28^\circ \), \( 82^\circ \), and \( \angle 1 \).
   - The sum of the angles in this triangle is \( 180^\circ \).
   - Therefore, we can write the equation:
     \[
     28^\circ + 82^\circ + \angle 1 = 180^\circ
     \]
   - Solving for \( \angle 1 \):
     \[
     \angle 1 = 180^\circ - 28^\circ - 82^\circ = 70^\circ
     \]
2. Next, consider the triangle on the right with angles \( 68^\circ \), \( \angle 2 \), and \( \angle 3 \).
   - The sum of the angles in this triangle is also \( 180^\circ \).
   - Therefore, we can write the equation:
     \[
     68^\circ + \angle 2 + \angle 3 = 180^\circ
     \]
3. Notice that \( \angle 1 \) and \( \angle 2 \) are supplementary because they form a straight line.
   - Therefore:
     \[
     \angle 1 + \angle 2 = 180^\circ
     \]
   - Substituting the value of \( \angle 1 \) we found earlier:
     \[
     70^\circ + \angle 2 = 180^\circ
     \]
   - Solving for \( \angle 2 \):
     \[
     \angle 2 = 180^\circ - 70^\circ = 110^\circ
     \]
4. Now, substitute \( \angle 2 \) back into the equation for the sum of angles in the triangle on the right:
   \[
   68^\circ + 110^\circ + \angle 3 = 180^\circ
   \]
   - Solving for \( \angle 3 \):
     \[
     \angle 3 = 180^\circ - 68^\circ - 110^\circ = 42^\circ
     \]
Therefore, the measure of \( \angle 3 \) is \( 42^\circ \), which corresponds to option (D).
Revised Answer: The answer is D.

\vspace{1em}

\textbf{Analysis}: The key mistake in the reasoning lies in the incorrect assumption about the relationship between \( \angle 1 \) and \( \angle 2 \). Here is the analysis:
\\
Incorrect Assumption:\\
It was assumed that \( \angle 1 + \angle 2 = 180^\circ \), implying that \( \angle 1 \) and \( \angle 2 \) are supplementary angles. \\
This is incorrect because \( \angle 1 \) and \( \angle 2 \) are actually vertical (opposite) angles, as they are formed by the intersection of two straight lines.

\end{promptbox}

\begin{promptbox}[Caption Incorrect Case]{darkblue}
\begin{minipage}{\textwidth}
    \includegraphics[height=4cm]{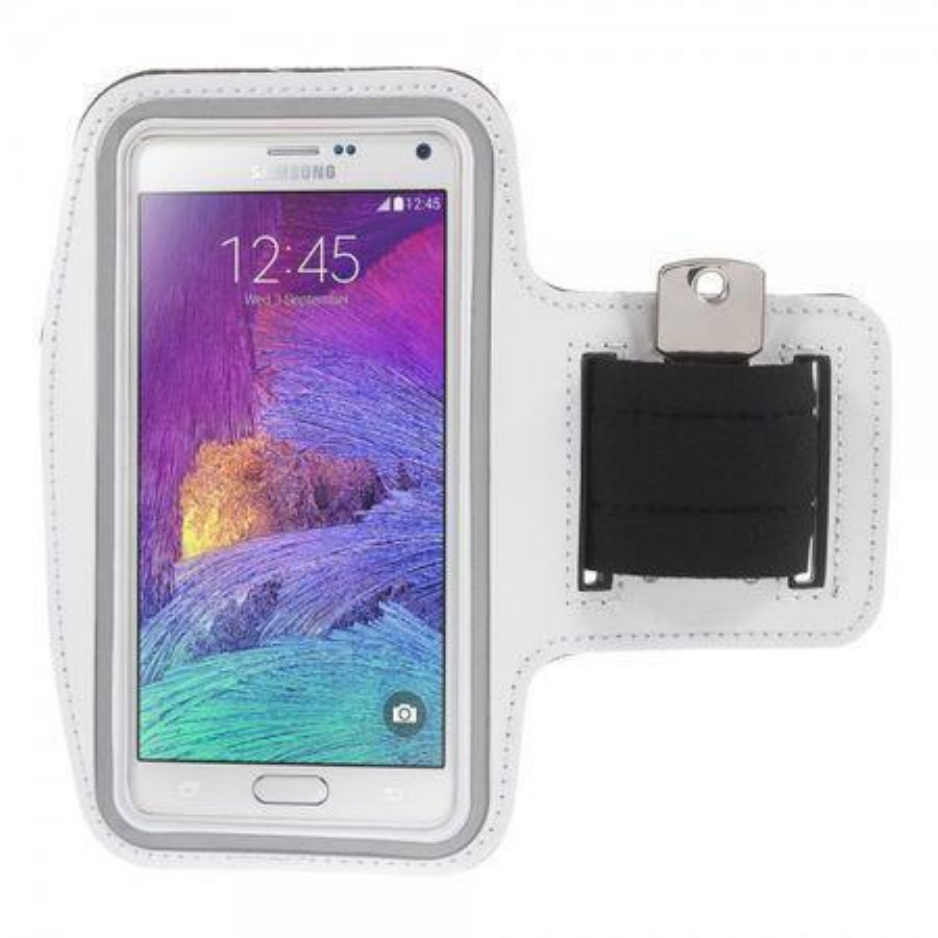}
\end{minipage}

\vspace{1em}

\textbf{Original Caption}: The image displays a white, protective armband designed to hold a smartphone. Specifically, the smartphone inserted into the armband appears to be a Samsung device. The design of the armband is structured to be worn around the upper arm, probably intended for use during physical activities such as running or working out. The front section of the armband is clear, allowing for visibility and usage of the smartphone's touchscreen. Here are the detailed components and features observed in the image: 1. Smartphone:   1) Brand and Model: The phone is from the Samsung brand, identifiable by the name imprinted on its front. The exact model is not explicitly stated, but its design characteristics suggest it could be from the Galaxy series.    2) Time Display: The phone's display shows the time as 12:45. This time might be relevant to showcasing the phone's screen capabilities.    3) Date Display: The date shown is Wednesday, September 3. This could either be a static image used for promotional purposes or a real-time capture.    4) UI Elements: The display shows typical smartphone UI elements, including the status bar with signal strength, WiFi connectivity, battery percentage, and possibly notifications. 2. Armband:    1) Material and Color: The armband material appears to be a blend of fabric and synthetic materials, predominantly in white color.    2) Clear Front Cover: The front cover of the armband is clear, providing visibility and functionality to the smartphone's screen.    3) Strap and Buckle: To the side, it has an adjustable strap with a buckle mechanism. The strap appears to be made of black elastic material, ensuring that the armband can fit various arm sizes securely. The buckle is metallic, providing a durable mechanism for securing the armband.3. Design and Ergonomics:    1) Stitching and Build Quality: The stitching around the armband looks precise and robust, indicating good build quality aimed at withstanding regular usage.    2) Port Accessibility: The design does not obstruct the smartphone's various ports and buttons, ensuring that users can access volume controls, home buttons, or charging ports without removing the phone from the armband. 4. Functional Aspects:    1) Purpose: The primary function of the armband is to hold the smartphone securely during physical activities. The clear cover allows users to interact with their device without removing it.    2) Practical Features: The adjustable strap ensures a snug fit for different arm sizes, while the metallic buckle provides added security to prevent the armband from coming loose. 

\vspace{1em}

\textbf{Incorrect Rewrite Question}: Based on market trends for smartphone accessories and the functionalities observed in the armband, analyze the potential target market for this product and predict its sales outlook in the next quarter, assuming an MSRP of \$29.99 and availability through major online retailers.

\vspace{1em}

\textbf{Incorrect Rewrite Answer}: Analyzing the market trends for smartphone accessories reveals a growing demand for hands-free solutions that cater to active lifestyles. Given the armband's functionalities, such as its adjustable strap, clear front cover, and secure buckle mechanism, it is poised to capture a significant share of the market. Target demographics likely include fitness enthusiasts, athletes, and individuals who enjoy outdoor activities. Assuming the MSRP of \$29.99, a competitive price point compared to similar products, and considering the armband's availability through major online retailers, we can predict the sales outlook for the next quarter. Historical sales data from similar products suggest a strong demand during the first quarter of the year, particularly in January and February, as consumers adopt New Year's resolutions related to fitness and wellness. Projecting a conservative growth rate of 10\% month-over-month and accounting for potential seasonality, we estimate the following sales figures: Month 1: 10,000 units, Month 2: 11,000 units, Month 3: 12,100 units. These numbers translate to quarterly revenue of approximately \$348,000, indicating a promising sales outlook for the armband in the next quarter.

\vspace{1em}

\textbf{Analysis}: Generated questions have very little relevance to the image.

\end{promptbox}

The examples show the comparison of cases before and after incorrect rewriting. The exceptions that occur during the rewriting process include generating information that is not in the image, incorrectly recognizing of information in the image, making errors in the mathematical logic reasoning process, and generating QA that is not related to the image.

\newpage

\subsection{Good Cases}
\label{app:good_case}

Analysis: Cases show the comparison of cases before and after good rewriting. Good rewrites bring more complex and diverse instructions, contain more knowledge, and expand the current monotonous datasets.

\begin{promptbox}[OCR Good Case]{darkblue}
\begin{minipage}{\textwidth}
    \includegraphics[height=8cm]{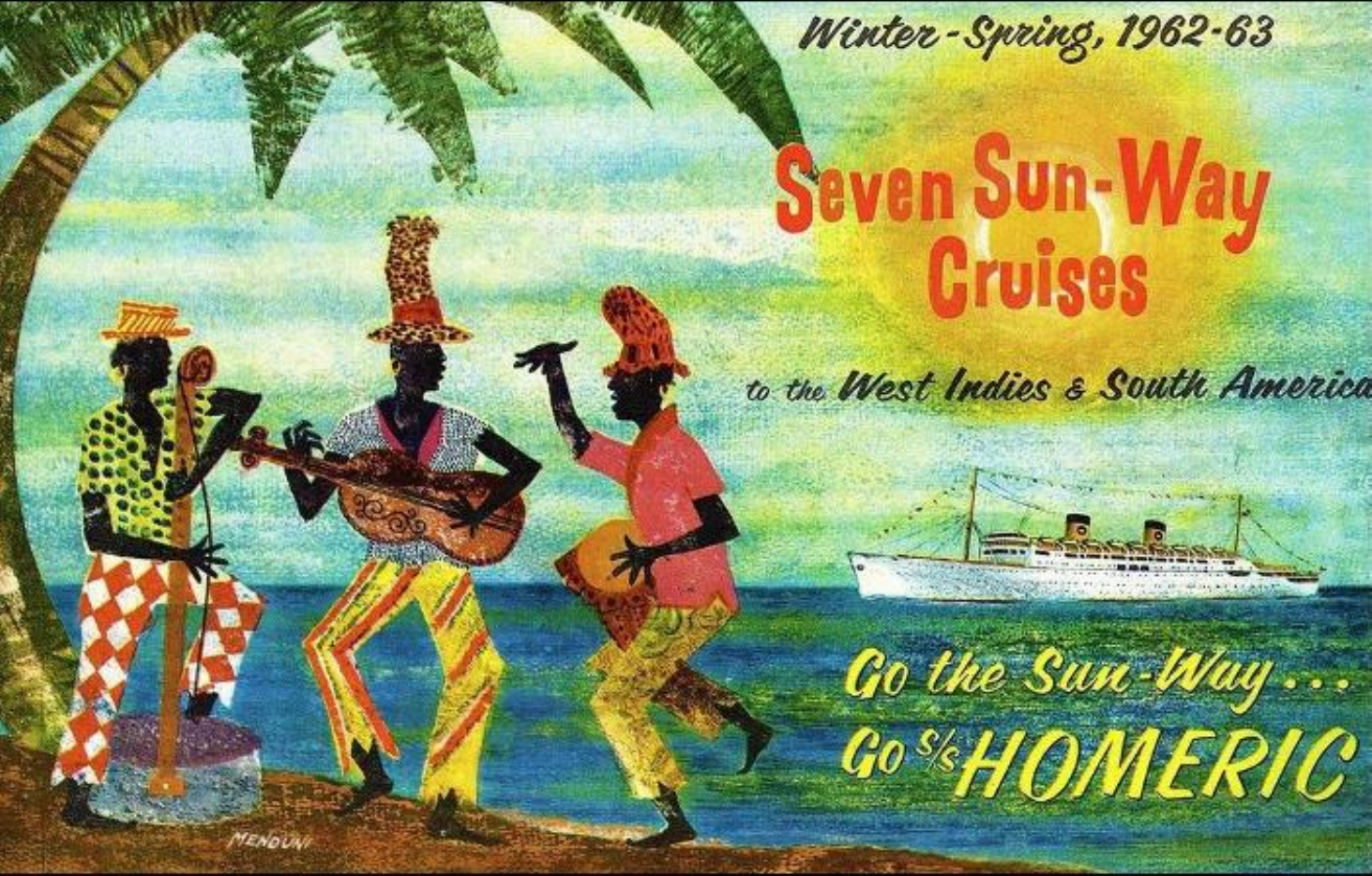}
\end{minipage}

\vspace{1em}

\textbf{Original Question}: what year is this?

\vspace{1em}

\textbf{Original Answer}: 1962-63

\vspace{1em}

\textbf{Rewrite Question}: According to the poster, what type of cruise experience can travelers expect during the Winter-Spring season of 1962-63, and how is this experience reflected in the visual elements of the advertisement?

\vspace{1em}

\textbf{Rewrite Answer}: Travelers can expect a vibrant and exotic getaway to the West Indies and South America during the Winter-Spring season of 1962-63, as promised by Seven Sun-Way Cruises. The visual elements of the poster effectively convey this experience by featuring three individuals in colorful traditional attire, playing musical instruments on a sunny beach setting. The use of tropical colors, palm trees, and a cruise ship on the horizon suggests a fun, lively, and adventurous atmosphere, which aligns with the idea of a warm and exotic getaway. The poster's focus on musicians and a picturesque seascape implies that the cruise experience will be filled with entertainment, relaxation, and opportunities to soak up the sun.

\end{promptbox}

\newpage

\begin{promptbox}[Chart Good Case]{darkblue}
\begin{minipage}{\textwidth}
    \includegraphics[height=8cm]{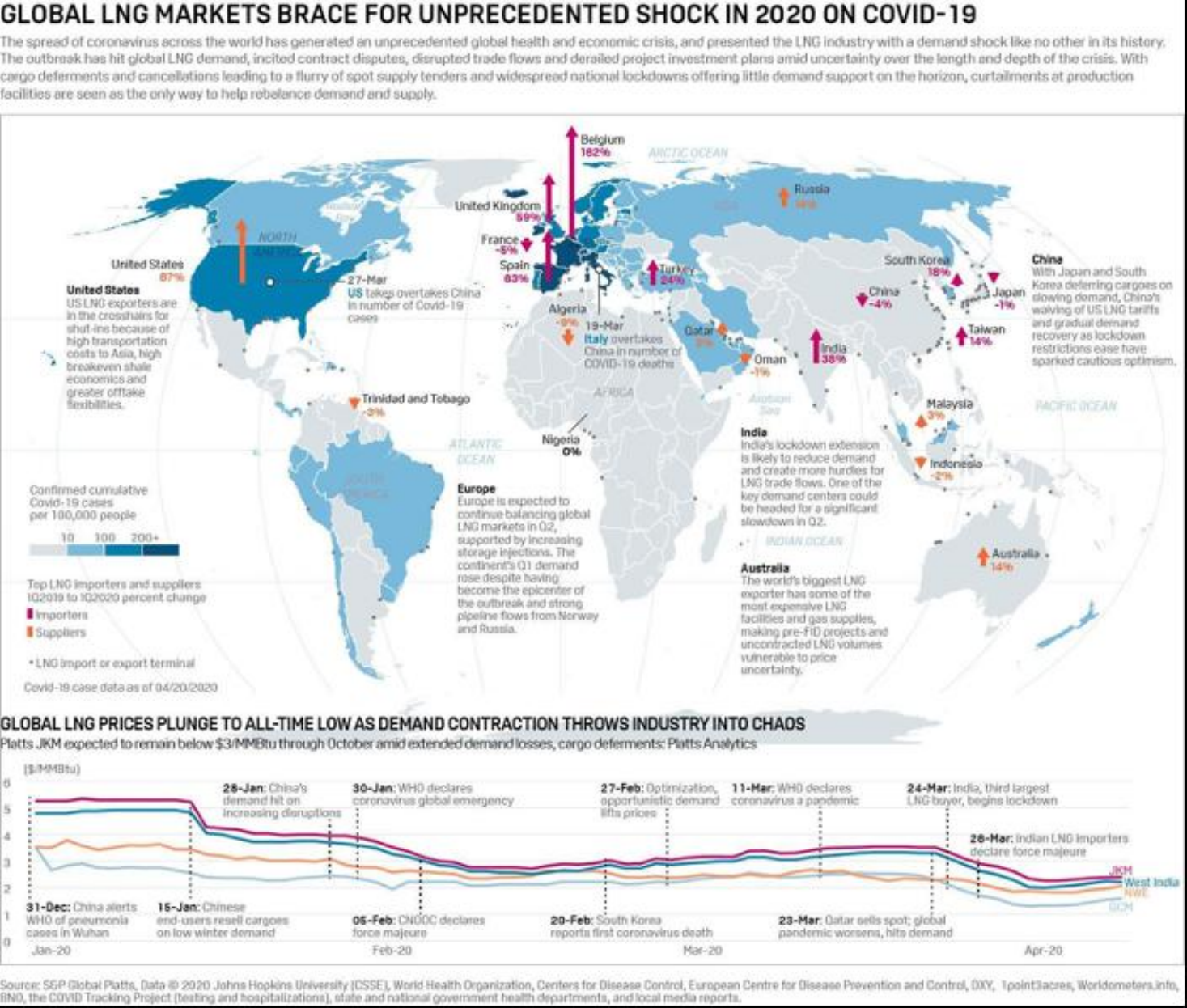}
\end{minipage}

\vspace{1em}

\textbf{Original Question}: On which day WHO declared Covid 19 global emergency?

\vspace{1em}

\textbf{Original Answer}: 30-Jan

\vspace{1em}

\textbf{Rewrite Question}: Analyze the global LNG market trends depicted in the image and discuss the potential long-term impacts of the COVID-19 pandemic on LNG demand and supply. Consider the role of major LNG importers and exporters, and how their strategies might evolve in response to the pandemic-induced market changes.

\vspace{1em}

\textbf{Rewrite Answer}: The image illustrates the significant disruption caused by the COVID-19 pandemic
in the global LNG market. The pandemic has led to a demand shock, with many countries experiencing reduced economic activity and lower energy consumption. This has resulted in
a flurry of spot supply tenders and widespread national lockdowns, which have further exacerbated the demand contraction. Major LNG importers such as Japan, South Korea, and China have seen significant reductions in LNG demand due to industrial slowdowns and decreased energy requirements. For instance, Japan and South Korea have deferred cargoes, while China has waived US LNG tariffs to support its economic recovery. These actions reflect the immediate response to the pandemic's impact on demand. On the supply side, major LNG exporters like Qatar, Australia, and the United States have had to navigate the
challenges posed by reduced demand and lower prices. Qatar, the world's largest LNG exporter, has had to sell spot cargoes as the pandemic worsens, highlighting the need for flexibility in supply contracts. In the long term, the pandemic is likely to accelerate the shift towards more flexible LNG supply contracts, as both buyers and sellers seek to mitigate risks associated with demand uncertainty. Additionally, the pandemic may spur investments in LNG infrastructure in regions with growing demand, such as Southeast Asia and Latin America, as countries seek to diversify their energy sources and enhance energy security. Overall, the COVID-19 pandemic has underscored the importance of agility and adaptability in the LNG market. As the global economy recovers, LNG demand is expected to rebound, but the industry will need to navigate a new landscape characterized by evolving market dynamics and changing energy consumption patterns.

\end{promptbox}

\begin{promptbox}[Math Good Case]{darkblue}
\begin{minipage}{\textwidth}
    \includegraphics[height=6cm]{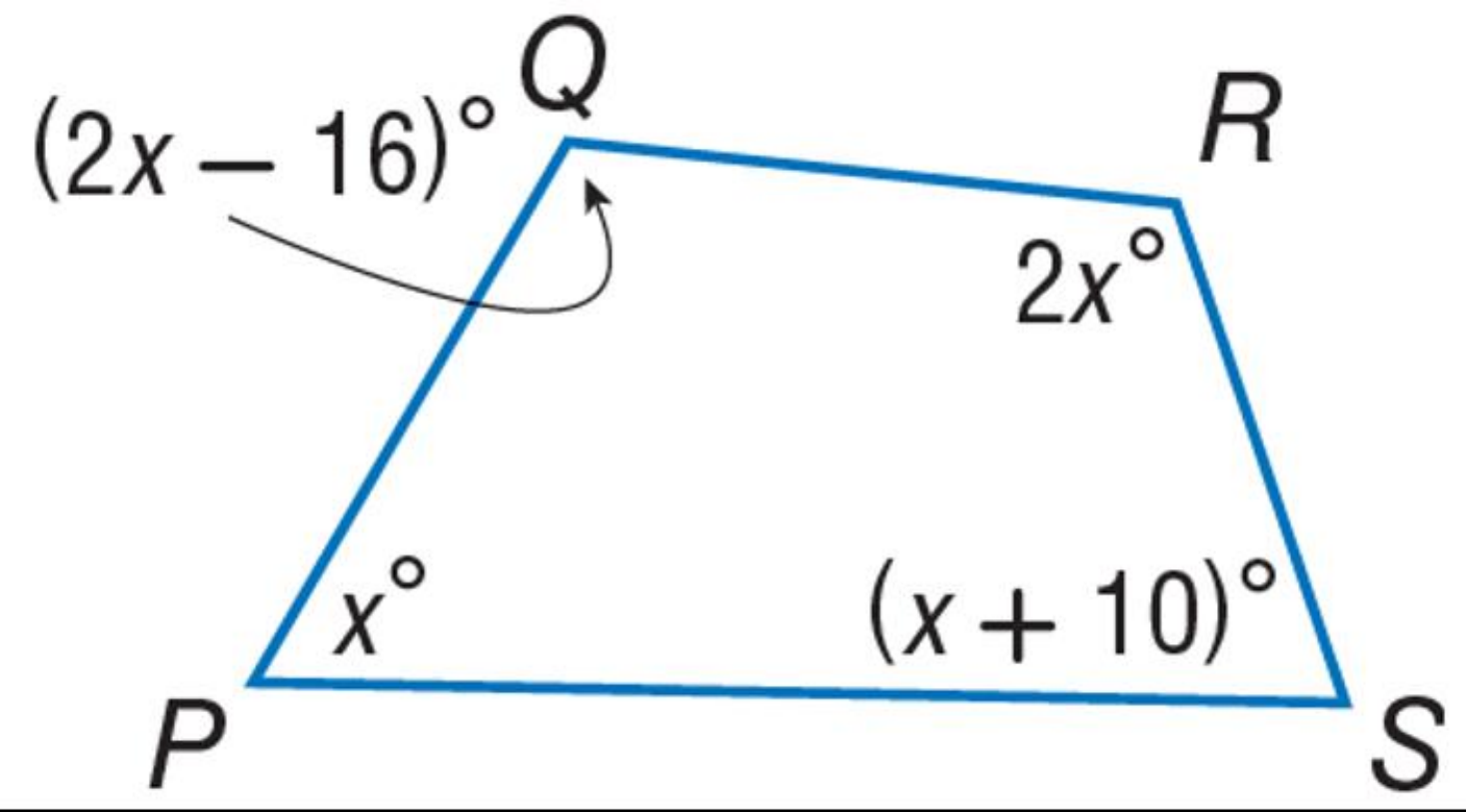}
\end{minipage}

\vspace{1em}

\textbf{Original Question}: What is the measure of angle QPS?

\vspace{1em}

\textbf{Original Answer}: The answer is 61

\vspace{1em}

\textbf{Rewrite Question}: What is the measure of angle QPS?

\vspace{1em}

\textbf{Rewrite Answer}: 
To find the measure of angle \( \angle QPS \), we need to use the properties of the quadrilateral \( PQRS \). The sum of the interior angles of any quadrilateral is \( 360^\circ \).

Given the angles:
\[
\begin{aligned}
&\angle P = x^\circ, \\
&\angle Q = (2x - 16)^\circ, \\
&\angle R = 2x^\circ, \\
&\angle S = (x + 10)^\circ,
\end{aligned}
\]

we can set up the equation for the sum of the interior angles:
\[
x + (2x - 16) + 2x + (x + 10) = 360.
\]

Combine like terms:
\[
x + 2x + 2x + x - 16 + 10 = 360,
\]
\[
6x - 6 = 360.
\]

Add 6 to both sides:
\[
6x = 366.
\]

Divide by 6:
\[
x = 61.
\]

Now, we need to find the measure of angle \( \angle QPS \), which is given as \( x \):
\[
x = 61^\circ.
\]

Therefore, the measure of angle \( \angle QPS \) is \( 61^\circ \).

\end{promptbox}

\begin{promptbox}[Caption Good Case]{darkblue}
\begin{minipage}{\textwidth}
    \includegraphics[height=8cm]{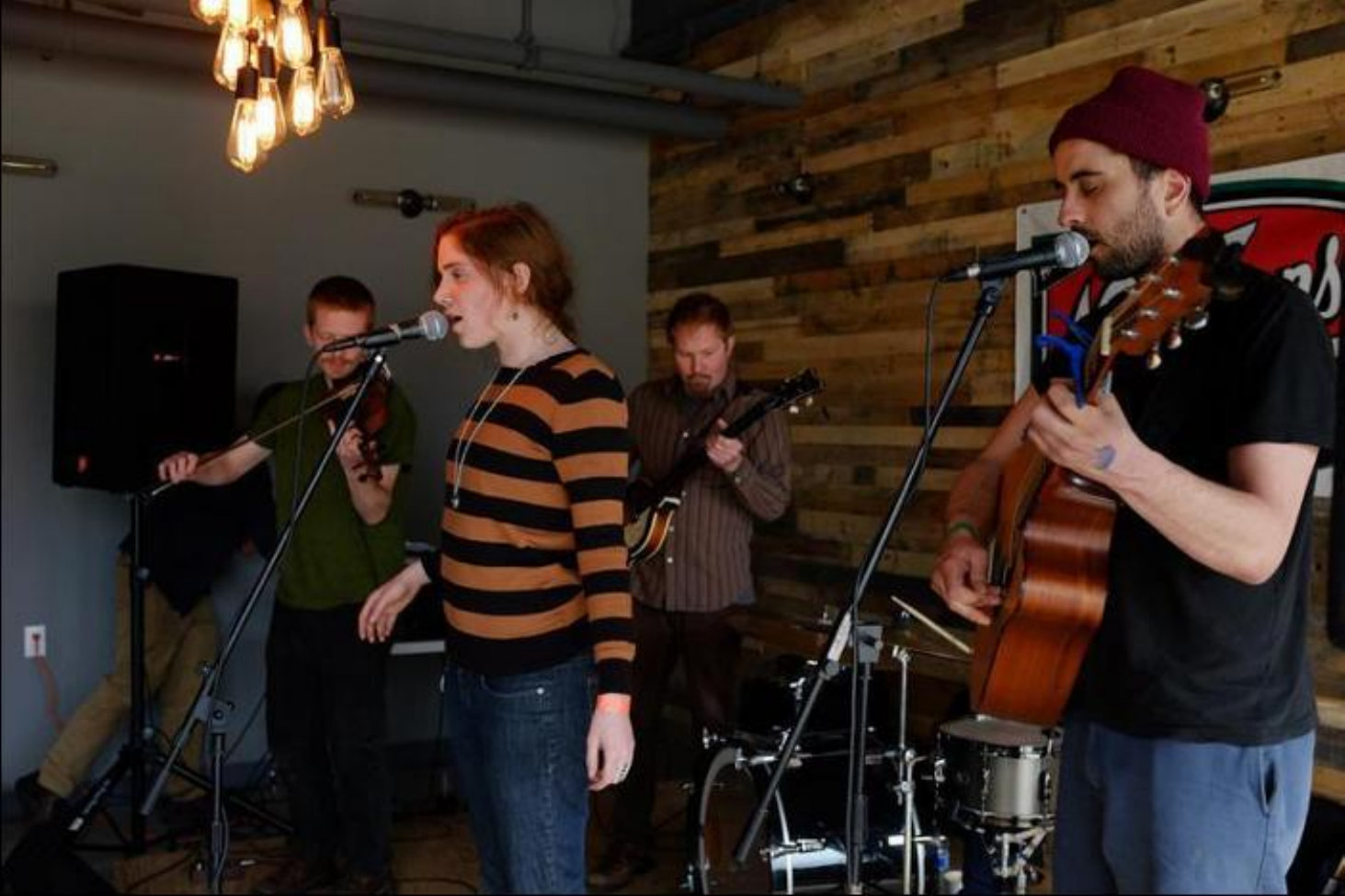}
\end{minipage}

\vspace{1em}

\textbf{Caption}: The image captures a group of four musicians, who appear to be performing
indoors. The setting suggests a relaxed and intimate atmosphere, possibly a small venue or a casual gathering. The background features a wooden plank wall, giving the
environment a rustic and cozy feel. To the upper left, there is a modern hanging light fixture that consists of several exposed bulbs suspended from cords, adding a touch of warmth to the scene. Starting from the left, the first musician is playing a violin. He is dressed in a green shirt and appears to be focused on his instrument. Next to him, slightly to the right and in the foreground, is a female vocalist. She has long hair, which is tied back, and she is wearing a black and orange striped sweater. She is singing into a microphone held by a stand. To her right is another musician who is playing a string instrument, possibly a banjo or a mandolin. He has a beard and is dressed casually in an earthy-toned shirt, and he is standing a bit further back compared to the others. Finally, on the far right is a man playing an acoustic guitar. He is wearing a beanie and a casual jacket. He is also singing into a microphone, contributing to the vocal harmony. Near his microphone, a small portion of a red and green circular sign can be seen, which might be part of the venue’s signage or decor. On the far right, there
is a drum set, though no drummer is visible in the image, suggesting that the band may have more members not captured in this particular frame or they are performing without a drummer at this moment. The musicians appear to be deeply engaged in their performance, with expressions suggesting concentration and genuine involvement in their
music. The use of acoustic instruments and their casual attire suggest that their musical style might be folk or indie. In summary, the image captures a quartet of musicians performing with a violin, a banjo or mandolin, an acoustic guitar, and vocals, in a setting that conveys a warm, intimate, and possibly rustic atmosphere.

\vspace{1em}

\textbf{Rewrite Question}: Determine the artist's possible musical genre, based on the instruments, attire, and atmospheric elements present in the scene. What supporting visual evidence justifies this inference?

\vspace{1em}

\textbf{Rewrite Answer}: Based on the visual elements, it appears that the musical genre is likely folk or indie. The use of acoustic instruments, such as the violin, banjo or mandolin, and acoustic guitar, suggests a focus on traditional or roots music. The casual attire and earthy-toned shirts worn by the musicians also support this inference, implying a relaxed, non-mainstream approach to music. The wooden plank wall and rustic decor in the background reinforce this interpretation, evoking a sense of warmth and authenticity often associated with folk or indie music. Furthermore, the use of microphones and the absence of electric instruments suggest a focus on acoustic performance, which is consistent with these genres.

\end{promptbox}

\section{Model Capabilities and Qualitative Examples}
\label{app:real_examples}

\begin{promptbox}[Comparison with Llava-OneVision-7B Case1]{darkblue}
\begin{minipage}{\textwidth}
    \includegraphics[height=8cm]{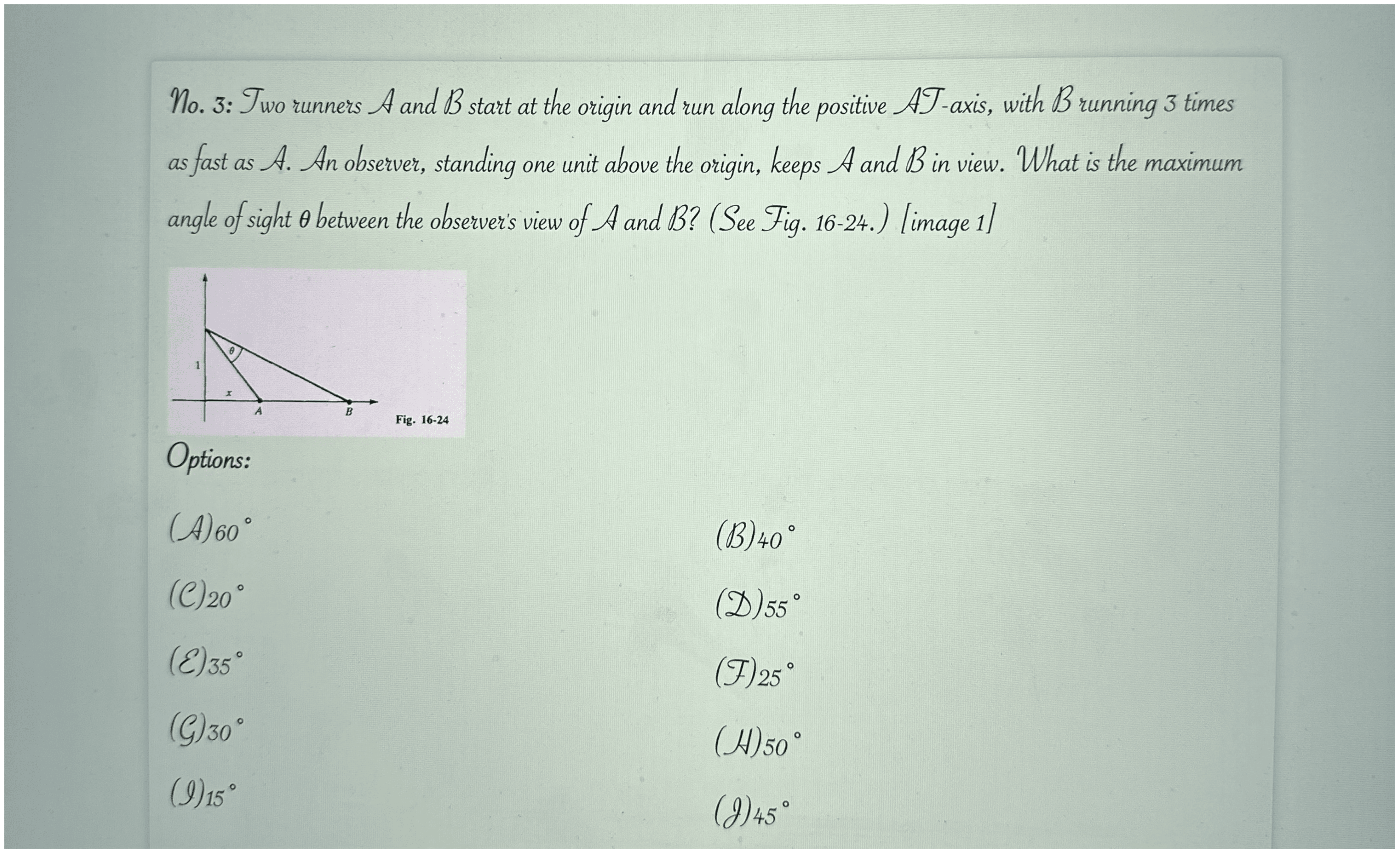}
\end{minipage}

\vspace{1em}

\textbf{\model-8B(\correct)}: To find the maximum angle of sight $\theta$ between the observer's view of A and B, we need to analyze the geometry of the situation.

1. Understanding the Problem: - Runner A starts at the origin and runs along the positive $x$-axis. -Runner B runs 3 times as fast as A and also starts from the origin. - The observer is standing one unit above the origin. \\ 

2. Setting Up the Geometry: - Let's denote the position of runner A at time $t$ as $A(t)$ and the position of runner B as $B(t)$. - Since B runs 3 times as fast as A, if A runs a distance $d$ in time $t$, B will run $3d$ in the same time $t$. \\

3. Finding the Coordinates: - At any time $t$, the coordinates of A are $(t, 0)$. - The coordinates of B are $(3t, 0)$. \\

4. Calculating the Angles: - The angle of sight $\theta$ is the difference between the angles formed by the lines connecting the observer to A and the observer to B. - The angle formed by the line connecting the observer to A is $\alpha = \tan^{-1}\left(\frac{1}{t}\right)$. - The angle formed by the line connecting the observer to B is $\beta = \tan^{-1}\left(\frac{1}{3t}\right)$. \\

5. Maximizing the Angle of Sight: - To maximize $\theta$, we need to find the maximum value of $\alpha - \beta$. - Using calculus, we can find the critical points by taking the derivative of $\alpha - \beta$ with respect to $t$ and setting it to zero. \\

6. Solving for $t$: - After solving the equation, we find that the maximum angle of sight occurs when $t = \frac{1}{\sqrt{3}}$. \\

7. Calculating the Maximum Angle:
\begin{itemize}
\item Substituting $t = \frac{1}{\sqrt{3}}$ into the expressions for $\alpha$ and $\beta$, we get:
\[
\alpha = \tan^{-1}(\sqrt{3}) = 60^\circ
\]
\[
\beta = \tan^{-1}\left(\frac{\sqrt{3}}{3}\right) = 30^\circ
\]
\item Therefore, the maximum angle of sight $\theta$ is:
\[
\theta = \alpha - \beta = 60^\circ - 30^\circ = 30^\circ
\]
\end{itemize}

8. Matching with Options: The closest option to $30^\circ$ is (G) 30°. Thus, the maximum angle of sight between the observer's view of A and B is $30^\circ$. Answer: (G)

\vspace{1em}

\textbf{Llava-OneVision-7B(\wrong)}: The correct answer is D

\end{promptbox}

\begin{promptbox}[Comparison with Llava-OneVision-7B Case2]{darkblue}
\begin{minipage}{\textwidth}
    \includegraphics[height=12cm]{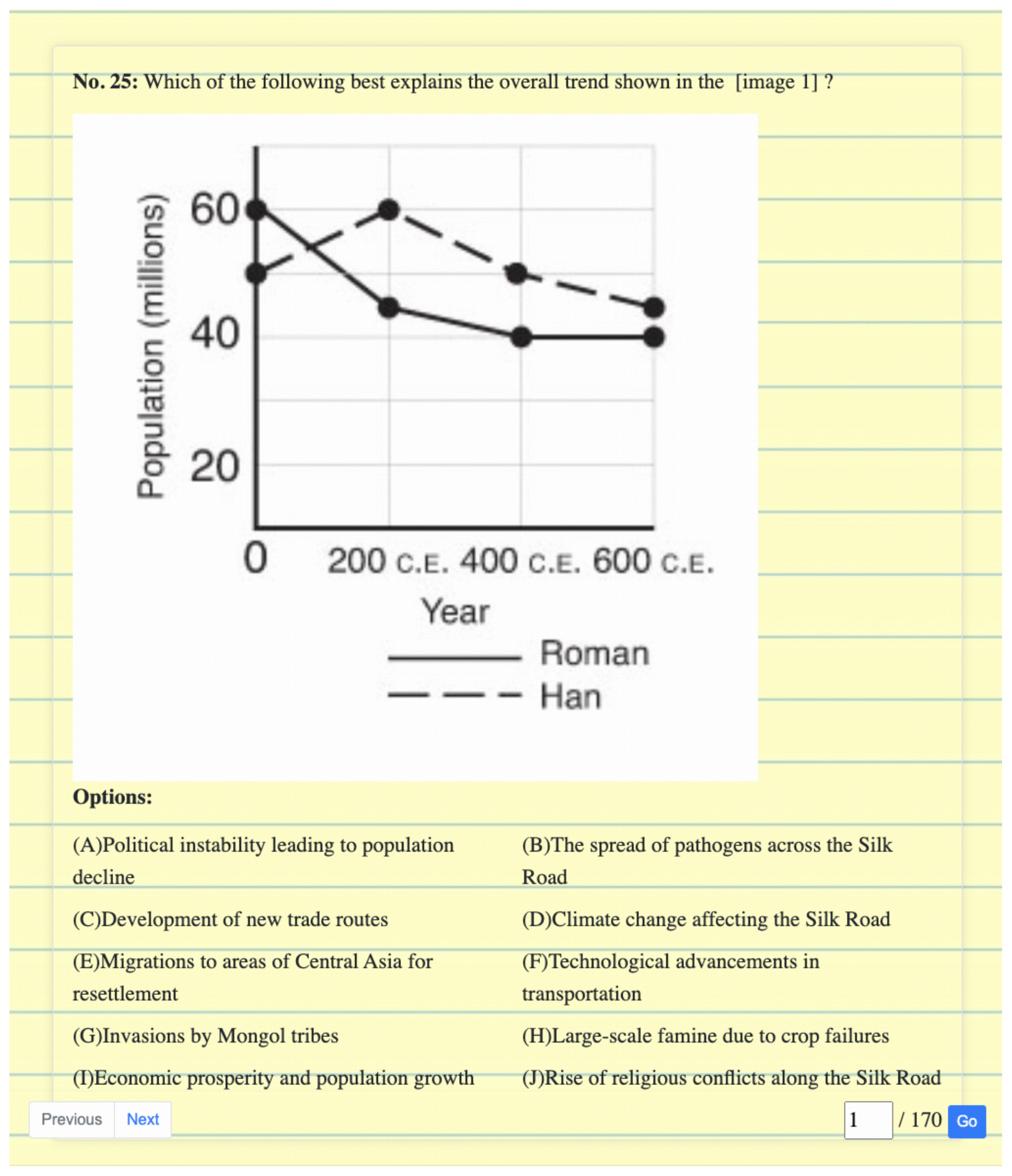}
\end{minipage}

\vspace{1em}

\textbf{\model-8B(\correct)}: To determine the best explanation for the overall trend shown in the graph, we need to analyze the population trends of the Roman and Han empires over time. \\
1. Identify the Trends: The graph shows two lines representing the population of the Roman and Han empires. The Roman line starts at a higher population and declines over time. The Han line starts at a lower population and also declines over time.\\
2. Evaluate the Options:\\ (A) Political instability leading to population decline: This option suggests that political instability could lead to a decline in population. However, it does not explain why the Roman population started higher and declined more significantly than the Han population.\\  (B) The spread of pathogens across the Silk Road: This option implies that the spread of diseases could cause population decline. While this could be a contributing factor, it does not fully explain the observed trends.\\ (C) Development of new trade routes: This option suggests that the development of new trade routes could lead to population growth or stability. However, the graph shows a decline in population for both empires, which contradicts this option.\\ (D) Climate change affecting the Silk Road: This option indicates that environmental factors could impact population trends. While climate change could contribute to population decline, it does not fully explain the specific trends observed in the graph.\\  (E) Migrations to areas of Central Asia for resettlement: This option suggests that migrations could lead to population changes. However, it does not explain why the Roman population started higher and declined more significantly than the Han population. ......\\ 
3. Conclusion:\\  - Based on the analysis, none of the options fully explain the observed trends in the graph. However, option (B) The spread of pathogens across the Silk Road is the most plausible explanation, as it suggests that disease could have contributed to the decline in population for both empires. Therefore, the answer is: Answer: B

\vspace{1em}

\textbf{Llava-OneVision-7B(\wrong)}: The overall trend shown in the image is a decline in population. The graph shows that both the Roman and Han populations decreased over time, with the Roman population declining more sharply than the Han population. This decline could be due to political instability leading to population decline, as option (A) suggests. However, without more context or information about the specific events that occurred during this time period, it is difficult to determine the exact cause of the decline. Nonetheless, based on the information provided in the image, the best explanation for the overall trend shown in the image is option (A).

\end{promptbox}

\begin{promptbox}[Comparison with Llava-OneVision-7B Case3]{darkblue}
\begin{minipage}{\textwidth}
    \includegraphics[height=14cm]{comparison_case/comparison_case2.pdf}
\end{minipage}

\vspace{1em}

\textbf{\model-8B(\correct)}: To find the sensitivity of the new rapid test for HIV, we need to use the information provided in the table. Sensitivity is defined as the proportion of true positive results out of all the actual positive cases.\\
\text{From the table, we can see the following values:}
\begin{itemize}
\item \text{True positive (TP): 378}
\item \text{False negative (FN): 2}
\item \text{True negative (TN): 98,823}
\item \text{False positive (FP): 99,220}
\end{itemize}

\text{The formula for sensitivity is:}
\[ \text{Sensitivity} = \frac{\text{True Positive (TP)}}{\text{True Positive (TP)} + \text{False Negative (FN)}} \]

\text{Plugging in the values from the table:}
\[ \text{Sensitivity} = \frac{378}{378 + 2} = \frac{378}{380} \]

\text{Now, we calculate the decimal value:}
\[ \frac{378}{380} \approx 0.995 \]

\text{Therefore, the sensitivity of the new rapid test for HIV is approximately 0.995.}

\text{Answer: (I)0.995}

\vspace{1em}

\textbf{Llava-OneVision-7B(\wrong)}: The sensitivity of the new rapid test for HIV is 0.985. Therefore, the answer is option B. \\Answer:B
\end{promptbox}

\begin{promptbox}[Multi-Image and Video-1]{darkblue}
\begin{minipage}{\textwidth}
    \includegraphics[width=\linewidth]{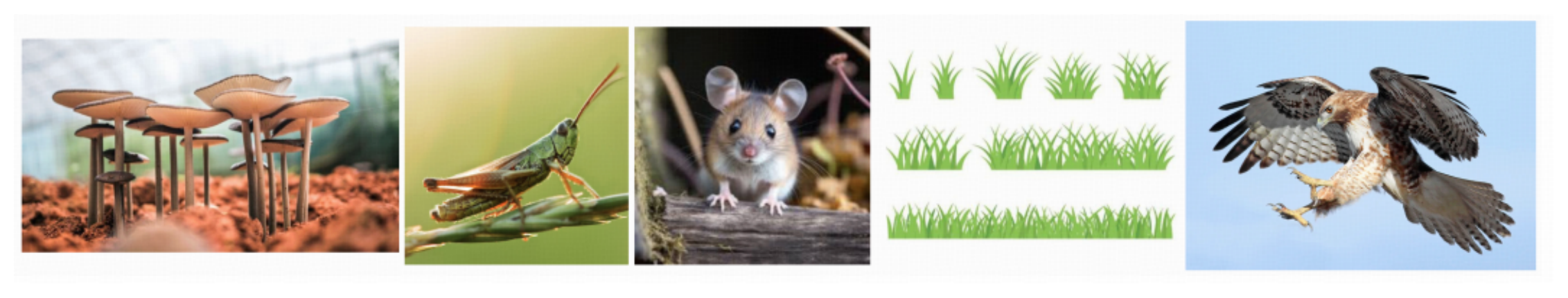}
\end{minipage}

\vspace{1em}

\textbf{User}: What are the accusations of the five species?

\vspace{0.5em}

\textbf{\model-8B}: The grasshopper is an herbivore, feeding on the grass. The mouse is a carnivore, feeding on the grasshopper. The hawk is a carnivore, feeding on the mouse. The grass is a producer, providing energy for the grasshopper. The mushroom is a decomposer, breaking down dead organic matter and recycling nutrients back into the ecosystem.

\vspace{1em}

\textbf{User}: Based on the five species in the images, draw a food chain. Explain the role of each species in the food chain.

\vspace{0.5em}

\textbf{\model-8B}: \\
The food chain can be represented as follows:\\

Grass → Grasshopper → Mouse → Hawk \\

In this food chain, the grass is the primary producer, converting sunlight into energy through photosynthesis. The grasshopper is a primary consumer that feeds on the grass, obtaining energy from it. The mouse is a secondary consumer that feeds on the grasshopper, obtaining energy from it. The hawk is a tertiary consumer that feeds on the mouse, obtaining energy from it. \\

Each species plays a crucial role in the food chain. The grass provides energy for the grasshopper, which in turn provides energy for the mouse. The mouse provides energy for the hawk, which is at the top of the food chain. Without each species, the food chain would collapse, and the ecosystem would be disrupted.\\

\end{promptbox}

\begin{promptbox}[Multi-Image and Video-2]{darkblue}
\begin{minipage}{\textwidth}
    \includegraphics[width=\linewidth]{realcase/realcase_video.pdf}
\end{minipage}

\vspace{1em}

\textbf{User}: Please describe the video in detail.

\vspace{0.5em}

\textbf{\model-8B}: The video begins with a close-up of two individuals, one wearing a white t-shirt and the other in a sleeveless top with a floral pattern. They are standing in a kitchen with a brick wall background, which includes various kitchen items such as bottles and utensils. The scene transitions to a close-up of hands meticulously arranging strips of bacon on a metal baking tray placed on a wooden cutting board. The hands carefully place each strip, ensuring they are evenly spaced. The focus remains on the preparation process, highlighting the precision involved in arranging the bacon. The video then shows a person placing the tray into an oven, with text appearing on the screen that reads '400°F 10-15 minutes,' indicating the cooking instructions. The final scene returns to the two individuals in the kitchen, now engaged in conversation while standing near a counter filled with various ingredients and kitchen tools, including bottles, vegetables, and a bowl.

\end{promptbox}

\clearpage
\begin{promptbox}[Chart \& Doc Understanding-1]{darkblue}

\begin{minipage}{\textwidth}
    \includegraphics[width=\linewidth]{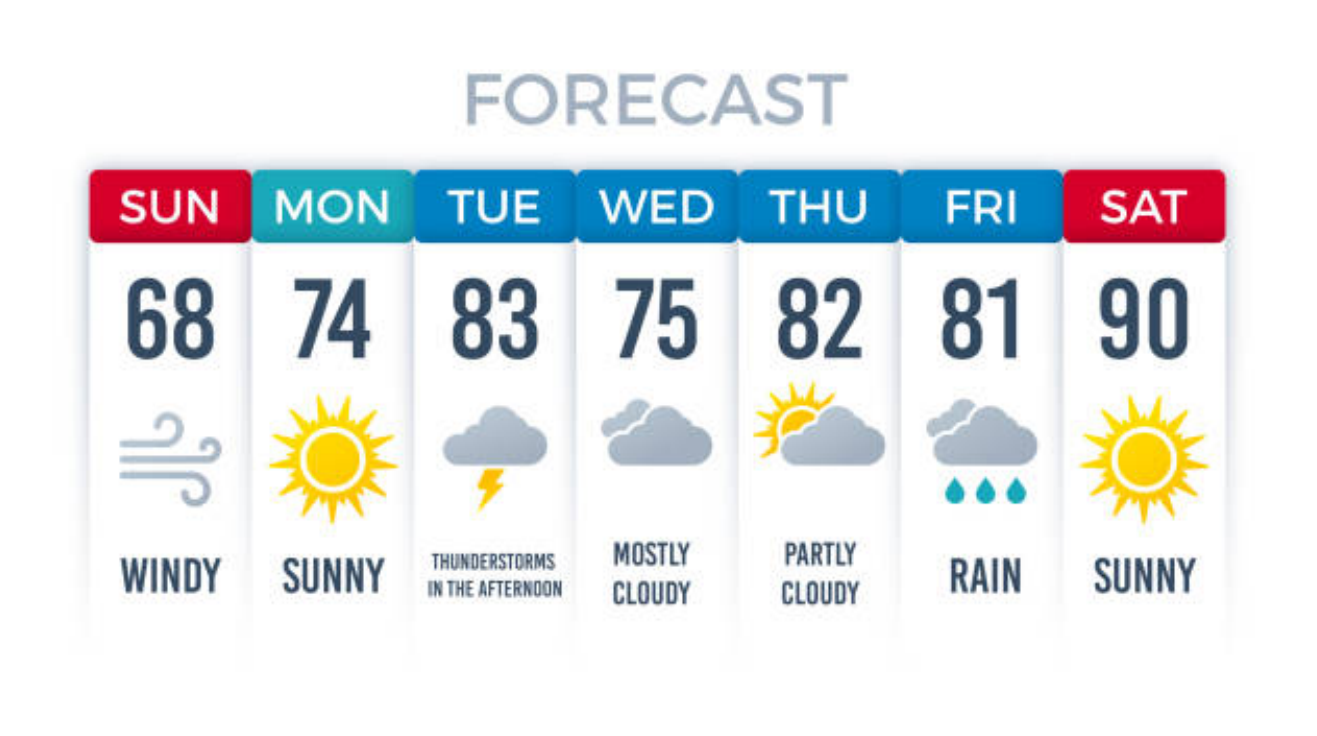}
\end{minipage}

\vspace{1em}

\textbf{User}: List the weather for Monday to Friday

\vspace{0.5em}

\textbf{\model-8B}:\\
Monday: Sunny \\
Tuesday: Thunderstorms in the afternoon \\
Wednesday: Mostly cloudy \\
Thursday: Partly cloudy \\
Friday: Rain \\

\end{promptbox}

\begin{promptbox}[Chart \& Doc Understanding-2]{darkblue}

\begin{minipage}{\textwidth}
    \includegraphics[height=8cm]{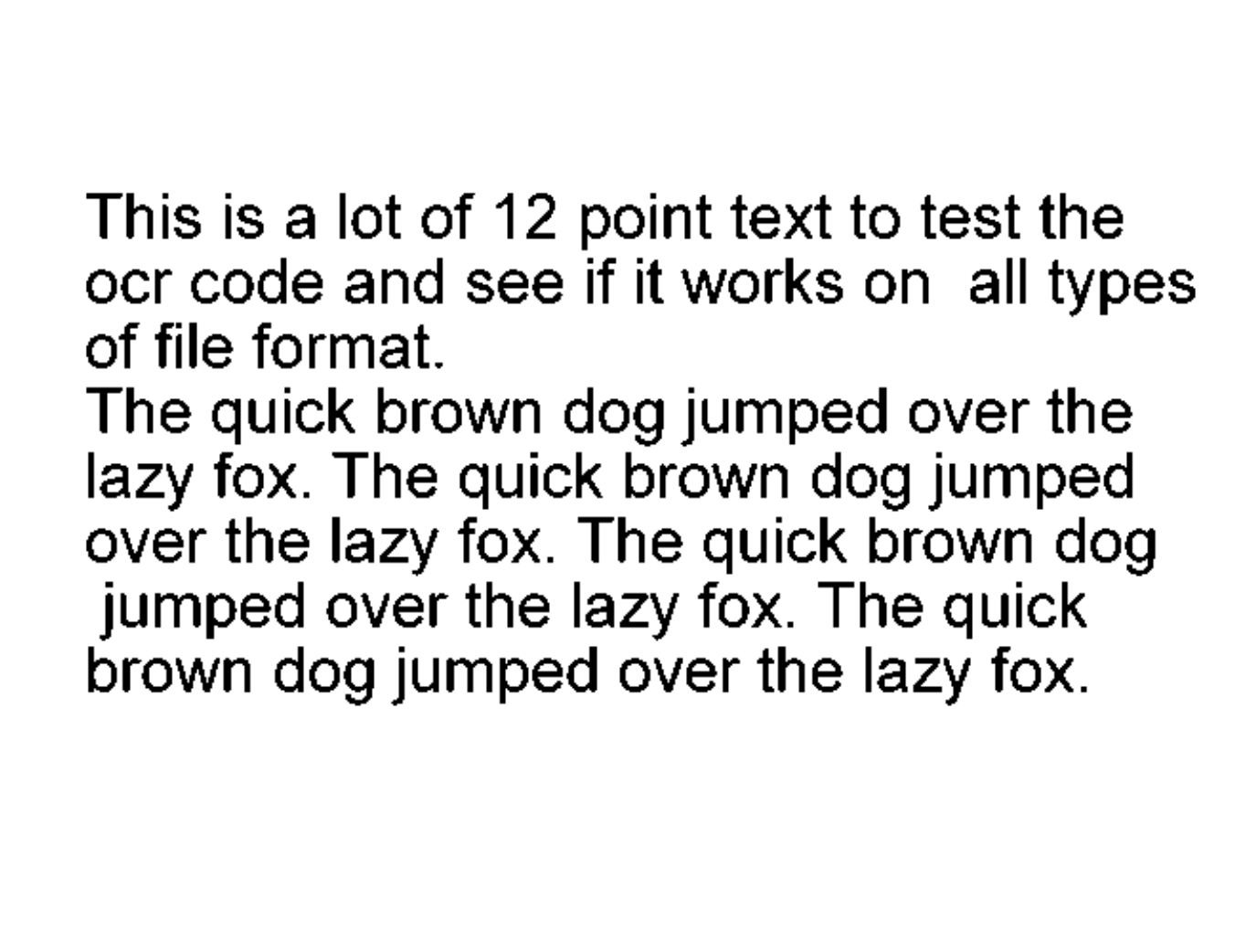}
\end{minipage}

\vspace{1em}

\textbf{User}: Read text in the image

\vspace{0.5em}

\textbf{\model-8B}: This is a lot of 12 point text to test the ocr code and see if it works on all types of file format. The quick brown dog jumped over the lazy fox. The quick brown dog jumped over the lazy fox. The quick brown dog jumped over the lazy fox. The quick brown dog jumped over the lazy fox. The quick brown dog jumped over the lazy fox.

\end{promptbox}

\clearpage
\begin{promptbox}[Multi-Discipline Knowledge and Mathematical Reasoning-1]{darkblue}
\begin{minipage}{\textwidth}
    \includegraphics[height=8cm]{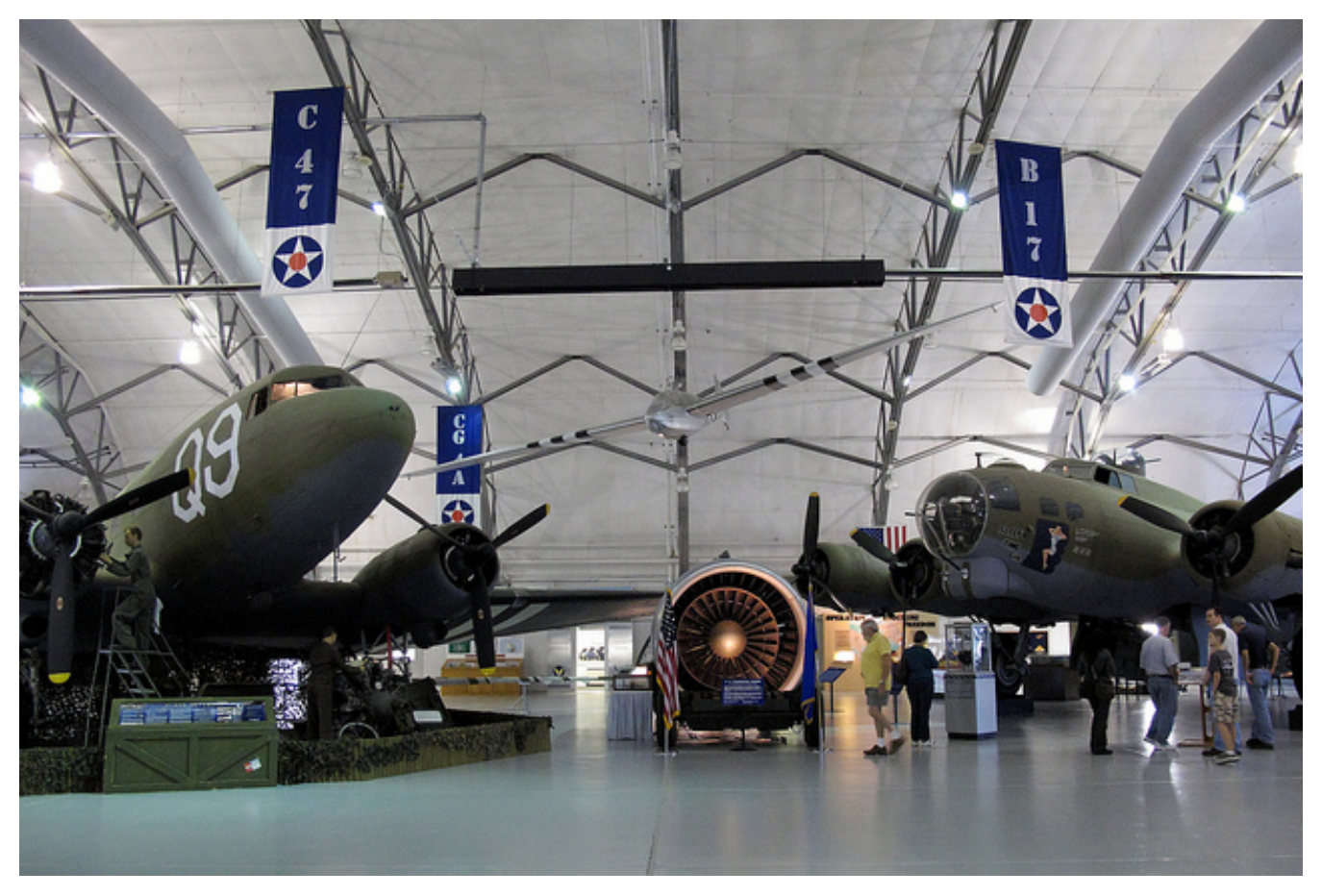}
\end{minipage}

\vspace{1em}

\textbf{User}: Answer the following question based on the provided image: What country do these planes belong to?

\vspace{0.5em}


\textbf{\model-8B}: The planes in the image belong to the United States. This can be inferred from several clues within the image: the design of the aircraft, which is consistent with American military planes from the World War II era; the presence of the U.S. Air Force roundel on the fuselage of the aircraft on the right; and the style of the hangar, which resembles those used by the U.S. military. Additionally, the informational plaques and the overall presentation suggest that this is a museum dedicated to American aviation history. \\

\end{promptbox}

\clearpage
\begin{promptbox}[Multi-Discipline Knowledge and Mathematical Reasoning-2]{darkblue}
\begin{minipage}{\textwidth}
    \includegraphics[height=7cm]{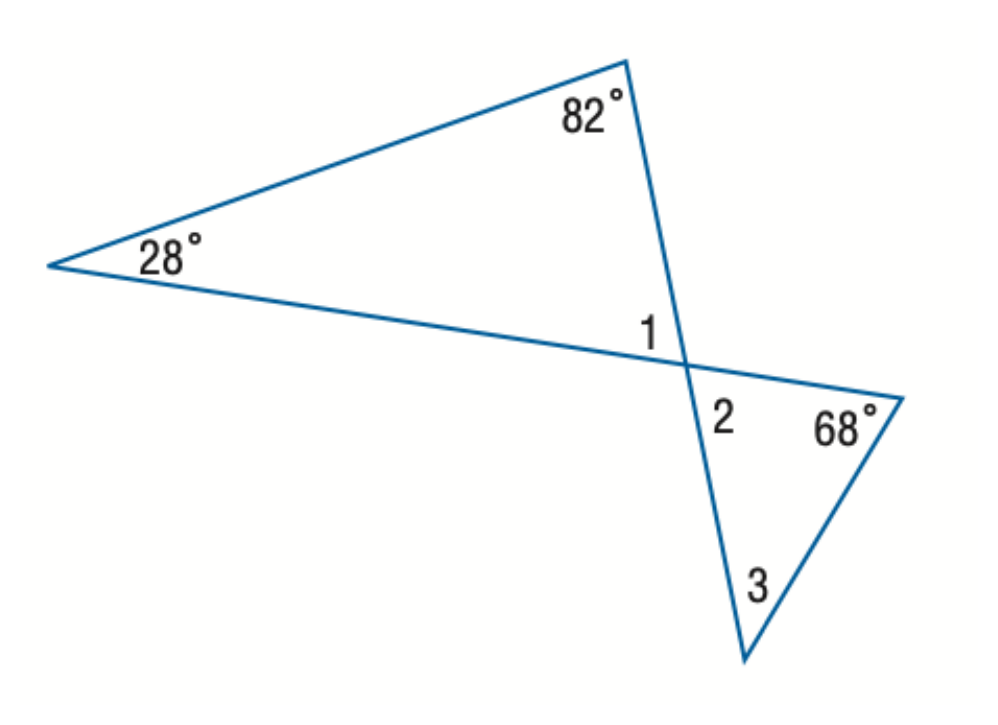} 
\end{minipage}

\vspace{1em}

\textbf{User}: Find the measure of angle 3.

\vspace{0.5em}

\textbf{\model-8B}: To find the measure of angle 3, we need to use the properties of triangles and the fact that the sum of angles in a triangle is always 180 degrees.

1. First, observe the triangle on the left with angles 28° and 82°.
   - The sum of these two angles is:
     \[
     28^\circ + 82^\circ = 110^\circ
     \]

2. Since the sum of all angles in a triangle must be 180°, the third angle in this triangle (which is angle 1) can be found by subtracting the sum of the known angles from 180°:
   \[
   180^\circ - 110^\circ = 70^\circ
   \]

3. Now, consider the triangle on the right with angles 68° and the angle we just found (70°).
   - The sum of these two angles is:
     \[
     68^\circ + 70^\circ = 138^\circ
     \]

4. Again, since the sum of all angles in a triangle must be 180°, the third angle in this triangle (which is angle 3) can be found by subtracting the sum of the known angles from 180°:
   \[
   180^\circ - 138^\circ = 42^\circ
   \]

Therefore, the measure of angle 3 is 42°.

\end{promptbox}

\clearpage
\begin{promptbox}[Multimodal Interactions \& Preferences-1]{darkblue}
\begin{minipage}{\textwidth}
    \includegraphics[height=8cm]{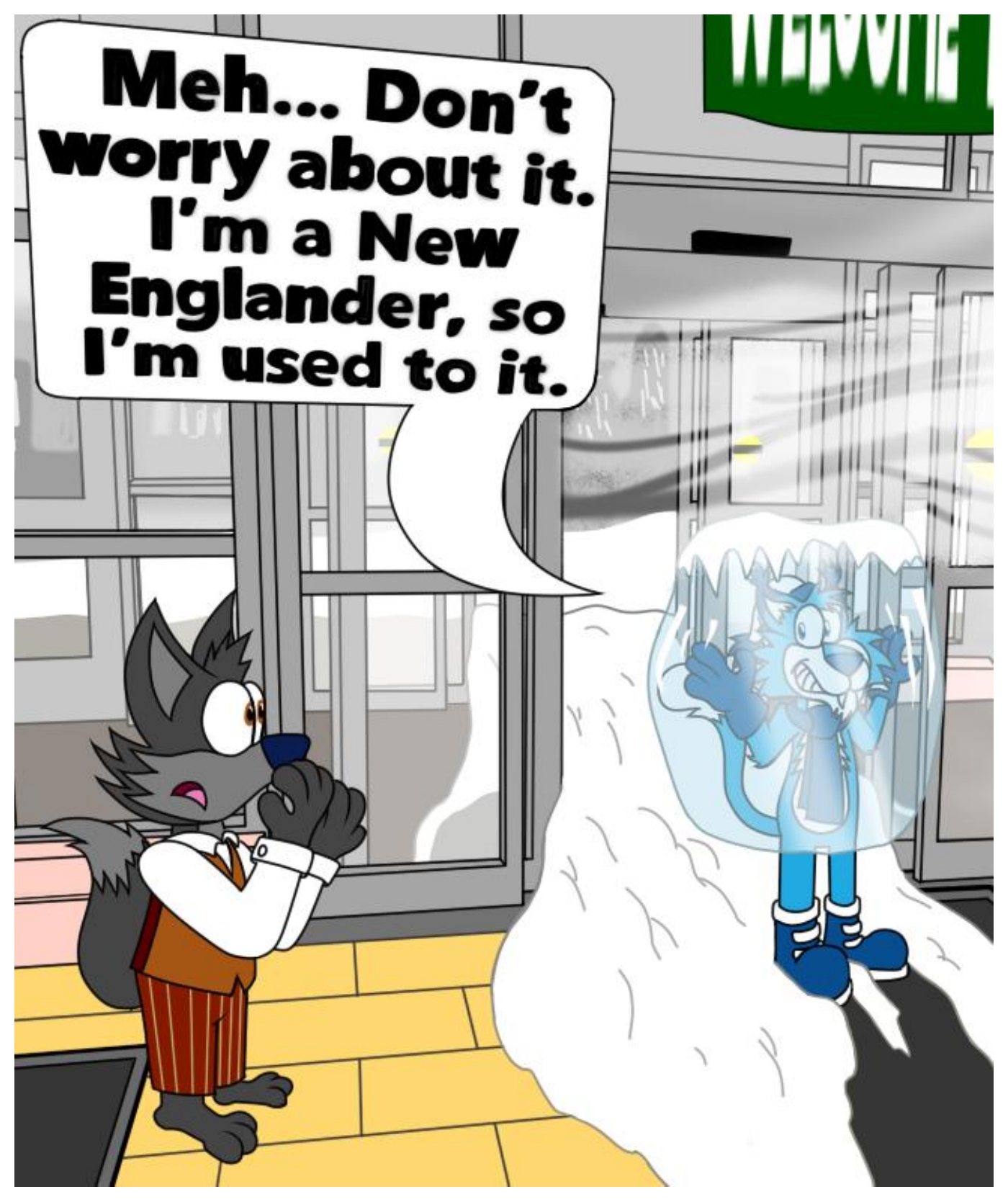}
\end{minipage}

\vspace{1em}

\textbf{User}: Please perfectly describe this cartoon illustration in as much detail as possible

\vspace{0.5em}


\textbf{\model-8B}: The image is a cartoon illustration featuring two anthropomorphic animal characters. On the left, there's a character resembling a gray wolf with large ears, wearing a white shirt with a black vest and red-striped pants. This character has a surprised expression on its face.

On the right side of the image, there's another character that appears to be a blue fox with a similar anthropomorphic design. This character is standing behind a transparent wall or barrier, which seems to be made of ice or glass. The fox character is wearing blue shoes and has a confident, almost smug expression on its face.

Above the fox character, there's a speech bubble containing text that reads: "Meh... Don't worry about it. I'm a New Englander, so I'm used to it." The background of the image shows a store entrance with a sign that says "WELCOME" in green letters. The floor of the store is tiled, and there are reflections visible on the glass doors.

The style of the illustration is cartoonish with bold outlines and bright colors. The overall tone of the image seems to be humorous, as indicated by the speech bubble and the expressions of the characters.

\end{promptbox}

\clearpage
\begin{promptbox}[Multimodal Interactions \& Preferences-2]{darkblue}
\begin{minipage}{\textwidth}
    \includegraphics[height=7cm]{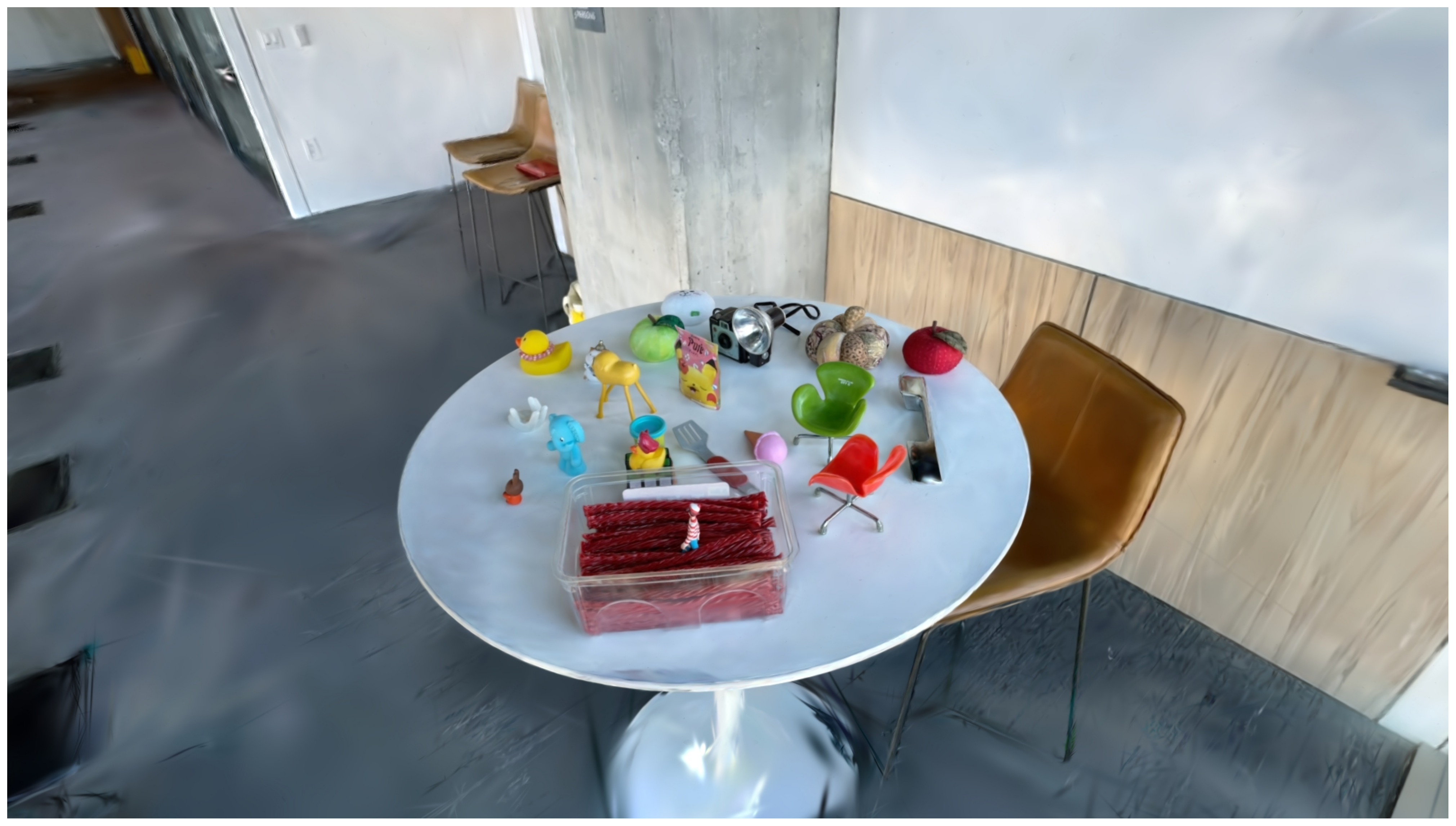}
\end{minipage}

\vspace{1em}

\textbf{User}: This is an image of a room. It could either be a real image captured in the room or a rendered image from a 3D scene reconstruction technique that is trained using real images of the room. A rendered image usually contains some visible artifacts (eg. blurred regions due to under-reconstructed areas) that do not faithfully represent the actual scene. You need to decide if its a real image or a rendered image by giving each image a photorealism score between 1 and 5.

\vspace{0.5em}

\textbf{\model-8B}: The image appears to be a rendered image rather than a real photograph. This assessment is based on several visual cues: the uniformity of the lighting, the lack of shadows that would typically be cast by objects in a real-world setting, and the smoothness of the surfaces which lacks the texture and imperfections usually found in real materials. Additionally, the colors are overly saturated and the reflections are too perfect, which is common in computer-generated imagery. The photorealism score for this image would likely be low, around 1 or 2, due to these factors.

\end{promptbox}

\end{document}